\documentclass[10pt,twocolumn,letterpaper]{article}

\usepackage[pagenumbers]{iccv} 

\newcommand{\finding}[1]{\textbf{#1}}

\definecolor{iccvblue}{rgb}{0.21,0.49,0.74}
\usepackage[pagebackref,breaklinks,colorlinks,allcolors=iccvblue]{hyperref}
\usepackage{multirow}
\usepackage{tabulary}
\usepackage[draft]{changes}
\usepackage{csquotes}
\usepackage{mathtools}

\definechangesauthor[color=BurntOrange]{MK}

\def\paperID{9063} 
\def\confName{ICCV}
\def\confYear{2025}

\title{Benchmarking Vision Foundation Models for \\ Input Monitoring in Autonomous Driving}

\author{
Mert\,Keser$^{1,2}$ \quad Halil Ibrahim\,Orhan$^{1,2}$ \quad Niki\,Amini-Naieni$^4$ \quad Gesina\,Schwalbe$^{3}$ \\ \quad  Alois\,Knoll$^2$ \quad Matthias\,Rottmann$^5$\\
$^1$ Continental AG \quad $^2$ Technical University of Munich \quad $^3$ University of Lübeck \\ \quad $^4$ University of Oxford \quad $^5$ University of Wuppertal \\
\tt\small {\{\href{mailto:mert.keser@continental.com}{mert.keser}, \href{mailto:halil.ibrahim.orhan-EXT@continental.com}{ibrahim.orhan}\}@continental.com} \\ 
\tt\small {\{\href{mailto:mert.keser@tum.de}{mert.keser}, \href{mailto:halil.ibrahim.orhan-EXT@continental.com}{ibrahim.orhan}, \href{mailto:k@tum.de}{k}\}@tum.de} \\
\tt\small {\href{mailto:niki.amini-naieni@eng.ox.ac.uk}{niki.amini-naieni@eng.ox.ac.uk}, \href{mailto:gesina.schwalbe@uni-luebeck.de}{gesina.schwalbe@uni-luebeck.de}, \href{mailto:rottmann@uni-wuppertal.de}{rottmann@uni-wuppertal.de} }
}

\begin{document}
\maketitle
\begin{abstract}

Deep neural networks (DNNs) remain challenged by distribution shifts in complex open-world domains like automated driving (AD): Robustness against yet unknown novel objects (semantic shift) or styles like lighting conditions (covariate shift) cannot be guaranteed.
Hence, reliable operation-time monitors for identification of out-of-training-data-distribution (OOD) scenarios are imperative.
Current approaches for OOD classification are untested for complex domains like AD, are limited in the kinds of shifts they detect, or even require supervision with OOD samples. 
To prepare for unanticipated shifts, we instead establish a framework around a principled, unsupervised and model-agnostic method that unifies detection of semantic and covariate shifts: Find a full model of the training data's feature distribution, to then use its density at new points as in-distribution (ID) score.
To implement this, we propose to combine Vision Foundation Models (VFMs) as feature extractors with  
density modeling techniques.
Through a comprehensive benchmark of 4 VFMs with different backbone architectures and 5 density-modeling techniques against established baselines, we provide the first systematic evaluation of OOD classification capabilities of VFMs across diverse conditions. A comparison with state-of-the-art binary OOD classification methods reveals that VFM embeddings with density estimation outperform existing approaches in identifying OOD inputs.
Additionally,  
we show that our method detects high-risk inputs likely to cause errors in downstream tasks, thereby improving overall performance. 
Overall, VFMs, when coupled with robust density modeling techniques, are promising to realize model-agnostic, unsupervised, reliable safety monitors in complex vision tasks.

\end{abstract}

\section{Introduction}
\label{sec:intro}



\begin{figure}
\centering
\includegraphics[width=\linewidth]{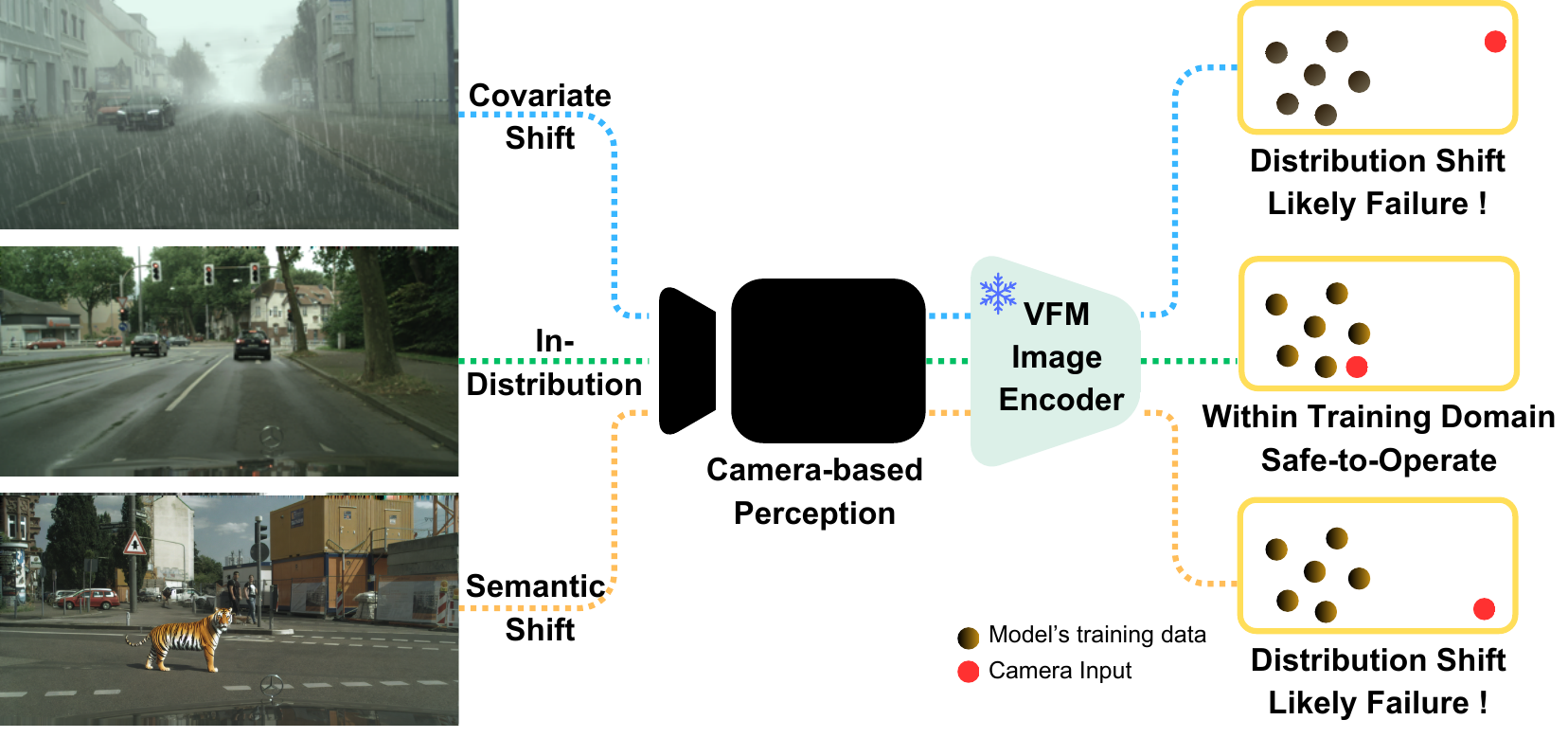}
\vspace*{-\baselineskip}
\caption{A monitoring system for autonomous driving that uses a pre-trained Vision Foundation Model’s image encoder to detect distribution shifts in input data. The system identifies whether camera inputs (red dots) fall within the model’s training distribution (brown dots) or represent covariate shifts (top) or semantic shifts (bottom), helping assess operational safety and potential failure risks.
}
\label{fig:Network_Monitoring}
\vspace*{-1.3\baselineskip}
\end{figure}



In recent decades, deep neural networks (DNNs) have become widely used for environmental perception due to their superior performance compared to traditional computer vision methods \cite{chen2017deeplab, cheng2024yolo, zhao2024detrs, pi2024perceptiongpt, shi2024transnext, liu2021swin, liu2023grounding}.
However, in complex open-world domains such as autonomous driving (AD) and robot navigation this performance is inevitably challenged by the \enquote{long tail} of the input distribution, i.e., unforeseeable out-of-distribution (OOD) scenarios for which DNN generalization cannot be guaranteed.
AD spans multiple operational levels, from driver assistance (L1-L2) to fully autonomous operation (L4-L5), with conditional automation (L3) bridging the transition \cite{SAEJ3016}. Across these levels, camera-based perception systems support critical functions, from lane-keeping and traffic sign recognition to full scene understanding \cite{CarADAS_LKA}. As these systems operate in increasingly diverse environments, a fundamental question emerges: \emph{How can we ensure that perception models function reliably in complex, open-world scenarios, particularly when encountering inputs that deviate from their training distribution?}
To mitigate these risks and ensure safe operation, reliable monitors that can detect and respond to such deviations have become mandatory \cite{euaiactcompromise2024, ISO8800}. 
Aside from model uncertainty and behavior monitoring \cite{rahman2021run, gupta2024detecting}, 
an important approach for OOD detection is input monitoring (\cref{fig:Network_Monitoring}), which identifies when input data fed to the system deviates from the training data distribution \cite{rahman2021run}.

Existing binary OOD classification methods suffer from critical limitations for AD applications: they often require direct access to model internals, are specialized to specific architectures, or need exposure to anticipated shift types during training \cite{huang2021importance, liu2020energy, macedo2021entropic, zhang2023decoupling, sun2022out}. These constraints compromise reliability when encountering truly unforeseen distribution shifts in open-world environments -- a critical limitation for autonomous driving systems that must handle diverse operational domains. Furthermore, accessing internal model layers becomes impractical in AD systems. A more principled approach would model the complete training data distribution in feature space, using density values or likelihoods as in-distribution (ID) scores for new samples. The challenge for complex visual domains lies in extracting sufficiently powerful features—standard DNNs and autoencoders (AE) trained on limited datasets typically fail to capture the necessary representational richness. 

This is precisely where VFMs offer a compelling solution. Trained on diverse, large-scale datasets, VFMs encode comprehensive semantic contexts and subtle visual distinctions within their latent representations \cite{caron2021emerging, oquab2023dinov2}, making them particularly well-suited for distribution modeling. Their appeal for input monitoring in AD stems from demonstrated strong generalization capabilities across diverse visual tasks and novel scenarios \cite{ilyas2024potential, nekrasov2024oodis}, potentially enabling robust detection of unforeseen distribution shifts without requiring model-specific adaptations or prior exposure to OOD examples.

In this study we investigate: \emph{How suitable are VFM image encodings for modeling a data distribution for OOD scenario detection in complex vision domains like AD?} In particular: \emph{How do they compare to standard DNN encodings and SoTA baseline binary-OOD detection methods? Is this OOD of use for improving overall system reliability in downstream applications?} 
Also, while most existing works focuses on either semantic OOD detection on object level \cite{chan2021segmentmeifyoucan,bevandic2021dense,chan2021entropy} or on covariate shifts detection \cite{bolte2019unsupervised,hell2021monitoring}, we are interested in monitoring capabilities for \emph{both} semantic, covariate, and combined shift detection on \emph{scene} level.
Our \textbf{contributions} are:
\begin{enumerate}[leftmargin=1.5em, label=(\roman*), itemsep=0pt, parsep=0pt, topsep=0pt]  

\item We propose a \textbf{modular pipeline} for \textbf{unsupervised}, \textbf{model-agnostic} detection of \textbf{distribution shifts} in input images (see \cref{fig:Network_Monitoring}), in particular both \textbf{semantic, covariate, and joint shifts}.
At its core we use rich VFM feature encodings for distribution density modeling.
\item We for the first time \textbf{benchmark} VFMs for OOD on the challenging open-world domain of \textbf{AD}. Our extensive tests compare 4 VFMs on 12 both synthetic and real datasets, 
and 7 OOD methods as baselines. Our results show the clear superiority of VFM encodings in unified distribution shift detection.

\item We demonstrate \textbf{practical usability} on a downstream AD perception application, showcasing that, despite being model-agnostic, VFMs successfully \textbf{identify OOD samples with higher risk of errors}.
\item We establish a comprehensive \textbf{benchmarking framework} for evaluating binary OOD classification methods in AD by providing diverse experimental scenarios combining different types of distribution shifts, datasets, and evaluation protocols.

\end{enumerate}

\section{Related Work}
\label{sec:relatedWork}



\subsection{Semantic Shift in Autonomous Driving}
\label{sec:semantic_shift}

In machine learning, training and testing sets are typically drawn from the same distribution, which poses the closed-world assumption \cite{rahman2021run} that the set of considered semantic classes will not change \cite{pinggera2016lost,hendrycks2016baseline}. To identify semantic shifts (novel classes), techniques like Monte Carlo dropout estimate neural network uncertainty \cite{gal2016dropout}, finding applications in semantic segmentation \cite{kendall2015bayesian} and object detection \cite{he2017mask, heidecker2021towards}. Recent advances integrate semantic segmentation with anomaly detection \cite{bevandic2021dense, chan2021entropy} and employ various reconstruction approaches \cite{vojir2021road, ohgushi2020road, wu2023discriminating} for object-level OOD detection.
Alternative feature-based approaches include SAFE \cite{wilson2023safe}, which teaches detectors to recognize sensitivity-aware features through input perturbation, and FOOD-ID \cite{li2022out}, which employs likelihood-based feature space mechanisms. However, existing solutions typically rely on model-specific architectures, OOD training examples, computationally expensive generative models, or are limited to specific anticipatable shift types. In contrast, pre-trained VFMs enable unsupervised, model-agnostic detection of distribution shifts without these limitations.



\subsection{Covariate Shift in Autonomous Driving}
\label{sec:covariate_shift}



The distribution of real-world scenarios often diverges from training data, a phenomenon known as covariate shift \cite{schneider2020improving}, distributional shift \cite{hell2021monitoring}, or domain shift \cite{lohdefink2020self}. In autonomous driving, this occurs due to changes in lighting conditions and weather patterns. Comparative studies \cite{hell2021monitoring} found likelihood regret \cite{xiao2020likelihood} and self-supervised approaches \cite{sehwag2021ssd} outperform variational autoencoders \cite{stocco2020misbehaviour} for covariate shift detection. Other methods include Earth Mover's Distance for linking domain mismatch with performance degradation \cite{lohdefink2020self}, feature map error analysis \cite{bolte2019unsupervised}, and GAN-based validation frameworks \cite{zhang2018deeproad}.

Similar to semantic shift detection, these approaches rely on model-specific architectures, require supervision with specific shift examples, or depend on computationally expensive generative models. Our work addresses these fundamental limitations by leveraging VFMs' rich representational power for model-agnostic distribution shift detection that unifies both covariate and semantic shift identification.

\subsection{Vision Foundation Models}
\label{sec:Vision_Foundation_Models}
FMs are large-scale machine learning models trained on diverse data, allowing them to adapt to varied tasks beyond their original objective \cite{wu2024prospective}. The field includes models like CLIP \cite{radford2021learning}, trained contrastively on image-text pairs; DINO \cite{caron2021emerging}, using self-supervised learning with self-distillation; DINOv2 \cite{oquab2023dinov2}, enhancing DINO with masked image modeling; Grounding-DINO \cite{liu2023grounding}, extending DETR-based detection \cite{zhang2022dino} with text guidance; InternImage \cite{wang2023internimage}, utilizing deformable convolutions for adaptive receptive fields; Florence-2 \cite{xiao2024florence}, employing hierarchical visual tokenization; and LLaVA \cite{liu2023visual}, connecting vision encoders to large language models through visual instruction tuning. This study focuses on the first four VFMs, selected for their architectural diversity (spanning Convolutional Networks, Vision Transformers, and Swin Transformers) and absence of test data leakage (see Supplementary). For a comprehensive overview of FMs, we refer to \cite{bommasani2021opportunities, awais2025foundation}, while \cite{zhou2024vision} provides a detailed review of VFMs specifically in AD.

Note that, while CLIP \cite{radford2021learning} and GroundingDINO \cite{liu2023grounding} may also accept language inputs, we here for now only consider their image encodings.

\section{Shift Detection Methodology \& Evaluation}

\subsection{Shift Detection Methodology: Overview}
\label{sec:problemDefinition}

We propose a method to detect any distribution shift in image data using a DNN $F\colon x\mapsto f$ for feature extraction (a VFM, standard DNN, or AE).
Concretely, we aim to classify whether a new image $x^*$ is an outlier with respect to the distribution of a given (monitor training) dataset $\mathcal{X}_\mathrm{monitor}$ (the ID data). 
In our case we choose $\mathcal{X}_\mathrm{monitor}=\mathcal{X}_\mathrm{perception}$ (e.g., Cityscapes train set \cite{cordts2016cityscapes}), the dataset the perception system is trained on.
During monitor \textbf{training}, the monitor is then parametrized on the (fixed) features $\mathcal{F}_\mathrm{monitor}\coloneqq F(\mathcal{X}_\mathrm{monitor})$ to capture the distribution $\mathcal{F}_\mathrm{monitor}$.
From the distribution model we then derive a function $c$ that provides an ID score $c(x^*)=\mathrm{score}(f^*)$ for any new input $x^*$, based on its similarity to or density in the monitor data feature distribution.
Using few ID and OOD samples, a decision threshold $t$ is determined.
For monitor \textbf{inference} on new samples $x*$, $c(x^*)>t$ is calculated to decide the OOD vs.\ ID binary classification.
For \textbf{testing}, we combine an ID dataset (e.g., Cityscapes validation set \cite{cordts2016cityscapes})  
%
with an OOD dataset (e.g., Foggy Cityscapes \cite{sakaridis2021acdc}),
both containing samples not seen by the monitor previously:
$\mathcal{X}_\mathrm{test} = \mathcal{X}_\mathrm{ID} \cup \mathcal{X}_\mathrm{OOD}$.
The OOD samples differ from $\mathcal{X}_\mathrm{monitor}$ through either semantic shifts (e.g., novel object classes) or covariate shifts (e.g., varying lighting conditions or adverse weather).

We consider the following \textbf{choices for the monitor} distribution modeling further detailed in the following in order of increasing complexity:
1) instance-based: averaged pairwise similarities to the training examples and its computationally efficient derivation using similarity to the mean feature vector, 2) a one-class support vector machine, 3) a Gaussian mixture model and 4) a normalizing flow.

\vspace{-2mm}
\paragraph{Average Pairwise Similarity (APS) and Mean Feature Similarity (MFS): }
In this method, we compute the similarity between a test image and training images using cosine similarity. The test set includes both $\mathcal{X}_\mathrm{ID}$ and $\mathcal{X}_\mathrm{OOD}$ images. Cosine similarity is chosen for several reasons: (1) its effectiveness in measuring the similarity between high-dimensional vectors, particularly when the magnitude of the vectors is less important than their direction; (2) its invariance to scaling, which is beneficial when dealing with normalized feature vectors; and (3) its natural alignment with the latent space of many pre-trained foundational models used for feature extraction. Notably, models like CLIP and DINO are often trained using contrastive losses based on cosine similarity, making this metric particularly suitable for comparing features in their embedding spaces. For a new test image $x^*$ with feature vector $f^*$, we calculate its average cosine similarity to all training samples.
\begin{equation}
c_{\text{APS}}(x^*) = \frac{1}{n} \sum_{i=1}^n \cos(f^*, f_i) = \frac{1}{n} \sum_{i=1}^n \frac{f^* \cdot f_i}{||f^*|| \, ||f_i||}
\end{equation}
where $\cos(f^*, f_i)$ denotes the cosine similarity between vectors $f^*$ and $f_i$, $\cdot$ denotes the dot product, and $||\cdot||$ is the L2 norm. The resulting similarity score $c(x^*)$ ranges from -1 to 1, with higher values indicating greater similarity.

For real-time applications, we also implement MFS, a computationally efficient variant that first computes a mean feature vector $\bar{f} = \frac{1}{n}\sum_{i=1}^n f_i$ from training samples, then calculates similarity to this prototype:
\begin{equation}
c_{\text{MFS}}(x^*) = \cos(f^*, \bar{f}) = \frac{f^* \cdot \bar{f}}{||f^*|| \, ||\bar{f}||}
\end{equation}
This reduces inference complexity from $O(n)$ to $O(1)$, making it suitable for monitoring systems with constrained computational resources.

\vspace{-2mm}
\paragraph{One-Class Support Vector Machine (OC-SVM):}
OC-SVM \cite{bounsiar2014one, alam2020one} is an unsupervised machine learning algorithm designed for novelty detection. It is particularly effective in scenarios where there is an abundance of data from one class (the "normal" class) and the objective is to identify instances that deviate from this norm. In our context, we employ OC-SVM to model the distribution of features extracted from $\mathcal{X}_\mathrm{monitor}$ images and subsequently detect samples from $\mathcal{X}_\mathrm{ID}$ and $\mathcal{X}_\mathrm{OOD}$ sets.
For a set of $\mathcal{X}_\mathrm{monitor}$ features $\mathcal{F}_\mathrm{monitor} = \{f_1, f_2, \ldots, f_n\}$ extracted from $\mathcal{X}_\mathrm{monitor}$ set, the OC-SVM aims to find a hyperplane that separates the data from the origin with maximum margin in the feature space. This is achieved by solving the following optimization problem:\\[-\baselineskip]
\begin{align}
\SwapAboveDisplaySkip
&\min_{w, \xi, \rho} \frac{1}{2} ||w||^2 + \frac{1}{\nu n} \sum_{i=1}^n \xi_i - \rho
\\
\text{subject to:}\quad
&(w \cdot \phi(f_i)) \geq \rho - \xi_i, \quad \xi_i \geq 0
\end{align}
where $w$ is the normal vector to the hyperplane, $\phi$ is a feature map that implicitly maps the input features to a higher-dimensional space, $\xi_i$ are slack variables, $\rho$ is the offset of the hyperplane from the origin, and $\nu \in (0, 1]$ is a parameter that controls the trade-off between maximizing the distance of the hyperplane from the origin and the fraction of training examples that are allowed to fall on the wrong side of the hyperplane.
For a new test image $x^*$ with feature vector $f^*$, the decision function is:
\begin{gather}
\SwapAboveDisplaySkip
c_{\text{OC-SVM}}(x^*) = (w \cdot \phi(f^*)) - \rho
\end{gather}
We utilize these decision function values as anomaly scores, with lower scores indicating a higher likelihood of the sample being OOD. 



\vspace{-2mm}
\paragraph{Gaussian Mixture Model (GMM):}
We model the features $\mathcal{F}_\mathrm{train}$ using a Gaussian mixture model (GMM) \cite{Murphy2021} of $K$ Gaussians with the formulation
\begin{gather}
\SwapAboveDisplaySkip
p(f_i | \Theta) = \sum_{k=1}^K \pi_k \mathcal{N}(f_i | \mu_k, \Sigma_k) ,
\end{gather}
Therein, $\Theta = \{\pi_1, \ldots, \pi_K, \mu_1, \ldots, \mu_K, \Sigma_1, \ldots, \Sigma_K\}$ represents the GMM parameters, where $\pi_k$ is the mixing coefficient (with $\sum_{k=1}^K \pi_k = 1$ ensuring a valid probability distribution), $\mu_k$ the mean vector, and $\Sigma_k$ the covariance matrix of the $k$-th Gaussian component.

The parameters $\Theta$ of the GMM are derived from the feature set $\mathcal{F}_\mathrm{train}$ using the Expectation-Maximization (EM) algorithm. To optimally balance complexity and fitting accuracy, the model's number of components $K$ is determined through the Akaike Information Criterion (AIC). This criterion is applied by testing a series of $K$ values on $\mathcal{F}_\mathrm{monitor}$. The choice of $K$ is discussed in the Appendix.





We further assume to have an ID set $\mathcal{X}_\mathrm{ID}$ with corresponding features $\mathcal{F}_\mathrm{ID}$ and analogously OOD set (containing either semantic or covariate shift) $\mathcal{X}_\mathrm{OOD}$ with corresponding features $\mathcal{F}_\mathrm{OOD}$. Let $\mathcal{X} = \mathcal{X}_\mathrm{ID} \cup \mathcal{X}_\mathrm{OOD}$ and $\mathcal{F} = \mathcal{F}_\mathrm{ID} \cup \mathcal{F}_\mathrm{OOD}$.
For a new test image $x^* \in \mathcal{X}$, we extract and normalize its feature vector $f^*$ using the given foundational model currently under inspection. To compute  confidence scores, the log-likelihood of $f^*$ under the learned GMM is computed as
    \begin{gather}
    \SwapAboveDisplaySkip
    c_{\text{GMM}}(x^*) = \log p(f^* | \Theta) = \log \left( \sum_{k=1}^{K} \pi_k \mathcal{N}(f^* | \mu_k, \Sigma_k) \right)
    \end{gather}
where \( \Theta = \{\pi_k, \mu_k, \Sigma_k\}_{k=1}^{K} \) is the set of parameters for the GMM, learned from $\mathcal{F}_\mathrm{train}$.

\vspace{-2mm}
\paragraph{Normalizing Flows (NF):} Normalizing flows \cite{papamakarios2021normalizing, dinh2016density} provide a deep learning framework for high-dimensional density estimation through invertible mappings. Let $P_F$ denote the probability measure of VFM features $f_1,\ldots,f_n$ from $\mathcal{F}_\mathrm{monitor}$, and $P_\mathcal{N}$ denote the measure of the multivariate standard normal distribution $\mathcal{N}(\cdot| 0, I)$. A normalizing flow learns an invertible map $g_\theta$ such that:
\begin{gather}
    \SwapAboveDisplaySkip
     P_F \circ g_\theta^{-1} \approx P_\mathcal{N}
\end{gather}
where $g_\theta^{-1}$ denotes the pre-image.
We want to use the density corresponding to $P_F$ at a feature vector $f$ to derive an ID-score for $f$.
Given $g_\theta$ as above, the density can be comfortably estimated as $\mathcal{N}(g_\theta(f)| 0, I)\cdot|\det(J_f g_\theta(f))|$, where $J_f g_\theta$ is the Jacobian of $g_\theta$ with respect to $f\in \mathbb{R}^d$.

We implement $g_\theta$ using a Real-valued Non-Volume Preserving (Real-NVP) \cite{dinh2016density} architecture with affine coupling layers. Each layer employs a binary mask to split input dimensions into $x_1$ and $x_2$, producing outputs:
\begin{align}
y_1 &= x_1
&y_2 &= x_2 \odot \exp(s(x_1)) + t(x_1)
\end{align}
Here, $s(\cdot)$ and $t(\cdot)$ are neural networks computing scale and translation respectively, and $\odot$ is the element-wise product. This architecture's Jacobian determinant simplifies to $\log|\det(J)| = \sum_i s(x_1)_i$, enabling efficient likelihood computation. For OOD detection, we normalize the negative log-likelihood of test samples to [0,1], where higher values indicate potential OOD samples.

\section{Experiments}


This section presents a comprehensive empirical evaluation, focusing on systematic experiments to assess the ability of VFMs to detect different distributional changes in the challenging AD domain. 
We begin by reviewing the datasets used in our experiments and provide implementation details for the methods benchmarked in this study. Our evaluation examines the detection of semantic shifts, covariate shifts, and their combination. The proposed unsupervised shift detection techniques, based on modeling the distribution of $\mathcal{F}_\mathrm{monitor}$, are systematically compared to state-of-the-art OOD detection methods. Additionally, we evaluate our technique using identical model architectures trained on different datasets to establish comparative baselines. This broad experimental evaluation is complemented with a downstream application: we employ our shift detection techniques  to filter out OOD inputs to semantic segmentation DNNs, and show that increasing filtering strengths notably improves segmentation performance. This suggests that our model-agnostic method specifically captures error-prone OOD samples despite being model-agnostic.

\vspace{-2mm}
\paragraph{Metrics:}
\label{sec:evaluationDetails}

For each test image $x^* \in \mathcal{X} = \mathcal{X}_{\text{ID}} \cup \mathcal{X}_{\text{OOD}}$, we compute a confidence score $c(x^*)$ using one of the models outlined above. Higher values of $c(x^*)$ indicate a greater likelihood that $x^*$ belongs to the ID data represented by the set $\mathcal{X}_\mathrm{ID}$.
To evaluate the model's performance in distinguishing between ID and OOD samples, we consider the following binary classification metrics:

\textbf{Area Under the Receiver Operating Characteristic Curve, AUROC:}
The ROC curve plots true against false positive rate for varying decision thresholds.
An AUROC value of 1 indicates perfect classification, while 0.5 corresponds to random guessing.

\textbf{Area Under the Precision-Recall Curve, AUPR:}
The AUPR 
is particularly useful for performance evaluation on  class-imbalanced test datasets.

\textbf{False Positive Rate (FPR) at 95\% True Positive Rate (TPR), FPR95:} This metric measures the FPR for the threshold at which the TPR is 95\%. Specifically, we determine the threshold $t$ that achieves a 95\% TPR for ID samples and compute the corresponding FPR for OOD samples at $t$. Lower FPR95 values mean fewer OOD samples are misclassified as ID, i.e., better OOD detection performance.

These metrics are measured on $\mathcal{X}_{\text{ID}}\cup\mathcal{X}_{\text{OOD}}$.


\vspace{-2mm}
\paragraph{Datasets:}
\label{sec:datasets}

Our experimental framework employs a diverse array of datasets to evaluate the robustness of our proposed method under various shifts. We carefully selected and curated datasets to simulate semantic shifts, covariate shifts, and their combination, providing a comprehensive testbed for our analysis.

To investigate semantic shifts, we use the Cityscapes train set \cite{cordts2016cityscapes} as our primary $\mathcal{X}_\mathrm{monitor}$, with its validation set serving as $\mathcal{X}_\mathrm{ID}$. We introduced OOD elements using two complementary datasets: the Lost and Found dataset \cite{pinggera2016lost}, which captures real-world road hazards in environmental conditions matching Cityscapes, and the bravo-synobjs dataset \cite{loiseau2023reliability}, which provides synthetic anomalies through generative inpainting of Cityscapes images. This combination enables evaluation of VFMs on both naturally occurring and synthetically generated semantic shifts while maintaining consistent environmental conditions with $\mathcal{X}_\mathrm{monitor}$.

For covariate shifts, we employed datasets that maintain semantic consistency with $\mathcal{X}_\mathrm{monitor}$ while introducing varying visual characteristics. Using the Cityscapes training set \cite{cordts2016cityscapes} as our monitoring dataset $\mathcal{X}_\mathrm{monitor}$, validation set of Cityscapes as $\mathcal{X}_\mathrm{ID}$, we evaluate on several OOD datasets $\mathcal{X}_\mathrm{OOD}$: Foggy Cityscapes \cite{sakaridis2018semantic} with graduated synthetic fog, bravo-synrain and bravo-synflare \cite{loiseau2023reliability} with synthetic rain and lens flare effects. For ACDC, we utilize its clear-weather training images as part of $\mathcal{X}_\mathrm{monitor}$, clear-weather validation images as part of $\mathcal{X}_\mathrm{ID}$ and its adverse-condition validation images (\ie fog, night, rain, snow) as $\mathcal{X}_\mathrm{OOD}$. This setup enables systematic evaluation of VFMs under both synthetic and real-world covariate shifts.

To explore the interplay between semantic and covariate shifts, we incorporated the SegmentMeIfYouCan (SMIYC) dataset \cite{chan2021segmentmeifyoucan} and the Indian Driving Dataset (IDD) \cite{varma2019idd}. These datasets introduce both novel semantic elements and environmental variations, allowing us to assess our method under complex, real-world distributional changes. See Appendix for examples and visualizations. 

\vspace{-2mm}
\paragraph{Pre-trained VFMs and Implementation Details:}
\label{sec:pretrainedModels}

We employ a diverse selection of pre-trained VFMs, including CLIP \cite{radford2021learning}, DINO \cite{caron2021emerging}, DINOv2 \cite{oquab2023dinov2}, and Grounding DINO \cite{liu2023grounding}, each representing a distinct paradigm in visual representation learning.
CLIP \cite{radford2021learning}
is evaluated across multiple architectures. Specifically, we utilize ResNet (RN) variants with varying depths (RN50, RN101) and width multipliers (RN50x4, RN50x16, RN50x64), alongside Vision Transformer (ViT) configurations with different patch sizes and model capacities (ViT-B/32, ViT-B/16, ViT-L/14)\footnote{CLIP: 
\tiny\url{https://github.com/openai/CLIP}}.
DINO is examined through a range of backbones, including ViT variants with diverse patch sizes (Dino-ViT-S/16, Dino-ViT-S/8, Dino-ViT-B/16, Dino-ViT-B/8) and a ResNet variant (Dino-RN50)\footnote{DINO: 
\tiny\url{https://github.com/facebookresearch/dino}}.
DINOv2 \cite{oquab2023dinov2}
is represented by Vision Transformer variants of increasing model capacity (Dinov2-ViT-S/14, Dinov2-ViT-B/14, Dinov2-ViT-L/14, Dinov2-ViT-G/14)\footnote{DINOv2: 
\tiny\url{https://github.com/facebookresearch/dinov2}}.
Grounding DINO \cite{liu2023grounding}
employs a Swin Transformer \cite{liu2021swin} as its image encoder, building upon the DETR-based detection architecture \cite{zhang2022dino}. In our experiments, we include Grounding DINO to evaluate the effectiveness of the Swin architecture \cite{liu2021swin} in comparison to ViT-based models, providing a broader perspective on the performance of different backbone architectures\footnote{Grounding DINO: 
\tiny\url{https://github.com/IDEA-Research/GroundingDINO}}.

We also compare the performance of the VFMs against their ImageNet-trained counterparts \cite{deng2009imagenet}. This baseline includes ResNet variants (RN50, RN101, RN152) \cite{he2016deep}, ViT architectures (ViT-B/16, ViT-L/16) \cite{dosovitskiy2020vit}, and Swin Transformer  backbones (SwinB, SwinT)\footnote{ImageNet-trained ResNets: {\tiny\url{https://github.com/pytorch/pytorch}}; ViT and Swin Transformers: \tiny\url{https://github.com/huggingface/pytorch-image-models}} \cite{liu2021swin}. Furthermore, we trained autoencoder (AE) on the primary $\mathcal{X}_\mathrm{monitor}$ dataset to serve as feature extractors, offering an additional point of comparison with the VFMs. Architectural specifications, comparisons with ImageNet-trained counterparts, and implementation considerations are presented in the Appendix.

\subsection{Semantic Shift}
\label{Semantic_Shift_Experiment}



\cref{tab:semantic_shift} shows the two top-performing backbones of each model for semantic shift detection (find the complete comparison in the Appendix). The results reveal the varying sensitivities of different VFMs and their backbones to input images containing rare semantic classes absent from the training data. Several key observations are discussed in the following.

\begin{table*}[t!]
\vspace*{-1.3\baselineskip}
\centering
\resizebox{\textwidth}{!}{%
\begin{tabular}{cc|c||ccccccccccccccc}
\hline
\multirow{3}{*}{\textbf{Model}}                                                    & \multirow{3}{*}{\textbf{Backbone}} & \multirow{3}{*}{\textbf{\begin{tabular}[c]{@{}c@{}}Latent\\ Space\\ Dim.\end{tabular}}} & \multicolumn{15}{c}{\textbf{Lost and Found}}                                                                                                                                                                                                                                                                                                                                                                                                                                                                                                                                                                                                                                                                                                                                                                                                                           \\ \cline{4-18} 
                                                                                   &                                    &                                                                                         & \multicolumn{3}{c|}{\textbf{APS}}                                                                                                                                    & \multicolumn{3}{c|}{\textbf{MFS}}                                                                                                                                                               & \multicolumn{3}{c|}{\textbf{OC-SVM}}                                                                                                                                 & \multicolumn{3}{c|}{\textbf{GMM}}                                                                                                                                    & \multicolumn{3}{c}{\textbf{NF}}                                                                                                                 \\ \cline{4-18} 
                                                                                   &                                    &                                                                                         & \multicolumn{1}{c|}{\textbf{AUROC}↑} & \multicolumn{1}{c|}{\textbf{AUPR}↑} & \multicolumn{1}{c|}{\textbf{FPR95}↓} & \multicolumn{1}{c|}{\textbf{\textbf{AUROC}↑}} & \multicolumn{1}{c|}{\textbf{\textbf{AUPR}↑}} & \multicolumn{1}{c|}{\textbf{\textbf{FPR95}↓}} & \multicolumn{1}{c|}{\textbf{AUROC}↑} & \multicolumn{1}{c|}{\textbf{AUPR}↑} & \multicolumn{1}{c|}{\textbf{FPR95}↓} & \multicolumn{1}{c|}{\textbf{AUROC}↑} & \multicolumn{1}{c|}{\textbf{AUPR}↑} & \multicolumn{1}{c|}{\textbf{FPR95}↓} & \multicolumn{1}{c|}{\textbf{AUROC}↑} & \multicolumn{1}{c|}{\textbf{AUPR}↑} & \textbf{FPR95}↓ \\ \hline
\multirow{2}{*}{\textbf{CLIP}}                                                     & \textbf{RN50x64}                   & 1024                                                                                    & \textbf{0.94}                                         & \textbf{0.93}                                        & \multicolumn{1}{c|}{\textbf{0.24}}                    & \textbf{0.93}                                                  & \textbf{0.93}                                                 & \multicolumn{1}{c|}{\textbf{0.27}}                             & \textbf{0.94}                                         & \textbf{0.93}                                        & \multicolumn{1}{c|}{\textbf{0.24}}                    & \textbf{0.94}                                         & \textbf{0.93}                                        & \multicolumn{1}{c|}{\textbf{0.23}}                    & 0.88                                                  & 0.84                                                 & 0.37                             \\
                                                                                   & \textbf{ViT-B/16}                  & 512                                                                                     & 0.88                                                  & 0.86                                                 & \multicolumn{1}{c|}{0.38}                             & 0.86                                                           & 0.85                                                          & \multicolumn{1}{c|}{0.41}                                      & 0.89                                                  & 0.88                                                 & \multicolumn{1}{c|}{0.43}                             & 0.91                                                  & 0.89                                                 & \multicolumn{1}{c|}{0.33}                             & 0.91                                                  & 0.88                                                 & 0.34                             \\ \hline
\multirow{2}{*}{\textbf{DINO}}                                                     & \textbf{ViT-B/16}                  & 768                                                                                     & 0.79                                                  & 0.75                                                 & \multicolumn{1}{c|}{0.52}                             & 0.72                                                           & 0.64                                                          & \multicolumn{1}{c|}{0.60}                                      & 0.79                                                  & 0.76                                                 & \multicolumn{1}{c|}{0.59}                             & 0.84                                                  & 0.82                                                 & \multicolumn{1}{c|}{0.55}                             & 0.75                                                  & 0.71                                                 & 0.76                             \\
                                                                                   & \textbf{ViT-B/8}                   & 768                                                                                     & 0.81                                                  & 0.76                                                 & \multicolumn{1}{c|}{0.49}                             & 0.73                                                           & 0.64                                                          & \multicolumn{1}{c|}{0.59}                                      & 0.77                                                  & 0.75                                                 & \multicolumn{1}{c|}{0.56}                             & 0.87                                                  & 0.86                                                 & \multicolumn{1}{c|}{0.45}                             & 0.82                                                  & 0.81                                                 & 0.59                             \\ \hline
\multirow{2}{*}{\textbf{DINOv2}}                                                   & \textbf{ViT-L/14}                  & 1024                                                                                    & 0.89                                                  & 0.87                                                 & \multicolumn{1}{c|}{0.43}                             & 0.91                                                           & 0.88                                                          & \multicolumn{1}{c|}{0.37}                                      & 0.84                                                  & 0.82                                                 & \multicolumn{1}{c|}{0.68}                             & 0.82                                                  & 0.77                                                 & \multicolumn{1}{c|}{0.52}                             & 0.83                                                  & 0.78                                                 & 0.54                             \\
                                                                                   & \textbf{ViT-G/14}                  & 1536                                                                                    & 0.90                                                  & 0.87                                                 & \multicolumn{1}{c|}{0.30}                             & 0.89                                                           & 0.86                                                          & \multicolumn{1}{c|}{0.37}                                      & 0.85                                                  & 0.82                                                 & \multicolumn{1}{c|}{0.63}                             & 0.86                                                  & 0.78                                                 & \multicolumn{1}{c|}{0.39}                             & 0.83                                                  & 0.75                                                 & 0.46                             \\ \hline
\multirow{2}{*}{\textbf{\begin{tabular}[c]{@{}c@{}}Grounding\\ DINO\end{tabular}}} & \textbf{SwinB}                     & 768                                                                                     & 0.88                                                  & 0.84                                                 & \multicolumn{1}{c|}{0.47}                             & 0.82                                                           & 0.76                                                          & \multicolumn{1}{c|}{0.52}                                      & 0.84                                                  & 0.80                                                 & \multicolumn{1}{c|}{0.54}                             & 0.91                                                  & 0.89                                                 & \multicolumn{1}{c|}{0.32}                             & 0.95                                                  & 0.94                                                 & \textbf{0.16}                    \\
                                                                                   & \textbf{SwinT}                     & 768                                                                                     & 0.85                                                  & 0.80                                                 & \multicolumn{1}{c|}{0.53}                             & 0.79                                                           & 0.73                                                          & \multicolumn{1}{c|}{0.60}                                      & 0.83                                                  & 0.79                                                 & \multicolumn{1}{c|}{0.59}                             & 0.92                                                  & 0.90                                                 & \multicolumn{1}{c|}{0.28}                             & \textbf{0.96}                                         & \textbf{0.95}                                        & \textbf{0.16}                    \\ \hline
                                                                                   &                                    &                                                                                         & \multicolumn{15}{c}{\textbf{Bravo-Synobj}}                                                                                                                                                                                                                                                                                                                                                                                                                                                                                                                                                                                                                                                                                                                                                                                                                             \\ \hline
\multirow{2}{*}{\textbf{CLIP}}                                                     & \textbf{RN50x64}                   & 1024                                                                                    & \textbf{0.93}                                         & \textbf{0.94}                                        & \multicolumn{1}{c|}{0.54}                             & \textbf{0.93}                                                  & \textbf{0.94}                                                 & \multicolumn{1}{c|}{0.54}                                      & \textbf{0.93}                                         & \textbf{0.95}                                        & \multicolumn{1}{c|}{\textbf{0.54}}                    & \textbf{0.94}                                         & \textbf{0.95}                                        & \multicolumn{1}{c|}{0.53}                             & 0.94                                                  & 0.96                                                 & 0.43                             \\
                                                                                   & \textbf{ViT-B/16}                  & 512                                                                                     & 0.91                                                  & 0.93                                                 & \multicolumn{1}{c|}{\textbf{0.50}}                    & 0.91                                                           & 0.93                                                          & \multicolumn{1}{c|}{\textbf{0.50}}                             & 0.91                                                  & 0.93                                                 & \multicolumn{1}{c|}{\textbf{0.54}}                    & 0.92                                                  & 0.94                                                 & \multicolumn{1}{c|}{\textbf{0.48}}                    & 0.93                                                  & 0.95                                                 & 0.49                             \\ \hline
\multirow{2}{*}{\textbf{DINO}}                                                     & \textbf{ViT-B/16}                  & 768                                                                                     & 0.81                                                  & 0.82                                                 & \multicolumn{1}{c|}{0.69}                             & 0.81                                                           & 0.82                                                          & \multicolumn{1}{c|}{0.69}                                      & 0.83                                                  & 0.85                                                 & \multicolumn{1}{c|}{0.75}                             & 0.86                                                  & 0.88                                                 & \multicolumn{1}{c|}{0.60}                             & 0.84                                                  & 0.87                                                 & 0.80                             \\
                                                                                   & \textbf{ViT-B/8}                   & 768                                                                                     & 0.82                                                  & 0.84                                                 & \multicolumn{1}{c|}{0.64}                             & 0.82                                                           & 0.84                                                          & \multicolumn{1}{c|}{0.64}                                      & 0.83                                                  & 0.86                                                 & \multicolumn{1}{c|}{0.70}                             & 0.86                                                  & 0.89                                                 & \multicolumn{1}{c|}{0.59}                             & 0.88                                                  & 0.91                                                 & 0.65                             \\ \hline
\multirow{2}{*}{\textbf{DINOv2}}                                                   & \textbf{ViT-L/14}                  & 1024                                                                                    & 0.78                                                  & 0.79                                                 & \multicolumn{1}{c|}{0.77}                             & 0.78                                                           & 0.79                                                          & \multicolumn{1}{c|}{0.77}                                      & 0.78                                                  & 0.81                                                 & \multicolumn{1}{c|}{0.80}                             & 0.82                                                  & 0.85                                                 & \multicolumn{1}{c|}{0.74}                             & 0.86                                                  & 0.88                                                 & 0.63                             \\
                                                                                   & \textbf{ViT-G/14}                  & 1536                                                                                    & 0.79                                                  & 0.80                                                 & \multicolumn{1}{c|}{0.73}                             & 0.79                                                           & 0.80                                                          & \multicolumn{1}{c|}{0.73}                                      & 0.81                                                  & 0.83                                                 & \multicolumn{1}{c|}{0.69}                             & 0.82                                                  & 0.85                                                 & \multicolumn{1}{c|}{0.71}                             & 0.86                                                  & 0.89                                                 & 0.69                             \\ \hline
\multirow{2}{*}{\textbf{\begin{tabular}[c]{@{}c@{}}Grounding\\ DINO\end{tabular}}} & \textbf{SwinB}                     & 768                                                                                     & 0.60                                                  & 0.56                                                 & \multicolumn{1}{c|}{0.83}                             & 0.61                                                           & 0.56                                                          & \multicolumn{1}{c|}{0.82}                                      & 0.58                                                  & 0.54                                                 & \multicolumn{1}{c|}{0.85}                             & 0.76                                                  & 0.68                                                 & \multicolumn{1}{c|}{0.58}                             & \textbf{0.99}                                         & \textbf{0.99}                                        & \textbf{0.02}                    \\
                                                                                   & \textbf{SwinT}                     & 768                                                                                     & 0.57                                                  & 0.54                                                 & \multicolumn{1}{c|}{0.89}                             & 0.58                                                           & 0.54                                                          & \multicolumn{1}{c|}{0.90}                                      & 0.57                                                  & 0.54                                                 & \multicolumn{1}{c|}{0.91}                             & 0.64                                                  & 0.58                                                 & \multicolumn{1}{c|}{0.78}                             & 0.91                                                  & 0.84                                                 & 0.20                             \\ \hline
\end{tabular}
}
\vspace*{-.3\baselineskip}
\caption{Comparison of VFMs with various backbones on the Lost and Found \cite{pinggera2016lost} and Bravo-Synobj\cite{loiseau2023reliability} datasets. focusing on semantic shift sensitivity. Results show AUROC↑. AUPR↑ (higher is better) and FPR95↓ (lower is better) metrics. Best values per metric in bold.}
\label{tab:semantic_shift}
\end{table*}

\begin{table*}[htbp!]
\centering
\huge
\scriptsize
\resizebox{\textwidth}{!}{%
\begin{tabular}{cc|c||cc|cc|cc|cc|cc|cc|cc|cc}
\hline
\multirow{3}{*}{\textbf{Model}} & \multirow{3}{*}{\textbf{Backbone}} & \multicolumn{1}{c||}{\multirow{3}{*}{\textbf{\begin{tabular}[c]{@{}c@{}}Latent\\ Space\\ Dim.\end{tabular}}}} & \multicolumn{2}{c|}{\textbf{\begin{tabular}[c]{@{}c@{}}Foggy \\ Cityscapes\\ (i=0.05)\end{tabular}}} & \multicolumn{2}{c|}{\textbf{\begin{tabular}[c]{@{}c@{}}Foggy \\ Cityscapes\\ (i = 0.2)\end{tabular}}} & \multicolumn{2}{c|}{\textbf{\begin{tabular}[c]{@{}c@{}}Bravo-\\ Synflare\end{tabular}}} & \multicolumn{2}{c|}{\textbf{\begin{tabular}[c]{@{}c@{}}Bravo-\\ Synrain\end{tabular}}} & \multicolumn{2}{c|}{\textbf{ACDC Fog}} & \multicolumn{2}{c|}{\textbf{ACDC Night}} & \multicolumn{2}{c|}{\textbf{ACDC Rain}} & \multicolumn{2}{c}{\textbf{ACDC Snow}} \\ \cline{4-19} 
& & & \multicolumn{2}{c|}{\textbf{FPR95}↓} & \multicolumn{2}{c|}{\textbf{FPR95}↓} & \multicolumn{2}{c|}{\textbf{FPR95}↓} & \multicolumn{2}{c|}{\textbf{FPR95}↓} & \multicolumn{2}{c|}{\textbf{FPR95}↓} & \multicolumn{2}{c|}{\textbf{FPR95}↓} & \multicolumn{2}{c|}{\textbf{FPR95}↓} & \multicolumn{2}{c}{\textbf{FPR95}↓} \\ \cline{4-19} 
& & & \textbf{GMM} & \textbf{NF} & \textbf{GMM} & \textbf{NF} & \textbf{GMM} & \textbf{NF} & \textbf{GMM} & \textbf{NF} & \textbf{GMM} & \textbf{NF} & \textbf{GMM} & \textbf{NF} & \textbf{GMM} & \textbf{NF} & \textbf{GMM} & \textbf{NF} \\ \hline
\multirow{2}{*}{\textbf{CLIP}} & \textbf{RN50x64} & 1024 & 0.76 & 0.67 & 0.26 & 0.21 & 0.62 & 0.47 & 0.01 & 0.04 & \textbf{0.06} & \textbf{0.04} & 0.13 & 0.23 & \textbf{0.26} & \textbf{0.22} & \textbf{0.05} & \textbf{0.03} \\
& \textbf{ViT-B/16} & 512 & \textbf{0.57} & 0.48 & 0.18 & 0.10 & 0.43 & 0.35 & 0.20 & 0.12 & 0.07 & 0.07 & 0.34 & 0.16 & 0.27 & 0.25 & 0.05 & 0.05 \\ \hline
\multirow{2}{*}{\textbf{DINO}} & \textbf{ViT-B/16} & 768 & 0.72 & 0.74 & 0.31 & 0.34 & 0.37 & 0.57 & 0.17 & 0.03 & 0.43 & 0.50 & 0.33 & 0.23 & 0.53 & 0.68 & 0.43 & 0.48 \\
& \textbf{ViT-B/8} & 768 & 0.72 & 0.52 & 0.35 & 0.09 & 0.51 & 0.33 & 0.10 & \textbf{0} & 0.35 & 0.35 & 0.48 & 0.27 & 0.52 & 0.58 & 0.38 & 0.34 \\ \hline
\multirow{2}{*}{\textbf{DINOv2}} & \textbf{ViT-L/14} & 1024 & 0.92 & 0.88 & 0.84 & 0.70 & 0.88 & 0.80 & 0.56 & 0.37 & 0.69 & 0.59 & 0.48 & 0.39 & 0.83 & 0.67 & 0.62 & 0.38 \\
& \textbf{ViT-G/14} & 1536 & 0.89 & 0.83 & 0.73 & 0.54 & 0.88 & 0.80 & 0.61 & 0.35 & 0.63 & 0.46 & 0.31 & 0.31 & 0.71 & 0.58 & 0.49 & 0.28 \\ \hline
\multirow{2}{*}{\textbf{\begin{tabular}[c]{@{}c@{}}Grounding \\ DINO\end{tabular}}} & \textbf{SwinB} & 768 & 0.79 & \textbf{0.36} & 0.17 & \textbf{0} & 0.26 & \textbf{0.05} & \textbf{0} & \textbf{0} & 0.16 & 0.08 & 0.10 & \textbf{0.02} & 0.60 & 0.35 & 0.24 & 0.04 \\
& \textbf{SwinT} & 768 & 0.72 & 0.42 & \textbf{0.10} & 0.02 & \textbf{0.25} & 0.07 & \textbf{0} & \textbf{0} & 0.08 & 0.13 & \textbf{0.04} & 0.08 & 0.52 & 0.34 & 0.24 & 0.07 \\ \hline
\end{tabular}}
\vspace*{-.3\baselineskip}
\caption{Comparison of VFMs with various backbones on covariate shift scenarios using GMM and NF methods. Results show FPR95↓ (lower is better) for different environmental conditions: synthetic fog with varying intensities (Foggy Cityscapes \cite{sakaridis2018semantic}), synthetic effects (Bravo-Synflare\cite{loiseau2023reliability}, Bravo-Synrain\cite{liu2023grounding}), and real adverse conditions (ACDC\cite{sakaridis2021acdc}). Best values for each dataset are highlighted in bold.}
\label{tab:covariate_shift}
\vspace*{-.5\baselineskip}
\end{table*}

\textbf{Architecture-Specific Distribution Modeling:} Our analysis reveals that Grounding DINO with Swin Transformer achieves notable performance when paired with flow-based density estimation (Lost and Found: FPR95: 0.16; Bravo-Synobj: FPR95: 0.02). The hierarchical design of Swin Transformer, with its local-global attention mechanism, captures richer semantic relationships than standard architectures. However, \finding{leveraging this enhanced semantic understanding requires appropriate distribution modeling}—while simpler approaches show limited effectiveness with Swin Transformer (APS, MFS, OC-SVM: FPR95: 0.47--0.54), normalizing flows unlock its full potential. This architecture-specific behavior demonstrates the importance of matching distribution modeling techniques to architectural characteristics.

\textbf{Real vs.\ Synthetic Performance:} Notably, \finding{CLIP's RN50x64 maintains consistent performance with conventional methods like GMM and OC-SVM on real-world data} (FPR95: 0.23--0.24 on Lost and Found), but struggles with synthetic anomalies (FPR95: 0.53--0.54 on Bravo-Synobj). In contrast, \finding{Grounding DINO with flow-based modeling maintains effectiveness across both real and synthetic scenarios}, suggesting superior generalization to diverse semantic shift types regardless of data origin.


\subsection{Covariate Shift}
\label{sec:covariateShift}

As previously, due to space limitations we present results for only the two top-performing backbones per model in terms of FPR95 scores, focusing on GMM and NF methods due to their superior detection capabilities and FPR95's sensitivity in distinguishing architectural performance. A comprehensive comparison is available in the Appendix.
The following key observations can be drawn from our experiments.



\textbf{Perturbation Intensity:} Performance generally improves with increased perturbation intensity, as evidenced by the Foggy Cityscapes \cite{sakaridis2018semantic} results (i=0.05 vs.\ i=0.2), indicating sensitivity to the magnitude of covariate shift.

\textbf{Architecture and Detection Performance:} Grounding DINO with Swin Transformer as highly effective in detecting most covariate shifts when paired with normalizing flows, achieving perfect detection (FPR95: 0) on several conditions. However, its performance exhibits an interesting pattern in rain detection: perfect detection on Bravo-Synrain, which features explicit rain drops on the camera lens, but degraded performance on ACDC Rain (FPR95: 0.34-0.35), which only captures environmental rain effects (see example images in Appendix). Similarly, CLIP's RN50x64 shows higher false positive rates on synthetic perturbations (fog: 0.67, flare: 0.47) but excels in real adverse conditions (ACDC Snow: 0.03), suggesting that architectural designs may be more sensitive to direct camera-level perturbations than to subtle environmental changes. 

\begin{table}[t!]
\centering
\small
\resizebox{\columnwidth}{!}{%
\begin{tabular}{>{\centering\arraybackslash}p{1.5cm}>{\centering\arraybackslash}p{1.5cm}||cc|cc|}
\hline
\multirow{3}{*}{\textbf{Model}} & \multirow{3}{*}{\textbf{Backbone}} & \multicolumn{2}{c|}{\textbf{SegmentMeIfYouCan}} & \multicolumn{2}{c|}{\textbf{Indian Driving Dataset}} \\
\cline{3-6}
& & \multicolumn{2}{c|}{\textbf{FPR95↓}} & \multicolumn{2}{c|}{\textbf{FPR95↓}} \\
\cline{3-6}
& & \textbf{GMM} & \textbf{NF} & \textbf{GMM} & \textbf{NF} \\
\hline
\multirow{2}{*}{\centering\textbf{CLIP}} & \textbf{RN50x64} & \textbf{0} & \textbf{0} & \textbf{0} & \textbf{0} \\
& \textbf{ViT-B/16} & \textbf{0} & \textbf{0} & \textbf{0} & \textbf{0} \\
\hline
\multirow{2}{*}{\centering\textbf{DINO}} & \textbf{ViT-B/16} & 0.01 & 0.21 & 0.26 & 0.26 \\
& \textbf{ViT-B/8} & \textbf{0} & \textbf{0} & 0.12 & 0.07 \\
\hline
\multirow{2}{*}{\centering\textbf{DINOv2}} & \textbf{ViT-L/14} & \textbf{0} & 0.01 & \textbf{0} & \textbf{0} \\
& \textbf{ViT-g/14} & \textbf{0} & \textbf{0} & \textbf{0} & \textbf{0} \\
\hline
\multirow{2}{*}{\centering\textbf{\begin{tabular}[c]{@{}c@{}}Grounding \\ DINO\end{tabular}}} & \textbf{SwinB} & \textbf{0} & \textbf{0} & 0.01 & \textbf{0} \\
& \textbf{SwinT} & \textbf{0} & \textbf{0} & 0.05 &  \textbf{0} \\
\hline
\end{tabular}
}
\caption{Sensitivity of VFMs on datasets combining semantic and covariate shifts. Results are shown for the SMIYC dataset \cite{chan2021segmentmeifyoucan} (obstacle track) and IDD \cite{varma2019idd}, using the two top-performing two backbones for each model. Highest scores are highlighted in bold.}
\label{tab:semanticadncovariate}
\vspace*{-\baselineskip}
\end{table}

\subsection{Semantic and Covariate Shift}
\label{sec:SemanticcovariateShift}
To evaluate the robustness of VFMs to simultaneous semantic and covariate shifts, we utilized the SMIYC dataset \cite{chan2021segmentmeifyoucan} and the Indian Driving Dataset (IDD) \cite{varma2019idd}. SMIYC introduces both novel objects and adverse weather conditions, while IDD presents a geographically distinct urban environment with unique objects and weather patterns. SMIYC contains two datasets, RoadAnomaly21 and RoadObstacle21, the latter of which was used for results in \cref{tab:semanticadncovariate}. As in previous experiments, Cityscapes \cite{cordts2016cityscapes} training images served as the reference, with validation images as ID samples. \cref{tab:semanticadncovariate} shows GMM results for each the two top-performing backbones, the full comparison can be found in the Appendix.

\finding{VFMs demonstrate strong robustness to both semantic and covariate shifts, with most models achieving near-perfect FPR95 scores} on both the SMIYC and Indian Driving Dataset. This is to be expected as both shifts at a time constitute a stronger deviation from the training data than previously. 
\subsection{Benchmarking Against OOD Methods and ImageNet-Trained Counterparts}
\label{sec:ComparativeBaselineOtherOOd}
We evaluated our approach against both established OOD detection methods \cite{azeem2024monitizer} and ImageNet-trained counterparts of the same architectures in the context of input monitoring for AD systems. To facilitate comparison, we implemented these methods using a DeepLabV3+ semantic segmentation network with ResNet50 backbone\footnote{DeepLabv3+-RN50 trained on Cityscapes:: 
\tiny\url{https://github.com/open-mmlab/mmsegmentation/tree/main/configs/deeplabv3plus}} trained on Cityscapes \cite{cordts2016cityscapes} dataset. Our VFM-based input monitoring technique was benchmarked against methods categorized by their detection mechanisms: feature-based approaches (GradNorm \cite{huang2021importance}), which analyze network activations and gradients; logit-based methods (Energy \cite{liu2020energy}, Entropy \cite{macedo2021entropic}, MaxLogit \cite{zhang2023decoupling}), which examine output characteristics; confidence-based methods (DICE \cite{sun2022dice}), which leverage confidence scores derived from the model; distance-based approaches (KNN \cite{sun2022out}), which assess feature space similarities; and ODIN \cite{liang2017enhancing}, which enhances OOD detection through temperature scaling and input preprocessing. These methods require specific network access: intermediate layers for feature-based, final layer for logit-based, and feature embeddings for distance-based approaches. In contrast, our VFM-based method operates independently of the monitored network, significantly enhancing its applicability in AD systems where network architectures may be proprietary or inaccessible.

For our experimental evaluation, we used Cityscapes \cite{cordts2016cityscapes} training set as $\mathcal{X}\mathrm{monitor}$ and its validation set as $\mathcal{X}_\mathrm{ID}$, testing across diverse OOD scenarios: real-world obstacles (Lost and Found \cite{pinggera2016lost}), synthetic anomalies (Bravo-SynObj \cite{loiseau2023reliability}), adverse weather conditions (Foggy Cityscapes \cite{sakaridis2018semantic}, Bravo-SynRain \cite{loiseau2023reliability}). For fair comparison, all methods were evaluated using ResNet50 backbones from both the segmentation network and CLIP, with our approach utilizing GMM and NF for distribution modeling. Additionally, we report results using Grounding DINO's SwinB backbone with NF, which achieved the best performance overall. The results are shown in \cref{tab:OODmethodbenchmark}.

\begin{table*}[t]
\tiny
\centering
\resizebox{1\textwidth}{!}{
\begin{tabular}{c|l|cc|cc|cc|cc}
\hline
\multirow{2}{*}{} & \multirow{2}{*}{\textbf{Method}} & 
\multicolumn{2}{c|}{\textbf{Lost and Found}} & 
\multicolumn{2}{c|}{\textbf{Bravo-Synobj}} & 
\multicolumn{2}{c|}{\textbf{Foggy Cityscapes (i=0.2)}} & 
\multicolumn{2}{c}{\textbf{Bravo-Synrain}} \\ \cline{3-10} 
& & \textbf{AUROC}↑ & \textbf{FPR95}↓ & \textbf{AUROC}↑ & \textbf{FPR95}↓ & \textbf{AUROC}↑ & \textbf{FPR95}↓ & \textbf{AUROC}↑ & \textbf{FPR95}↓ \\ \hline
\multirow{7}{*}{\textbf{\begin{tabular}[c]{@{}c@{}}OOD\\Methods\end{tabular}}} 
& \textbf{DICE}~\cite{sun2022dice} & 0.50 & 0.94 & 0.57 & 0.91 & 0.69 & 0.92 & 0.77 & 0.80 \\
& \textbf{Energy}~\cite{liu2020energy} & 0.50 & 0.97 & 0.50 & 0.97 & 0.49 & 0.99 & 0.50 & 0.99 \\
& \textbf{Entropy}~\cite{macedo2021entropic} & 0.60 & 0.88 & 0.61 & 0.86 & 0.64 & 0.90 & 0.70 & 0.86 \\
& \textbf{GradNorm}~\cite{huang2021importance} & 0.80 & 0.43 & 0.45 & 0.94 & 0.98 & 0.06 & 0.99 & \textbf{0.00} \\
& \textbf{KNN}~\cite{sun2022out} & 0.68 & 0.78 & 0.49 & 0.91 & 0.57 & 0.88 & 0.42 & 0.94 \\
& \textbf{MaxLogit}~\cite{zhang2023decoupling} & 0.56 & 0.60 & 0.56 & 0.95 & 0.63 & 0.95 & 0.62 & 0.97 \\
& \textbf{ODIN}~\cite{liang2017enhancing} & 0.56 & 0.94 & 0.56 & 0.95 & 0.59 & 0.93 & 0.58 & 0.97 \\ \hline
\multirow{3}{*}{\textbf{\begin{tabular}[c]{@{}c@{}}Image-Net\\Trained\\Counterparts\end{tabular}}} 
& \textbf{RN50-GMM} & 0.76 & 0.62 & 0.84 & 0.75 & 0.92 & 0.28 & 0.94 & 0.23 \\
& \textbf{RN50-NF} & 0.75 & 0.63 & 0.87 & 0.74 & 0.93 & 0.25 & 0.97 & 0.12 \\
& \textbf{SwinB-NF} & 0.86 & 0.44 & 0.91 & 0.53 & 0.92 & 0.35 & 0.99 & 0.05 \\ \hline
\multirow{3}{*}{\textbf{\begin{tabular}[c]{@{}c@{}}VFMs\end{tabular}}}
& \textbf{Ours (RN50-GMM)} & 0.90 & 0.36 & 0.89 & 0.66 & 0.85 & 0.47 & 0.85 & 0.43 \\
& \textbf{Ours (RN50-NF)} & 0.82 & 0.52 & 0.92 & 0.57 & 0.86 & 0.53 & 0.95 & 0.16 \\
& \textbf{Ours (SwinB-NF)} & \textbf{0.95} & \textbf{0.16} & \textbf{0.99} & \textbf{0.02} & \textbf{1.00} & \textbf{0.00} & \textbf{1.00} & 0.02 \\ \hline
\end{tabular}
}
\vspace*{-\baselineskip}
\caption{Comparison of OOD detection between baseline methods and our VFM-based approach across semantic shifts (Lost and Found, Bravo-Synobj) and covariate shifts (Foggy Cityscapes, Bravo-Synrain). All baselines use a DeepLabV3+ network with ResNet50 backbone. Our approach is evaluated with CLIP's RN50 using GMM and NF for distribution modeling. Best results per dataset in bold.}
\label{tab:OODmethodbenchmark}
\vspace*{-\baselineskip}
\end{table*}

OOD detection methods show varying performance across semantic and covariate shifts. In particular, confidence-based methods (DICE \cite{sun2022dice}) and distance-based methods (KNN \cite{sun2022out}) demonstrate moderate performance on Lost and Found (AUROC: 0.50-0.68) and Bravo-Synobj (AUROC: 0.49-0.57), suggesting their limited capability in capturing complex semantic shifts. This limitation likely arises from their reliance on model confidence scores (DICE) and feature similarity measures (KNN), which are often unreliable under high semantic variations.

Logit-based approaches (Energy \cite{liu2020energy}, Entropy \cite{macedo2021entropic}, MaxLogit \cite{zhang2023decoupling}) perform consistently but modestly across all datasets (AUROC: 0.50-0.70), indicating their relative robustness but limited discriminative power. Their uniform performance suggests that while these methods leverage logit magnitudes to separate ID and OOD samples, they may struggle with fine-grained anomalies that do not cause strong logit deviations.

Notably, feature-based methods like GradNorm \cite{huang2021importance} show strong performance on covariate shifts (Foggy Cityscapes AUROC: 0.98, FPR95: 0.06; Bravo-Synrain AUROC: 0.99, FPR95: 0.00) and perform reasonably well on real semantic shifts (Lost and Found AUROC: 0.80, FPR95: 0.43) but struggle significantly with synthetic anomalies (Bravo-Synobj AUROC: 0.45, FPR95: 0.94), likely because GradNorm is more sensitive to global image transformations and natural distortions, while synthetic object insertions do not trigger strong gradient deviations.

ImageNet-trained models perform well on covariate shifts but lag behind VFMs on semantic shifts. Using identical architectures (ResNet50, SwinB) trained solely on ImageNet with the same distribution modeling techniques, we found competitive performance on covariate shifts. SwinB-NF excelled on weather conditions (Bravo-Synrain: AUROC: 0.99, FPR95: 0.05) and RN50-NF performed strongly on fog detection (Foggy Cityscapes: AUROC: 0.93, FPR95: 0.25). However, a significant gap appears for semantic shifts, particularly with SwinB (ImageNet SwinB-NF on Bravo-Synobj: FPR95: 0.53 vs. VFM SwinB-NF: FPR95: 0.02). This suggests ImageNet pre-training captures features sufficient for environmental variations, while VFMs' diverse training objectives and broader data exposure significantly enhance detection of novel objects and semantic anomalies.


\subsection{Input Monitoring}
\label{sec:InputMonitoringSemanticSegmentation}

To evaluate the practical applicability of our input monitoring approach, we applied it to semantic segmentation using DeepLabv3+ \cite{chen2018encoder} with ResNet-50, ResNet-101 \cite{he2016deep}, and MobileNetV3 \cite{howard2019searching} backbones. These models were trained on Cityscapes \cite{cordts2016cityscapes} and evaluated on IDD \cite{varma2019idd}.  We employed Grounding DINO's SwinB backbone with normalizing flows to model the distribution of Cityscapes training images. For each test image in IDD, we computed its log-likelihood under this learned distribution, providing a measure of similarity to the training distribution. 
To systematically evaluate the effect of input monitoring, we established baselines using the entire IDD test set, then applied increasingly selective filtering based on the computed log-likelihoods. Specifically, we defined three filtering thresholds that retain the top 75\% (low), 50\% (medium), and 25\% (high) of IDD images ranked by their log-likelihood scores.
Segmentation performance was evaluated using mean Intersection over Union (mIoU) and mean Accuracy (mAcc) for each filtering level.

\begin{table}[t]
\centering
\resizebox{\columnwidth}{!}{%
\begin{tabular}{lcccccc}
\hline
\multirow{2}{*}{\textbf{\begin{tabular}[c]{@{}l@{}}Filter\\Strength\end{tabular}}} & \multicolumn{2}{c}{\textbf{ResNet50}} & \multicolumn{2}{c}{\textbf{ResNet101}} & \multicolumn{2}{c}{\textbf{MobileNetv3}} \\
\textbf{} & \textbf{mIoU↑} & \textbf{mAcc↑} & \textbf{mIoU↑} & \textbf{mAcc↑} & \textbf{mIoU↑} & \textbf{mAcc↑} \\
\hline
None          & 43.58 & 62.84 & 45.85  & 65.1  & 34.22 & 52.25 \\
Low           & 44.73 & 63.62 & 46.81 & 65.81 & 35.4 & 53.32  \\
Medium        & 46.86 & 64.82 & 48.35  & 66.54 & 36.59 & 54.11 \\
High          & 49.38 & 65.19 & 50.52 & 66.42 & 38.36  & 54.02 \\
\hline
\end{tabular}}
\vspace*{-.3\baselineskip}
\caption{Semantic segmentation performance of DeepLabv3+ with ResNet-50, ResNet-101, and MobileNetV3 backbones on IDD 
for different input monitoring filter strengths.}
\label{tab:semanticinputmonitoring}
\vspace*{-\baselineskip}
\end{table}

Results are summarized in \cref{tab:semanticinputmonitoring}. \finding{Our input monitoring approach clearly enhances semantic segmentation performance on OOD data}. Utilizing Grounding DINO's \cite{liu2023grounding} SwinB backbone with NF, the method quantifies distributional shifts between test and training sets, enabling adaptive input filtering. The results reveal a clear trade-off: stricter monitoring yields fewer retained images but clear accuracy gains. Stronger performance improvements can be expected on even more heterogeneous datasets, containing, e.g., Indian and German driving scenes. 
\section{Conclusion}





We presented an unsupervised, model-agnostic binary OOD classification framework that leverages latent representations of pre-trained off-the-shelf VFMs together with density modeling for joint detection of both semantic and covariate shifts. This is for the first time thoroughly tested in the challenging domain of AD. 
Our extensive experiments show that simple techniques for modeling the VFM feature distributions are sufficient to detect semantic and/or covariate OOD shifts with astonishingly high accuracy. VFMs clearly outperform non-foundation model latent representations, and classical OOD methods. 
Lastly, despite our method being model-agnostic, down-stream perception modules exhibit clear performance improvements when filtering out data that deviates too much from the training distribution. This promises that VMF latent space distributions can help to identify inputs with higher risk of inducing errors in computer vision DNNs, even in complex open-world domains.

\section*{Acknowledgment}
We acknowledge the financial support by the German Federal Ministry for Economic Affairs and Climate Action (BMWK) within the project nxtAIM – Next Generation AI Methods and the German Federal Ministry of Education and Research (BMBF) within the project REFRAME (grant no. 01IS24073).

{
    \small
    \bibliographystyle{ieeenat_fullname}
    \bibliography{main}

\begin{thebibliography}{91}
\providecommand{\natexlab}[1]{#1}
\providecommand{\url}[1]{\texttt{#1}}
\expandafter\ifx\csname urlstyle\endcsname\relax
  \providecommand{\doi}[1]{doi: #1}\else
  \providecommand{\doi}{doi: \begingroup \urlstyle{rm}\Url}\fi

\bibitem[Alam et~al.(2020)Alam, Sonbhadra, Agarwal, and Nagabhushan]{alam2020one}
Shamshe Alam, Sanjay~Kumar Sonbhadra, Sonali Agarwal, and P Nagabhushan.
\newblock One-class support vector classifiers: A survey.
\newblock \emph{Knowledge-Based Systems}, 196:\penalty0 105754, 2020.

\bibitem[Awais et~al.(2025)Awais, Naseer, Khan, Anwer, Cholakkal, Shah, Yang, and Khan]{awais2025foundation}
Muhammad Awais, Muzammal Naseer, Salman Khan, Rao~Muhammad Anwer, Hisham Cholakkal, Mubarak Shah, Ming-Hsuan Yang, and Fahad~Shahbaz Khan.
\newblock Foundation models defining a new era in vision: a survey and outlook.
\newblock \emph{IEEE Transactions on Pattern Analysis and Machine Intelligence}, 2025.

\bibitem[Azeem et~al.(2024)Azeem, Grobelna, Kanav, K{\v{r}}et{\'\i}nsk{\`y}, Mohr, and Rieder]{azeem2024monitizer}
Muqsit Azeem, Marta Grobelna, Sudeep Kanav, Jan K{\v{r}}et{\'\i}nsk{\`y}, Stefanie Mohr, and Sabine Rieder.
\newblock Monitizer: Automating design and evaluation of neural network monitors.
\newblock In \emph{International Conference on Computer Aided Verification}, pages 265--279. Springer, 2024.

\bibitem[Bevandi{\'c} et~al.(2019)Bevandi{\'c}, Kre{\v{s}}o, Or{\v{s}}i{\'c}, and {\v{S}}egvi{\'c}]{bevandic2019simultaneous}
Petra Bevandi{\'c}, Ivan Kre{\v{s}}o, Marin Or{\v{s}}i{\'c}, and Sini{\v{s}}a {\v{S}}egvi{\'c}.
\newblock Simultaneous semantic segmentation and outlier detection in presence of domain shift.
\newblock In \emph{Pattern Recognition: 41st DAGM German Conference, DAGM GCPR 2019, Dortmund, Germany, September 10--13, 2019, Proceedings 41}, pages 33--47. Springer, 2019.

\bibitem[Bevandi{\'c} et~al.(2021)Bevandi{\'c}, Kre{\v{s}}o, Or{\v{s}}i{\'c}, and {\v{S}}egvi{\'c}]{bevandic2021dense}
Petra Bevandi{\'c}, Ivan Kre{\v{s}}o, Marin Or{\v{s}}i{\'c}, and Sini{\v{s}}a {\v{S}}egvi{\'c}.
\newblock Dense outlier detection and open-set recognition based on training with noisy negative images.
\newblock \emph{arXiv preprint arXiv:2101.09193}, 2021.

\bibitem[Bolte et~al.(2019)Bolte, Kamp, Breuer, Homoceanu, Schlicht, Huger, Lipinski, and Fingscheidt]{bolte2019unsupervised}
Jan-Aike Bolte, Markus Kamp, Antonia Breuer, Silviu Homoceanu, Peter Schlicht, Fabian Huger, Daniel Lipinski, and Tim Fingscheidt.
\newblock Unsupervised domain adaptation to improve image segmentation quality both in the source and target domain.
\newblock In \emph{proceedings of the IEEE/CVF conference on computer vision and pattern recognition workshops}, pages 0--0, 2019.

\bibitem[Bommasani et~al.(2021)Bommasani, Hudson, Adeli, Altman, Arora, von Arx, Bernstein, Bohg, Bosselut, Brunskill, et~al.]{bommasani2021opportunities}
Rishi Bommasani, Drew~A Hudson, Ehsan Adeli, Russ Altman, Simran Arora, Sydney von Arx, Michael~S Bernstein, Jeannette Bohg, Antoine Bosselut, Emma Brunskill, et~al.
\newblock On the opportunities and risks of foundation models.
\newblock \emph{arXiv preprint arXiv:2108.07258}, 2021.

\bibitem[Bounsiar and Madden(2014)]{bounsiar2014one}
Abdenour Bounsiar and Michael~G Madden.
\newblock One-class support vector machines revisited.
\newblock In \emph{2014 International Conference on Information Science \& Applications (ICISA)}, pages 1--4. IEEE, 2014.

\bibitem[{CarADAS}(2024)]{CarADAS_LKA}
{CarADAS}.
\newblock {Understanding ADAS: Lane Keep Assist}, 2024.
\newblock Accessed: 4 March 2025.

\bibitem[Caron et~al.(2021)Caron, Touvron, Misra, J{\'e}gou, Mairal, Bojanowski, and Joulin]{caron2021emerging}
Mathilde Caron, Hugo Touvron, Ishan Misra, Herv{\'e} J{\'e}gou, Julien Mairal, Piotr Bojanowski, and Armand Joulin.
\newblock Emerging properties in self-supervised vision transformers.
\newblock In \emph{Proceedings of the IEEE/CVF international conference on computer vision}, pages 9650--9660, 2021.

\bibitem[Chan et~al.(2021{\natexlab{a}})Chan, Lis, Uhlemeyer, Blum, Honari, Siegwart, Fua, Salzmann, and Rottmann]{chan2021segmentmeifyoucan}
Robin Chan, Krzysztof Lis, Svenja Uhlemeyer, Hermann Blum, Sina Honari, Roland Siegwart, Pascal Fua, Mathieu Salzmann, and Matthias Rottmann.
\newblock Segmentmeifyoucan: A benchmark for anomaly segmentation.
\newblock \emph{arXiv preprint arXiv:2104.14812}, 2021{\natexlab{a}}.

\bibitem[Chan et~al.(2021{\natexlab{b}})Chan, Rottmann, and Gottschalk]{chan2021entropy}
Robin Chan, Matthias Rottmann, and Hanno Gottschalk.
\newblock Entropy maximization and meta classification for out-of-distribution detection in semantic segmentation.
\newblock In \emph{Proceedings of the ieee/cvf international conference on computer vision}, pages 5128--5137, 2021{\natexlab{b}}.

\bibitem[Chen et~al.(2017)Chen, Papandreou, Kokkinos, Murphy, and Yuille]{chen2017deeplab}
Liang-Chieh Chen, George Papandreou, Iasonas Kokkinos, Kevin Murphy, and Alan~L Yuille.
\newblock Deeplab: Semantic image segmentation with deep convolutional nets, atrous convolution, and fully connected crfs.
\newblock \emph{IEEE transactions on pattern analysis and machine intelligence}, 40\penalty0 (4):\penalty0 834--848, 2017.

\bibitem[Chen et~al.(2018)Chen, Zhu, Papandreou, Schroff, and Adam]{chen2018encoder}
Liang-Chieh Chen, Yukun Zhu, George Papandreou, Florian Schroff, and Hartwig Adam.
\newblock Encoder-decoder with atrous separable convolution for semantic image segmentation.
\newblock In \emph{Proceedings of the European conference on computer vision (ECCV)}, pages 801--818, 2018.

\bibitem[Cheng et~al.(2024)Cheng, Song, Ge, Liu, Wang, and Shan]{cheng2024yolo}
Tianheng Cheng, Lin Song, Yixiao Ge, Wenyu Liu, Xinggang Wang, and Ying Shan.
\newblock Yolo-world: Real-time open-vocabulary object detection.
\newblock In \emph{Proceedings of the IEEE/CVF Conference on Computer Vision and Pattern Recognition}, pages 16901--16911, 2024.

\bibitem[Cordts et~al.(2016)Cordts, Omran, Ramos, Rehfeld, Enzweiler, Benenson, Franke, Roth, and Schiele]{cordts2016cityscapes}
Marius Cordts, Mohamed Omran, Sebastian Ramos, Timo Rehfeld, Markus Enzweiler, Rodrigo Benenson, Uwe Franke, Stefan Roth, and Bernt Schiele.
\newblock The cityscapes dataset for semantic urban scene understanding.
\newblock In \emph{Proceedings of the IEEE conference on computer vision and pattern recognition}, pages 3213--3223, 2016.

\bibitem[{Council of the European Union}(2024)]{euaiactcompromise2024}
{Council of the European Union}.
\newblock Proposal for a regulation of the european parliament and of the council laying down harmonised rules on artificial intelligence (artificial intelligence act) and amending certain union legislative acts - analysis of the final compromise text with a view to agreement.
\newblock \url{https://data.consilium.europa.eu/doc/document/ST-5662-2024-INIT/en/pdf}, 2024.
\newblock Accessed: 2024-03-23.

\bibitem[Deng et~al.(2009)Deng, Dong, Socher, Li, Li, and Fei-Fei]{deng2009imagenet}
Jia Deng, Wei Dong, Richard Socher, Li-Jia Li, Kai Li, and Li Fei-Fei.
\newblock Imagenet: A large-scale hierarchical image database.
\newblock In \emph{2009 IEEE conference on computer vision and pattern recognition}, pages 248--255. Ieee, 2009.

\bibitem[Dinh et~al.(2016)Dinh, Sohl-Dickstein, and Bengio]{dinh2016density}
Laurent Dinh, Jascha Sohl-Dickstein, and Samy Bengio.
\newblock Density estimation using real nvp.
\newblock \emph{arXiv preprint arXiv:1605.08803}, 2016.

\bibitem[Dosovitskiy et~al.(2021)Dosovitskiy, Beyer, Kolesnikov, Weissenborn, Zhai, Unterthiner, Dehghani, Minderer, Heigold, Gelly, Uszkoreit, and Houlsby]{dosovitskiy2020vit}
Alexey Dosovitskiy, Lucas Beyer, Alexander Kolesnikov, Dirk Weissenborn, Xiaohua Zhai, Thomas Unterthiner, Mostafa Dehghani, Matthias Minderer, Georg Heigold, Sylvain Gelly, Jakob Uszkoreit, and Neil Houlsby.
\newblock An image is worth 16x16 words: Transformers for image recognition at scale.
\newblock \emph{ICLR}, 2021.

\bibitem[Gal and Ghahramani(2016)]{gal2016dropout}
Yarin Gal and Zoubin Ghahramani.
\newblock Dropout as a bayesian approximation: Representing model uncertainty in deep learning.
\newblock In \emph{international conference on machine learning}, pages 1050--1059. PMLR, 2016.

\bibitem[Gao et~al.(2024)Gao, Li, Salzmann, and He]{gao2024generalize}
Zhitong Gao, Bingnan Li, Mathieu Salzmann, and Xuming He.
\newblock Generalize or detect? towards robust semantic segmentation under multiple distribution shifts.
\newblock \emph{arXiv preprint arXiv:2411.03829}, 2024.

\bibitem[Grci{\'c} et~al.(2022)Grci{\'c}, Bevandi{\'c}, and {\v{S}}egvi{\'c}]{grcic2022densehybrid}
Matej Grci{\'c}, Petra Bevandi{\'c}, and Sini{\v{s}}a {\v{S}}egvi{\'c}.
\newblock Densehybrid: Hybrid anomaly detection for dense open-set recognition.
\newblock In \emph{European Conference on Computer Vision}, pages 500--517. Springer, 2022.

\bibitem[Grci{\'c} et~al.(2023)Grci{\'c}, {\v{S}}ari{\'c}, and {\v{S}}egvi{\'c}]{grcic2023advantages}
Matej Grci{\'c}, Josip {\v{S}}ari{\'c}, and Sini{\v{s}}a {\v{S}}egvi{\'c}.
\newblock On advantages of mask-level recognition for outlier-aware segmentation.
\newblock In \emph{Proceedings of the IEEE/CVF Conference on Computer Vision and Pattern Recognition}, pages 2937--2947, 2023.

\bibitem[Gupta et~al.(2024)Gupta, Chakraborty, and Bansal]{gupta2024detecting}
Aryaman Gupta, Kaustav Chakraborty, and Somil Bansal.
\newblock Detecting and mitigating system-level anomalies of vision-based controllers.
\newblock In \emph{2024 IEEE International Conference on Robotics and Automation (ICRA)}, pages 9953--9959. IEEE, 2024.

\bibitem[He et~al.(2016)He, Zhang, Ren, and Sun]{he2016deep}
Kaiming He, Xiangyu Zhang, Shaoqing Ren, and Jian Sun.
\newblock Deep residual learning for image recognition.
\newblock In \emph{Proceedings of the IEEE conference on computer vision and pattern recognition}, pages 770--778, 2016.

\bibitem[He et~al.(2017)He, Gkioxari, Doll{\'a}r, and Girshick]{he2017mask}
Kaiming He, Georgia Gkioxari, Piotr Doll{\'a}r, and Ross Girshick.
\newblock Mask r-cnn.
\newblock In \emph{Proceedings of the IEEE international conference on computer vision}, pages 2961--2969, 2017.

\bibitem[Heidecker et~al.(2021)Heidecker, Hannan, Bieshaar, and Sick]{heidecker2021towards}
Florian Heidecker, Abdul Hannan, Maarten Bieshaar, and Bernhard Sick.
\newblock Towards corner case detection by modeling the uncertainty of instance segmentation networks.
\newblock In \emph{Pattern Recognition. ICPR International Workshops and Challenges: Virtual Event, January 10--15, 2021, Proceedings, Part IV}, pages 361--374. Springer, 2021.

\bibitem[Hell et~al.(2021)Hell, Hinz, Liu, Goyal, Pei, Lytvynenko, Knoll, and Yiqiang]{hell2021monitoring}
Franz Hell, Gereon Hinz, Feng Liu, Sakshi Goyal, Ke Pei, Tetiana Lytvynenko, Alois Knoll, and Chen Yiqiang.
\newblock Monitoring perception reliability in autonomous driving: Distributional shift detection for estimating the impact of input data on prediction accuracy.
\newblock In \emph{Proceedings of the 5th ACM Computer Science in Cars Symposium}, pages 1--9, 2021.

\bibitem[Hendrycks and Gimpel(2016)]{hendrycks2016baseline}
Dan Hendrycks and Kevin Gimpel.
\newblock A baseline for detecting misclassified and out-of-distribution examples in neural networks.
\newblock \emph{arXiv preprint arXiv:1610.02136}, 2016.

\bibitem[Howard et~al.(2019)Howard, Sandler, Chu, Chen, Chen, Tan, Wang, Zhu, Pang, Vasudevan, et~al.]{howard2019searching}
Andrew Howard, Mark Sandler, Grace Chu, Liang-Chieh Chen, Bo Chen, Mingxing Tan, Weijun Wang, Yukun Zhu, Ruoming Pang, Vijay Vasudevan, et~al.
\newblock Searching for mobilenetv3.
\newblock In \emph{Proceedings of the IEEE/CVF international conference on computer vision}, pages 1314--1324, 2019.

\bibitem[Huang et~al.(2021)Huang, Geng, and Li]{huang2021importance}
Rui Huang, Andrew Geng, and Yixuan Li.
\newblock On the importance of gradients for detecting distributional shifts in the wild.
\newblock \emph{Advances in Neural Information Processing Systems}, 34:\penalty0 677--689, 2021.

\bibitem[Ilyas et~al.(2024)Ilyas, Freeman, and Rottmann]{ilyas2024potential}
Sadia Ilyas, Ido Freeman, and Matthias Rottmann.
\newblock On the potential of open-vocabulary models for object detection in unusual street scenes.
\newblock \emph{arXiv preprint arXiv:2408.11221}, 2024.

\bibitem[{International Organization for Standardization}(2024)]{ISO8800}
{International Organization for Standardization}.
\newblock {ISO/PAS 8800:2024 – Road Vehicles – Safety and Artificial Intelligence}, 2024.
\newblock Accessed: 4 March 2025.

\bibitem[Kamath et~al.(2021)Kamath, Singh, LeCun, Synnaeve, Misra, and Carion]{kamath2021mdetr}
Aishwarya Kamath, Mannat Singh, Yann LeCun, Gabriel Synnaeve, Ishan Misra, and Nicolas Carion.
\newblock Mdetr-modulated detection for end-to-end multi-modal understanding.
\newblock In \emph{Proceedings of the IEEE/CVF international conference on computer vision}, pages 1780--1790, 2021.

\bibitem[Kassab et~al.(2024)Kassab, Mattamala, Zhang, and Fallon]{kassab2024language}
Christina Kassab, Matias Mattamala, Lintong Zhang, and Maurice Fallon.
\newblock Language-extended indoor slam (lexis): A versatile system for real-time visual scene understanding.
\newblock In \emph{2024 IEEE International Conference on Robotics and Automation (ICRA)}, pages 15988--15994. IEEE, 2024.

\bibitem[Kendall and Gal(2017)]{kendall2017uncertainties}
Alex Kendall and Yarin Gal.
\newblock What uncertainties do we need in bayesian deep learning for computer vision?
\newblock \emph{Advances in neural information processing systems}, 30, 2017.

\bibitem[Kendall et~al.(2015)Kendall, Badrinarayanan, and Cipolla]{kendall2015bayesian}
Alex Kendall, Vijay Badrinarayanan, and Roberto Cipolla.
\newblock Bayesian segnet: Model uncertainty in deep convolutional encoder-decoder architectures for scene understanding.
\newblock \emph{arXiv preprint arXiv:1511.02680}, 2015.

\bibitem[Krasin et~al.(2017)Krasin, Duerig, Alldrin, Ferrari, Abu-El-Haija, Kuznetsova, Rom, Uijlings, Popov, Veit, et~al.]{krasin2017openimages}
Ivan Krasin, Tom Duerig, Neil Alldrin, Vittorio Ferrari, Sami Abu-El-Haija, Alina Kuznetsova, Hassan Rom, Jasper Uijlings, Stefan Popov, Andreas Veit, et~al.
\newblock Openimages: A public dataset for large-scale multi-label and multi-class image classification.
\newblock \emph{Dataset available from https://github. com/openimages}, 2\penalty0 (3):\penalty0 18, 2017.

\bibitem[Krishna et~al.(2017)Krishna, Zhu, Groth, Johnson, Hata, Kravitz, Chen, Kalantidis, Li, Shamma, et~al.]{krishna2017visual}
Ranjay Krishna, Yuke Zhu, Oliver Groth, Justin Johnson, Kenji Hata, Joshua Kravitz, Stephanie Chen, Yannis Kalantidis, Li-Jia Li, David~A Shamma, et~al.
\newblock Visual genome: Connecting language and vision using crowdsourced dense image annotations.
\newblock \emph{International journal of computer vision}, 123:\penalty0 32--73, 2017.

\bibitem[Li et~al.(2022)Li, Zhang, Zhou, Shi, and Luo]{li2022out}
Ruoqi Li, Chongyang Zhang, Hao Zhou, Chao Shi, and Yan Luo.
\newblock Out-of-distribution identification: Let detector tell which i am not sure.
\newblock In \emph{European Conference on Computer Vision}, pages 638--654. Springer, 2022.

\bibitem[Liang et~al.(2017)Liang, Li, and Srikant]{liang2017enhancing}
Shiyu Liang, Yixuan Li, and Rayadurgam Srikant.
\newblock Enhancing the reliability of out-of-distribution image detection in neural networks.
\newblock \emph{arXiv preprint arXiv:1706.02690}, 2017.

\bibitem[Lin et~al.(2014)Lin, Maire, Belongie, Hays, Perona, Ramanan, Doll{\'a}r, and Zitnick]{lin2014microsoft}
Tsung-Yi Lin, Michael Maire, Serge Belongie, James Hays, Pietro Perona, Deva Ramanan, Piotr Doll{\'a}r, and C~Lawrence Zitnick.
\newblock Microsoft coco: Common objects in context.
\newblock In \emph{Computer Vision--ECCV 2014: 13th European Conference, Zurich, Switzerland, September 6-12, 2014, Proceedings, Part V 13}, pages 740--755. Springer, 2014.

\bibitem[Liu et~al.(2023{\natexlab{a}})Liu, Li, Wu, and Lee]{liu2023visual}
Haotian Liu, Chunyuan Li, Qingyang Wu, and Yong~Jae Lee.
\newblock Visual instruction tuning.
\newblock \emph{Advances in neural information processing systems}, 36:\penalty0 34892--34916, 2023{\natexlab{a}}.

\bibitem[Liu et~al.(2023{\natexlab{b}})Liu, Zeng, Ren, Li, Zhang, Yang, Li, Yang, Su, Zhu, et~al.]{liu2023grounding}
Shilong Liu, Zhaoyang Zeng, Tianhe Ren, Feng Li, Hao Zhang, Jie Yang, Chunyuan Li, Jianwei Yang, Hang Su, Jun Zhu, et~al.
\newblock Grounding dino: Marrying dino with grounded pre-training for open-set object detection.
\newblock \emph{arXiv preprint arXiv:2303.05499}, 2023{\natexlab{b}}.

\bibitem[Liu et~al.(2020)Liu, Wang, Owens, and Li]{liu2020energy}
Weitang Liu, Xiaoyun Wang, John Owens, and Yixuan Li.
\newblock Energy-based out-of-distribution detection.
\newblock \emph{Advances in neural information processing systems}, 33:\penalty0 21464--21475, 2020.

\bibitem[Liu et~al.(2023{\natexlab{c}})Liu, Ding, Tian, Pang, Belagiannis, Reid, and Carneiro]{liu2023residual}
Yuyuan Liu, Choubo Ding, Yu Tian, Guansong Pang, Vasileios Belagiannis, Ian Reid, and Gustavo Carneiro.
\newblock Residual pattern learning for pixel-wise out-of-distribution detection in semantic segmentation.
\newblock In \emph{Proceedings of the IEEE/CVF International Conference on Computer Vision}, pages 1151--1161, 2023{\natexlab{c}}.

\bibitem[Liu et~al.(2021)Liu, Lin, Cao, Hu, Wei, Zhang, Lin, and Guo]{liu2021swin}
Ze Liu, Yutong Lin, Yue Cao, Han Hu, Yixuan Wei, Zheng Zhang, Stephen Lin, and Baining Guo.
\newblock Swin transformer: Hierarchical vision transformer using shifted windows.
\newblock In \emph{Proceedings of the IEEE/CVF international conference on computer vision}, pages 10012--10022, 2021.

\bibitem[Lohdefink et~al.(2020)Lohdefink, Fehrling, Klingner, Huger, Schlicht, Schmidt, and Fingscheidt]{lohdefink2020self}
Jonas Lohdefink, Justin Fehrling, Marvin Klingner, Fabian Huger, Peter Schlicht, Nico~M Schmidt, and Tim Fingscheidt.
\newblock Self-supervised domain mismatch estimation for autonomous perception.
\newblock In \emph{Proceedings of the IEEE/CVF Conference on Computer Vision and Pattern Recognition Workshops}, pages 334--335, 2020.

\bibitem[Loiseau et~al.(2023)Loiseau, Vu, Chen, P{\'e}rez, and Cord]{loiseau2023reliability}
Thibaut Loiseau, Tuan-Hung Vu, Mickael Chen, Patrick P{\'e}rez, and Matthieu Cord.
\newblock Reliability in semantic segmentation: Can we use synthetic data?
\newblock \emph{arXiv preprint arXiv:2312.09231}, 2023.

\bibitem[Mac{\^e}do et~al.(2021)Mac{\^e}do, Ren, Zanchettin, Oliveira, and Ludermir]{macedo2021entropic}
David Mac{\^e}do, Tsang~Ing Ren, Cleber Zanchettin, Adriano~LI Oliveira, and Teresa Ludermir.
\newblock Entropic out-of-distribution detection.
\newblock In \emph{2021 international joint conference on neural networks (IJCNN)}, pages 1--8. IEEE, 2021.

\bibitem[Murphy(2021)]{Murphy2021}
Kevin~P. Murphy.
\newblock \emph{Machine Learning: A Probabilistic Perspective}.
\newblock MIT Press, Cambridge, MA, 2 edition, 2021.

\bibitem[Nayal et~al.(2023)Nayal, Yavuz, Henriques, and G{\"u}ney]{nayal2023rba}
Nazir Nayal, Misra Yavuz, Joao~F Henriques, and Fatma G{\"u}ney.
\newblock Rba: Segmenting unknown regions rejected by all.
\newblock In \emph{Proceedings of the IEEE/CVF International Conference on Computer Vision}, pages 711--722, 2023.

\bibitem[Nekrasov et~al.(2024)Nekrasov, Zhou, Ackermann, Hermans, Leibe, and Rottmann]{nekrasov2024oodis}
Alexey Nekrasov, Rui Zhou, Miriam Ackermann, Alexander Hermans, Bastian Leibe, and Matthias Rottmann.
\newblock Oodis: Anomaly instance segmentation benchmark.
\newblock \emph{arXiv preprint arXiv:2406.11835}, 2024.

\bibitem[Ohgushi et~al.(2020)Ohgushi, Horiguchi, and Yamanaka]{ohgushi2020road}
Toshiaki Ohgushi, Kenji Horiguchi, and Masao Yamanaka.
\newblock Road obstacle detection method based on an autoencoder with semantic segmentation.
\newblock In \emph{proceedings of the Asian conference on computer vision}, 2020.

\bibitem[Oquab et~al.(2023)Oquab, Darcet, Moutakanni, Vo, Szafraniec, Khalidov, Fernandez, Haziza, Massa, El-Nouby, et~al.]{oquab2023dinov2}
Maxime Oquab, Timoth{\'e}e Darcet, Th{\'e}o Moutakanni, Huy Vo, Marc Szafraniec, Vasil Khalidov, Pierre Fernandez, Daniel Haziza, Francisco Massa, Alaaeldin El-Nouby, et~al.
\newblock Dinov2: Learning robust visual features without supervision.
\newblock \emph{arXiv preprint arXiv:2304.07193}, 2023.

\bibitem[Papamakarios et~al.(2021)Papamakarios, Nalisnick, Rezende, Mohamed, and Lakshminarayanan]{papamakarios2021normalizing}
George Papamakarios, Eric Nalisnick, Danilo~Jimenez Rezende, Shakir Mohamed, and Balaji Lakshminarayanan.
\newblock Normalizing flows for probabilistic modeling and inference.
\newblock \emph{Journal of Machine Learning Research}, 22\penalty0 (57):\penalty0 1--64, 2021.

\bibitem[Pi et~al.(2024)Pi, Yao, Gao, Zhang, and Zhang]{pi2024perceptiongpt}
Renjie Pi, Lewei Yao, Jiahui Gao, Jipeng Zhang, and Tong Zhang.
\newblock Perceptiongpt: Effectively fusing visual perception into llm.
\newblock In \emph{Proceedings of the IEEE/CVF Conference on Computer Vision and Pattern Recognition}, pages 27124--27133, 2024.

\bibitem[Pinggera et~al.(2016)Pinggera, Ramos, Gehrig, Franke, Rother, and Mester]{pinggera2016lost}
Peter Pinggera, Sebastian Ramos, Stefan Gehrig, Uwe Franke, Carsten Rother, and Rudolf Mester.
\newblock Lost and found: detecting small road hazards for self-driving vehicles.
\newblock In \emph{2016 IEEE/RSJ International Conference on Intelligent Robots and Systems (IROS)}, pages 1099--1106. IEEE, 2016.

\bibitem[Plummer et~al.(2015)Plummer, Wang, Cervantes, Caicedo, Hockenmaier, and Lazebnik]{plummer2015flickr30k}
Bryan~A Plummer, Liwei Wang, Chris~M Cervantes, Juan~C Caicedo, Julia Hockenmaier, and Svetlana Lazebnik.
\newblock Flickr30k entities: Collecting region-to-phrase correspondences for richer image-to-sentence models.
\newblock In \emph{Proceedings of the IEEE international conference on computer vision}, pages 2641--2649, 2015.

\bibitem[Radford et~al.(2021)Radford, Kim, Hallacy, Ramesh, Goh, Agarwal, Sastry, Askell, Mishkin, Clark, et~al.]{radford2021learning}
Alec Radford, Jong~Wook Kim, Chris Hallacy, Aditya Ramesh, Gabriel Goh, Sandhini Agarwal, Girish Sastry, Amanda Askell, Pamela Mishkin, Jack Clark, et~al.
\newblock Learning transferable visual models from natural language supervision.
\newblock In \emph{International conference on machine learning}, pages 8748--8763. PMLR, 2021.

\bibitem[Rahman et~al.(2021)Rahman, Corke, and Dayoub]{rahman2021run}
Quazi~Marufur Rahman, Peter Corke, and Feras Dayoub.
\newblock Run-time monitoring of machine learning for robotic perception: A survey of emerging trends.
\newblock \emph{IEEE Access}, 9:\penalty0 20067--20075, 2021.

\bibitem[Rai et~al.(2024)Rai, Cermelli, Caputo, and Masone]{rai2024mask2anomaly}
Shyam~Nandan Rai, Fabio Cermelli, Barbara Caputo, and Carlo Masone.
\newblock Mask2anomaly: Mask transformer for universal open-set segmentation.
\newblock \emph{IEEE Transactions on Pattern Analysis and Machine Intelligence}, 2024.

\bibitem[Ren et~al.(2024)Ren, Jiang, Liu, Zeng, Liu, Gao, Huang, Ma, Jiang, Chen, et~al.]{ren2024grounding}
Tianhe Ren, Qing Jiang, Shilong Liu, Zhaoyang Zeng, Wenlong Liu, Han Gao, Hongjie Huang, Zhengyu Ma, Xiaoke Jiang, Yihao Chen, et~al.
\newblock Grounding dino 1.5: Advance the" edge" of open-set object detection.
\newblock \emph{arXiv preprint arXiv:2405.10300}, 2024.

\bibitem[Rottmann et~al.(2020)Rottmann, Colling, Hack, Chan, H{\"u}ger, Schlicht, and Gottschalk]{rottmann2020prediction}
Matthias Rottmann, Pascal Colling, Thomas~Paul Hack, Robin Chan, Fabian H{\"u}ger, Peter Schlicht, and Hanno Gottschalk.
\newblock Prediction error meta classification in semantic segmentation: Detection via aggregated dispersion measures of softmax probabilities.
\newblock In \emph{2020 International Joint Conference on Neural Networks (IJCNN)}, pages 1--9. IEEE, 2020.

\bibitem[Russakovsky et~al.(2015)Russakovsky, Deng, Su, Krause, Satheesh, Ma, Huang, Karpathy, Khosla, Bernstein, et~al.]{russakovsky2015imagenet}
Olga Russakovsky, Jia Deng, Hao Su, Jonathan Krause, Sanjeev Satheesh, Sean Ma, Zhiheng Huang, Andrej Karpathy, Aditya Khosla, Michael Bernstein, et~al.
\newblock Imagenet large scale visual recognition challenge.
\newblock \emph{International journal of computer vision}, 115:\penalty0 211--252, 2015.

\bibitem[{SAE International}(2021)]{SAEJ3016}
{SAE International}.
\newblock {Taxonomy and Definitions for Terms Related to Driving Automation Systems for On-Road Motor Vehicles (SAE J3016)}, 2021.
\newblock Accessed: 4 March 2025.

\bibitem[Sakaridis et~al.(2018)Sakaridis, Dai, and Van~Gool]{sakaridis2018semantic}
Christos Sakaridis, Dengxin Dai, and Luc Van~Gool.
\newblock Semantic foggy scene understanding with synthetic data.
\newblock \emph{International Journal of Computer Vision}, 126:\penalty0 973--992, 2018.

\bibitem[Sakaridis et~al.(2021)Sakaridis, Dai, and Van~Gool]{sakaridis2021acdc}
Christos Sakaridis, Dengxin Dai, and Luc Van~Gool.
\newblock Acdc: The adverse conditions dataset with correspondences for semantic driving scene understanding.
\newblock In \emph{Proceedings of the IEEE/CVF International Conference on Computer Vision}, pages 10765--10775, 2021.

\bibitem[Schneider et~al.(2020)Schneider, Rusak, Eck, Bringmann, Brendel, and Bethge]{schneider2020improving}
Steffen Schneider, Evgenia Rusak, Luisa Eck, Oliver Bringmann, Wieland Brendel, and Matthias Bethge.
\newblock Improving robustness against common corruptions by covariate shift adaptation.
\newblock \emph{Advances in neural information processing systems}, 33:\penalty0 11539--11551, 2020.

\bibitem[Sehwag et~al.(2021)Sehwag, Chiang, and Mittal]{sehwag2021ssd}
Vikash Sehwag, Mung Chiang, and Prateek Mittal.
\newblock Ssd: A unified framework for self-supervised outlier detection.
\newblock \emph{arXiv preprint arXiv:2103.12051}, 2021.

\bibitem[Shao et~al.(2019)Shao, Li, Zhang, Peng, Yu, Zhang, Li, and Sun]{shao2019objects365}
Shuai Shao, Zeming Li, Tianyuan Zhang, Chao Peng, Gang Yu, Xiangyu Zhang, Jing Li, and Jian Sun.
\newblock Objects365: A large-scale, high-quality dataset for object detection.
\newblock In \emph{Proceedings of the IEEE/CVF international conference on computer vision}, pages 8430--8439, 2019.

\bibitem[Shi(2024)]{shi2024transnext}
Dai Shi.
\newblock Transnext: Robust foveal visual perception for vision transformers.
\newblock In \emph{Proceedings of the IEEE/CVF Conference on Computer Vision and Pattern Recognition}, pages 17773--17783, 2024.

\bibitem[Stocco et~al.(2020)Stocco, Weiss, Calzana, and Tonella]{stocco2020misbehaviour}
Andrea Stocco, Michael Weiss, Marco Calzana, and Paolo Tonella.
\newblock Misbehaviour prediction for autonomous driving systems.
\newblock In \emph{Proceedings of the ACM/IEEE 42nd international conference on software engineering}, pages 359--371, 2020.

\bibitem[Sun and Li(2022)]{sun2022dice}
Yiyou Sun and Yixuan Li.
\newblock Dice: Leveraging sparsification for out-of-distribution detection.
\newblock In \emph{European Conference on Computer Vision}, pages 691--708. Springer, 2022.

\bibitem[Sun et~al.(2022)Sun, Ming, Zhu, and Li]{sun2022out}
Yiyou Sun, Yifei Ming, Xiaojin Zhu, and Yixuan Li.
\newblock Out-of-distribution detection with deep nearest neighbors.
\newblock In \emph{International Conference on Machine Learning}, pages 20827--20840. PMLR, 2022.

\bibitem[Varma et~al.(2019)Varma, Subramanian, Namboodiri, Chandraker, and Jawahar]{varma2019idd}
Girish Varma, Anbumani Subramanian, Anoop Namboodiri, Manmohan Chandraker, and CV Jawahar.
\newblock Idd: A dataset for exploring problems of autonomous navigation in unconstrained environments.
\newblock In \emph{2019 IEEE winter conference on applications of computer vision (WACV)}, pages 1743--1751. IEEE, 2019.

\bibitem[Voj{\'\i}{\v{r}} and Matas(2023)]{vojivr2023image}
Tom{\'a}{\v{s}} Voj{\'\i}{\v{r}} and Ji{\v{r}}{\'\i} Matas.
\newblock Image-consistent detection of road anomalies as unpredictable patches.
\newblock In \emph{Proceedings of the IEEE/CVF Winter Conference on Applications of Computer Vision}, pages 5491--5500, 2023.

\bibitem[Vojir et~al.(2021)Vojir, {\v{S}}ipka, Aljundi, Chumerin, Reino, and Matas]{vojir2021road}
Tomas Vojir, Tom{\'a}{\v{s}} {\v{S}}ipka, Rahaf Aljundi, Nikolay Chumerin, Daniel~Olmeda Reino, and Jiri Matas.
\newblock Road anomaly detection by partial image reconstruction with segmentation coupling.
\newblock In \emph{Proceedings of the IEEE/CVF International Conference on Computer Vision}, pages 15651--15660, 2021.

\bibitem[Wang et~al.(2023)Wang, Dai, Chen, Huang, Li, Zhu, Hu, Lu, Lu, Li, et~al.]{wang2023internimage}
Wenhai Wang, Jifeng Dai, Zhe Chen, Zhenhang Huang, Zhiqi Li, Xizhou Zhu, Xiaowei Hu, Tong Lu, Lewei Lu, Hongsheng Li, et~al.
\newblock Internimage: Exploring large-scale vision foundation models with deformable convolutions.
\newblock In \emph{Proceedings of the IEEE/CVF conference on computer vision and pattern recognition}, pages 14408--14419, 2023.

\bibitem[Wilson et~al.(2023)Wilson, Fischer, Dayoub, Miller, and S{\"u}nderhauf]{wilson2023safe}
Samuel Wilson, Tobias Fischer, Feras Dayoub, Dimity Miller, and Niko S{\"u}nderhauf.
\newblock Safe: Sensitivity-aware features for out-of-distribution object detection.
\newblock In \emph{Proceedings of the IEEE/CVF International Conference on Computer Vision}, pages 23565--23576, 2023.

\bibitem[Wu and Deng(2023)]{wu2023discriminating}
Aming Wu and Cheng Deng.
\newblock Discriminating known from unknown objects via structure-enhanced recurrent variational autoencoder.
\newblock In \emph{Proceedings of the IEEE/CVF Conference on Computer Vision and Pattern Recognition}, pages 23956--23965, 2023.

\bibitem[Wu et~al.(2024)Wu, Gao, Gao, Yu, Chu, Yu, Gong, Chang, Tseng, Chen, et~al.]{wu2024prospective}
Jianhua Wu, Bingzhao Gao, Jincheng Gao, Jianhao Yu, Hongqing Chu, Qiankun Yu, Xun Gong, Yi Chang, H~Eric Tseng, Hong Chen, et~al.
\newblock Prospective role of foundation models in advancing autonomous vehicles.
\newblock \emph{Research}, 7:\penalty0 0399, 2024.

\bibitem[Xiao et~al.(2024)Xiao, Wu, Xu, Dai, Hu, Lu, Zeng, Liu, and Yuan]{xiao2024florence}
Bin Xiao, Haiping Wu, Weijian Xu, Xiyang Dai, Houdong Hu, Yumao Lu, Michael Zeng, Ce Liu, and Lu Yuan.
\newblock Florence-2: Advancing a unified representation for a variety of vision tasks.
\newblock In \emph{Proceedings of the IEEE/CVF Conference on Computer Vision and Pattern Recognition}, pages 4818--4829, 2024.

\bibitem[Xiao et~al.(2020)Xiao, Yan, and Amit]{xiao2020likelihood}
Zhisheng Xiao, Qing Yan, and Yali Amit.
\newblock Likelihood regret: An out-of-distribution detection score for variational auto-encoder.
\newblock \emph{Advances in neural information processing systems}, 33:\penalty0 20685--20696, 2020.

\bibitem[Zhang et~al.(2022)Zhang, Li, Liu, Zhang, Su, Zhu, Ni, and Shum]{zhang2022dino}
H Zhang, F Li, S Liu, L Zhang, H Su, J Zhu, LM Ni, and HY Shum.
\newblock Dino: Detr with improved denoising anchor boxes for end-to-end object detection. arxiv 2022.
\newblock \emph{arXiv preprint arXiv:2203.03605}, 5, 2022.

\bibitem[Zhang et~al.(2024)Zhang, Li, Qi, Yang, and Ahuja]{zhang2024csl}
Hao Zhang, Fang Li, Lu Qi, Ming-Hsuan Yang, and Narendra Ahuja.
\newblock Csl: Class-agnostic structure-constrained learning for segmentation including the unseen.
\newblock In \emph{Proceedings of the AAAI Conference on Artificial Intelligence}, pages 7078--7086, 2024.

\bibitem[Zhang et~al.(2018)Zhang, Zhang, Zhang, Liu, and Khurshid]{zhang2018deeproad}
Mengshi Zhang, Yuqun Zhang, Lingming Zhang, Cong Liu, and Sarfraz Khurshid.
\newblock Deeproad: Gan-based metamorphic testing and input validation framework for autonomous driving systems.
\newblock In \emph{Proceedings of the 33rd ACM/IEEE International Conference on Automated Software Engineering}, pages 132--142, 2018.

\bibitem[Zhang and Xiang(2023)]{zhang2023decoupling}
Zihan Zhang and Xiang Xiang.
\newblock Decoupling maxlogit for out-of-distribution detection.
\newblock In \emph{Proceedings of the IEEE/CVF Conference on Computer Vision and Pattern Recognition}, pages 3388--3397, 2023.

\bibitem[Zhao et~al.(2024)Zhao, Lv, Xu, Wei, Wang, Dang, Liu, and Chen]{zhao2024detrs}
Yian Zhao, Wenyu Lv, Shangliang Xu, Jinman Wei, Guanzhong Wang, Qingqing Dang, Yi Liu, and Jie Chen.
\newblock Detrs beat yolos on real-time object detection.
\newblock In \emph{Proceedings of the IEEE/CVF Conference on Computer Vision and Pattern Recognition}, pages 16965--16974, 2024.

\bibitem[Zhou et~al.(2024)Zhou, Liu, Yurtsever, Zagar, Zimmer, Cao, and Knoll]{zhou2024vision}
Xingcheng Zhou, Mingyu Liu, Ekim Yurtsever, Bare~Luka Zagar, Walter Zimmer, Hu Cao, and Alois~C Knoll.
\newblock Vision language models in autonomous driving: A survey and outlook.
\newblock \emph{IEEE Transactions on Intelligent Vehicles}, 2024.

\end{thebibliography}
}


\clearpage
\appendix
\setcounter{page}{1}
\maketitlesupplementary

This supplementary material provides comprehensive analyses and additional details that complement the main paper. Concretely, details are provided on:
\begin{itemize}
    \item \ref{sec:additional_results_discussion}: reasoning for the \textbf{scope}
    \item \ref{sec:models_and_datasets}:
    model and \textbf{dataset properties}
    with particular emphasis on both semantic and covariate distribution shift coverage, supported by illustrative examples.
    \item \ref{sec:Implementation Details}: \textbf{implementation considerations} essential for reproducing our experimental framework. 
    \item \ref{sec:Detailed Experimental Results}: additional tables with complete experimental \textbf{results across all model architectures and backbones}
    \item  \ref{sec:Limitations_and_Future_Directions}: More detailed discussion of \textbf{limitations and future directions}.
\end{itemize}    

\section{Discussion of the Scope}
\label{sec:additional_results_discussion} 
This section further argues specific choices of scope:
\begin{itemize}
    \item \ref{sec:input_vs_pixel_monitoring}: \textbf{Why input monitoring} (Comparison with pixel-wise OOD segmentation)
    \item \ref{sec:ComparativeBaselineAnalysis}: \textbf{Why VFMs} instead of standard feature extractors
    \item \ref{sec:realtime_capability} \textbf{Why real-time capability} may reasonably be achieved
\end{itemize}

\subsection{Input Monitoring and Pixel-Level OOD Segmentation}
\label{sec:input_vs_pixel_monitoring}

\begin{figure}[ht!]
    \centering

    \begin{minipage}{\linewidth}
        \centering
        \begin{subfigure}{0.47\linewidth}
            \centering
            \includegraphics[width=\linewidth]{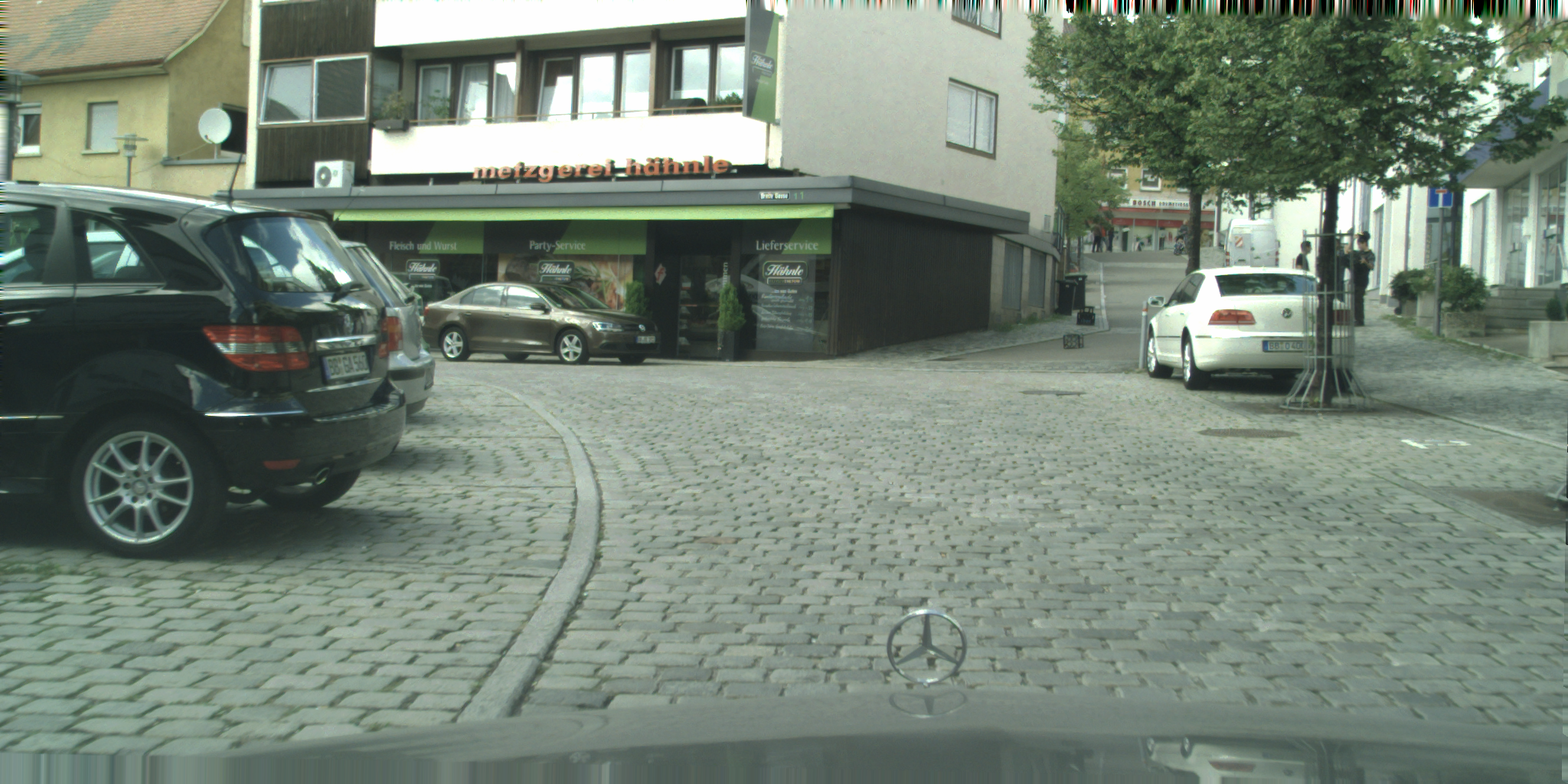}

        \end{subfigure}
        \hfill
        \begin{subfigure}{0.47\linewidth}
            \centering
            \includegraphics[width=\linewidth]{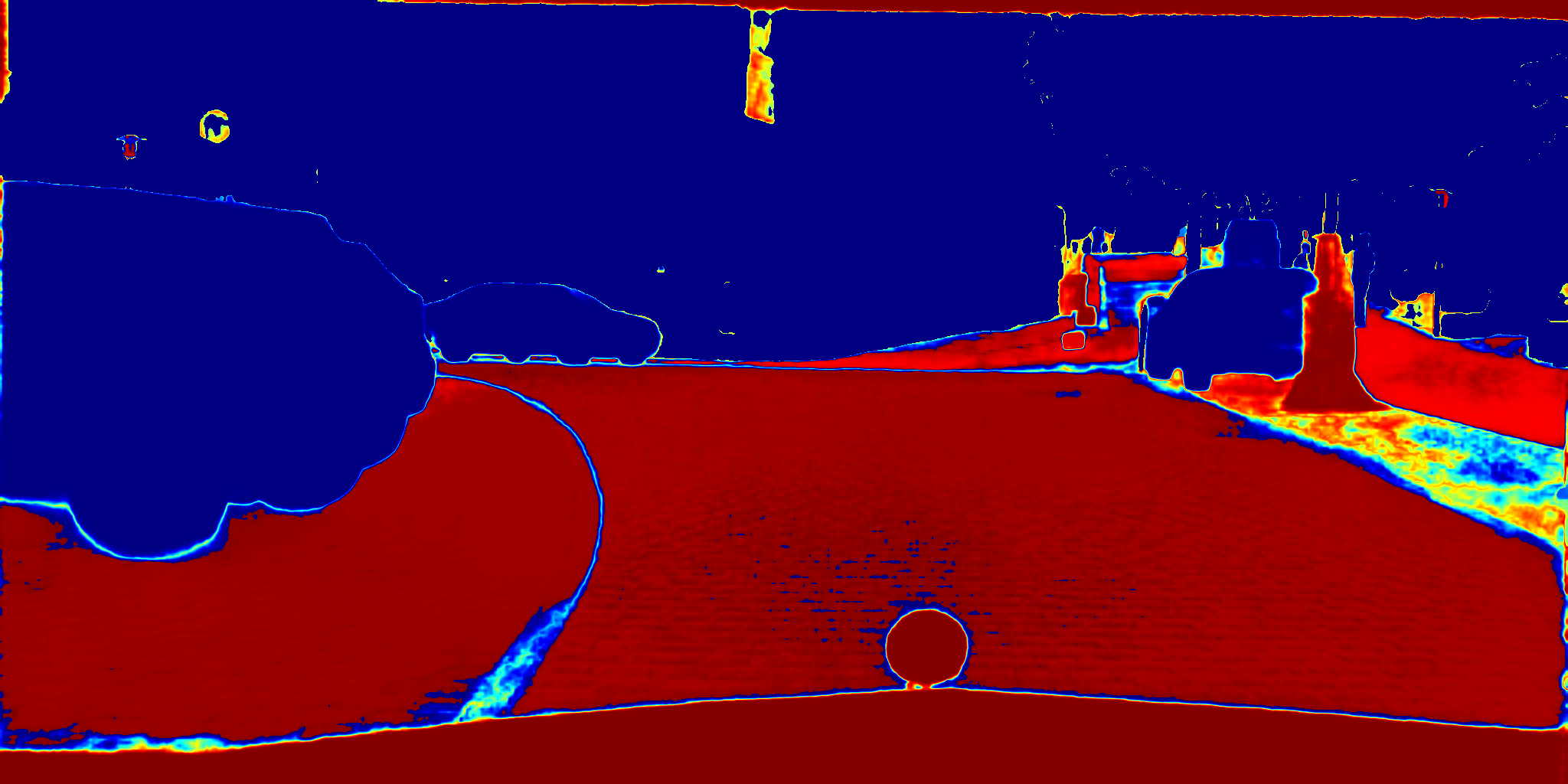}

        \end{subfigure}
        \vspace{0.5cm}
        \centering
        \small \textbf{Lost and Found Dataset} 
    \end{minipage}

    \begin{minipage}{\linewidth}
        \centering
        \begin{subfigure}{0.47\linewidth}
            \centering
            \includegraphics[width=\linewidth]{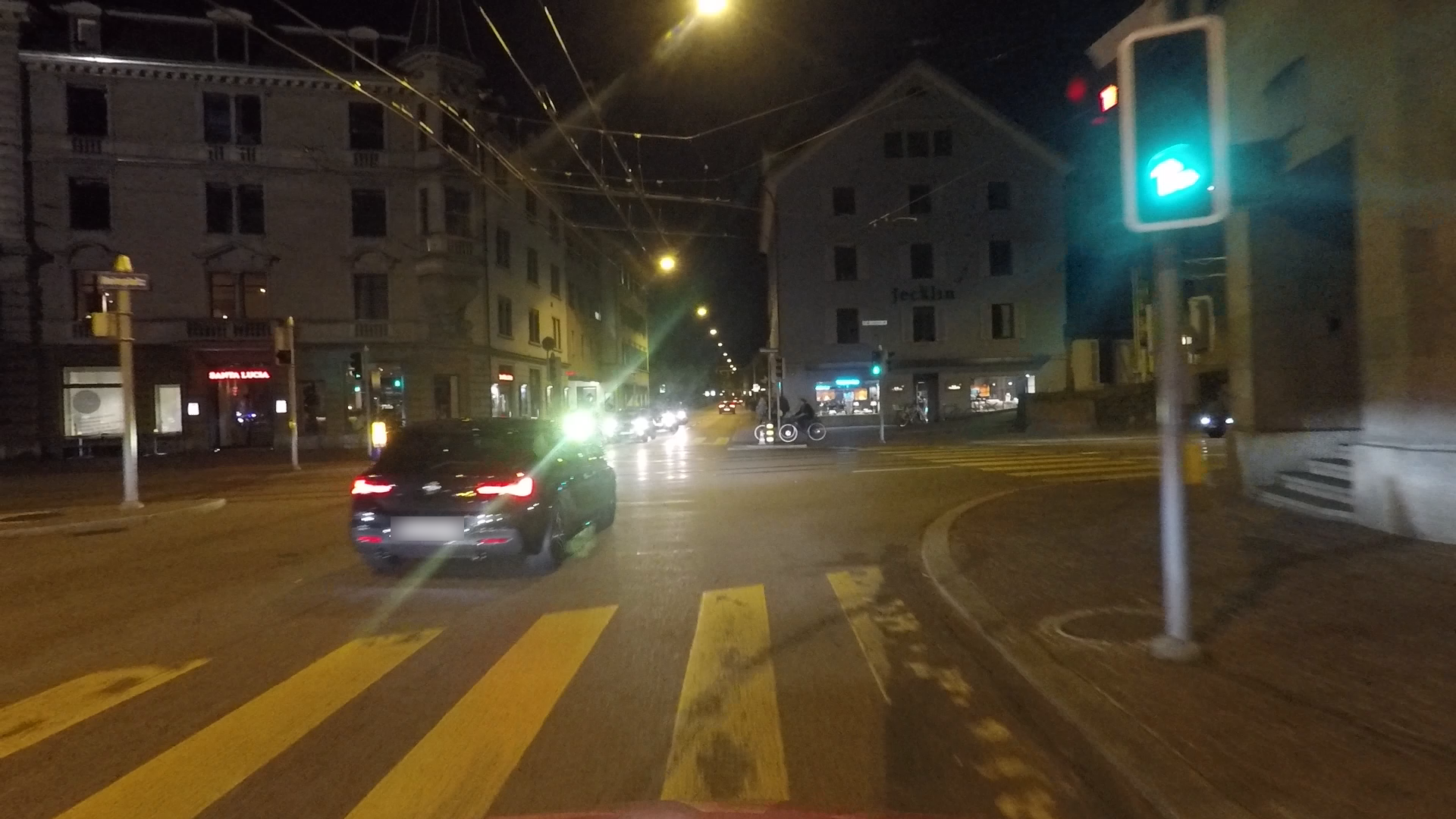}
        \end{subfigure}
        \hfill
        \begin{subfigure}{0.47\linewidth}
            \centering
            \includegraphics[width=\linewidth]{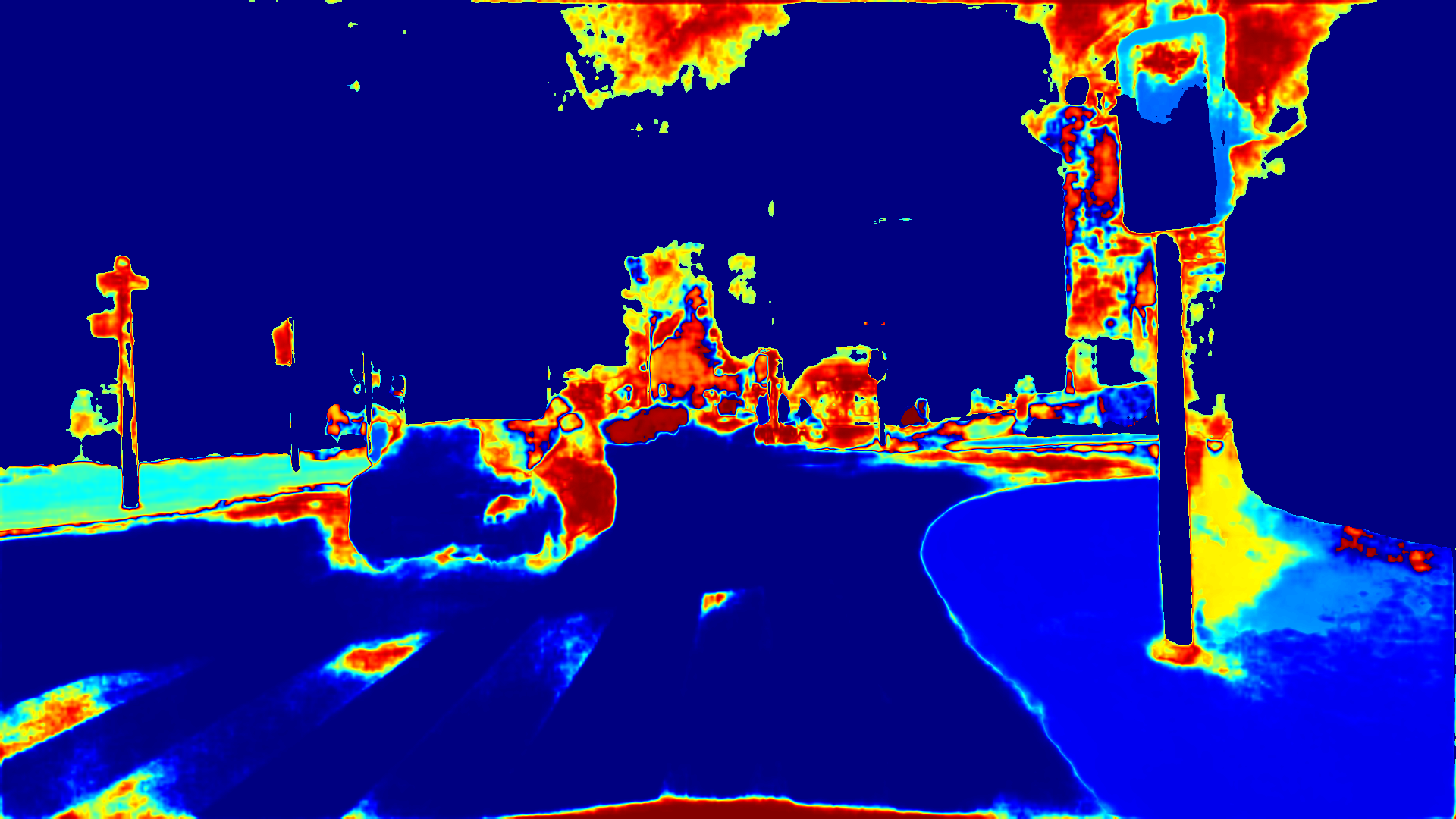}
        \end{subfigure}
        \vspace{0.5cm}
        \centering
        \small \textbf{ACDC Night Dataset}
    \end{minipage}

    \begin{minipage}{\linewidth}
        \centering
        \begin{subfigure}{0.47\linewidth}
            \centering
            \includegraphics[width=\linewidth]{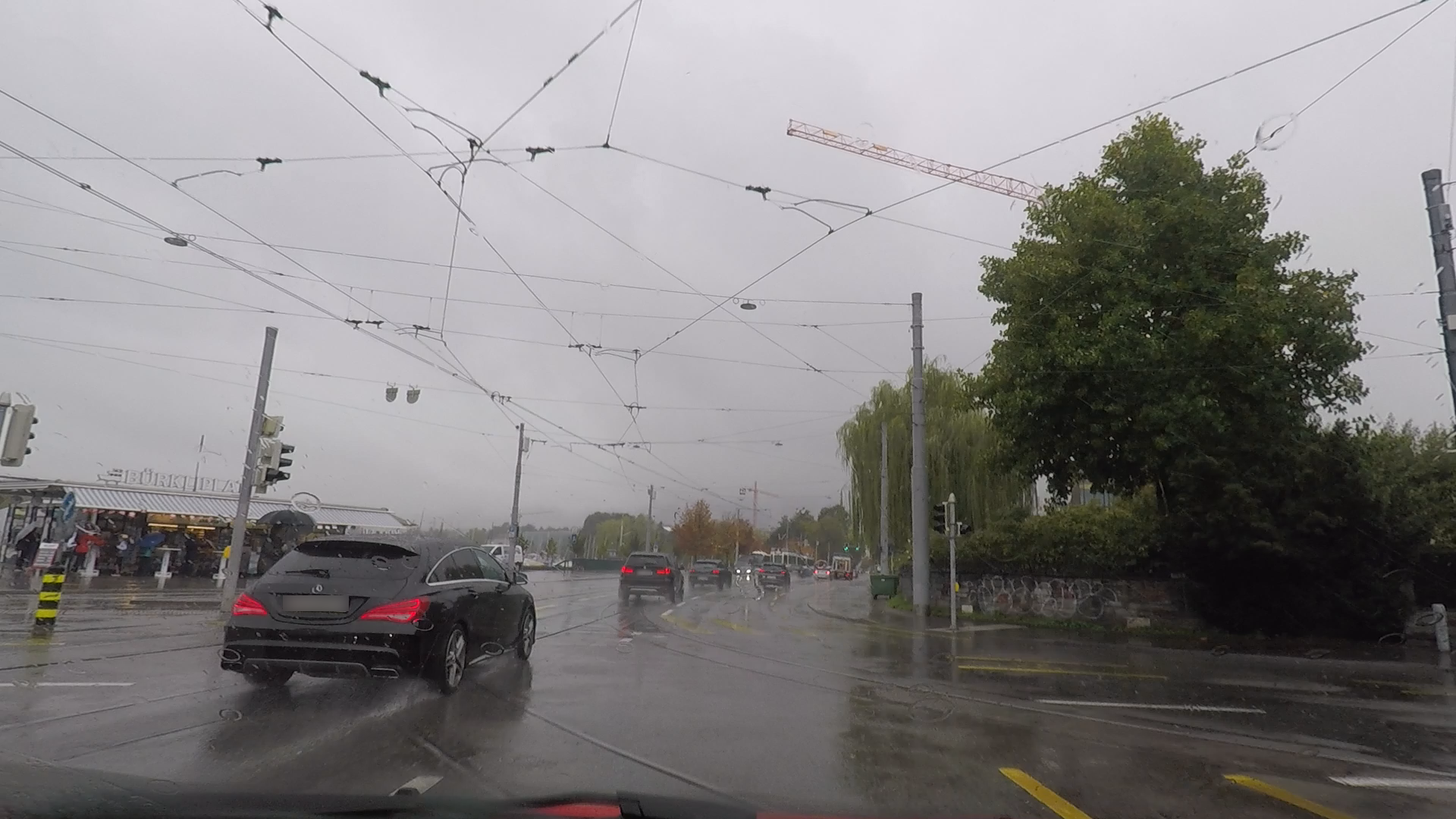}

        \end{subfigure}
        \hfill
        \begin{subfigure}{0.47\linewidth}
            \centering
            \includegraphics[width=\linewidth]{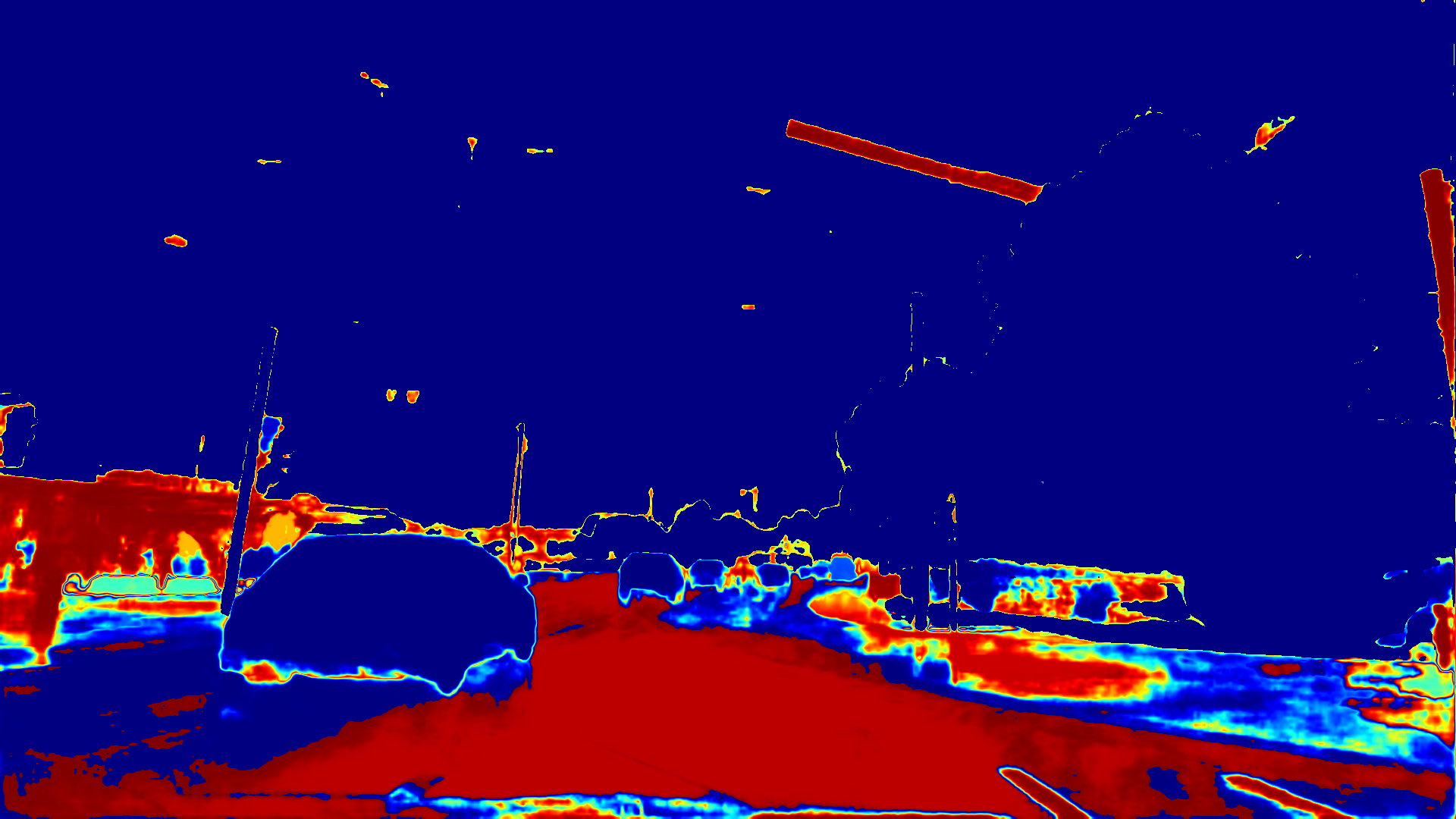}

        \end{subfigure}
        \vspace{0.5cm}
        \centering
        \small \textbf{ACDC Rain Dataset} 
    \end{minipage}

    \begin{minipage}{\linewidth}
        \centering
        \begin{subfigure}{0.47\linewidth}
            \centering
            \includegraphics[width=\linewidth]{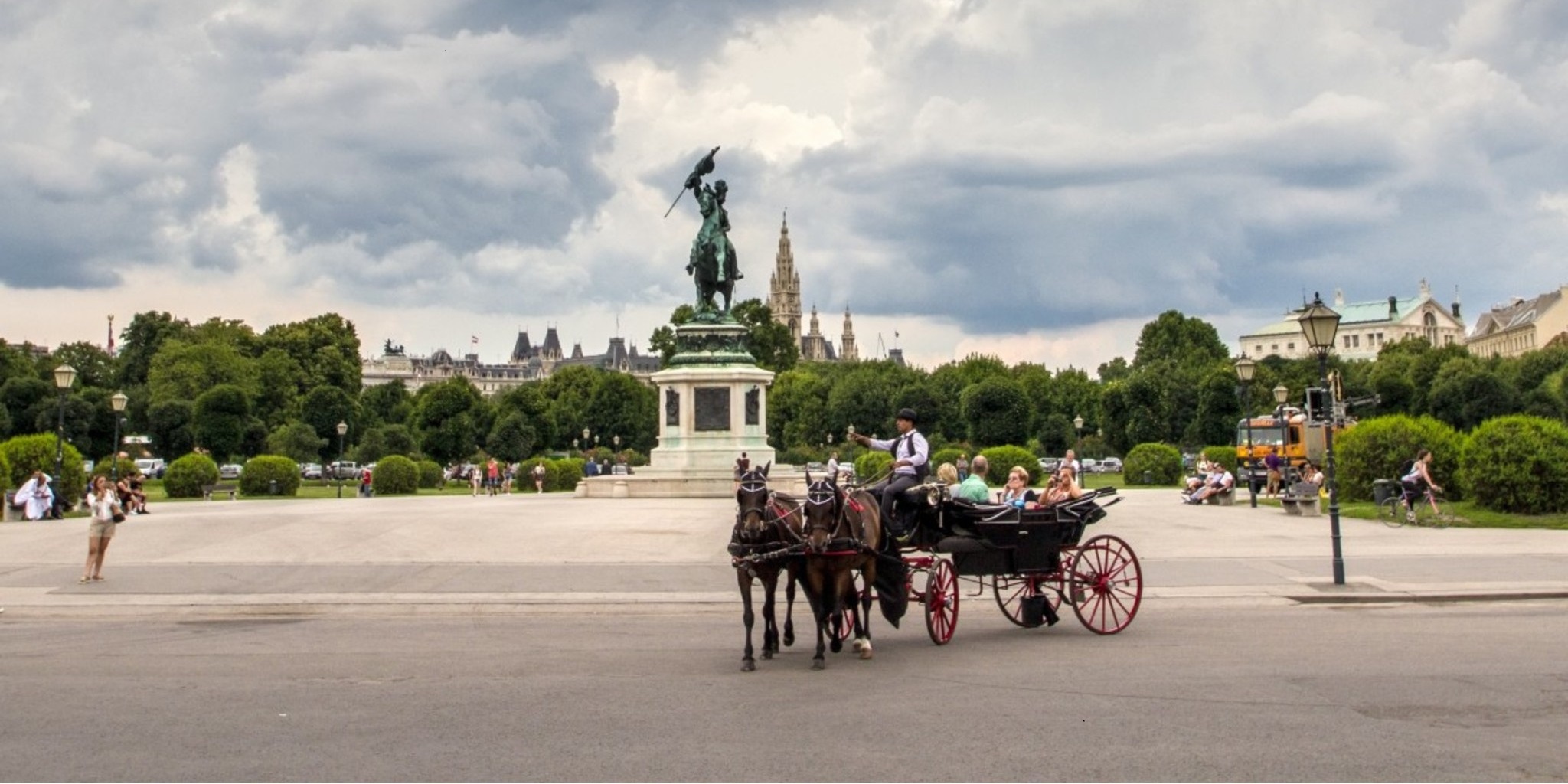}
        \end{subfigure}
        \hfill
        \begin{subfigure}{0.47\linewidth}
            \centering
            \includegraphics[width=\linewidth]{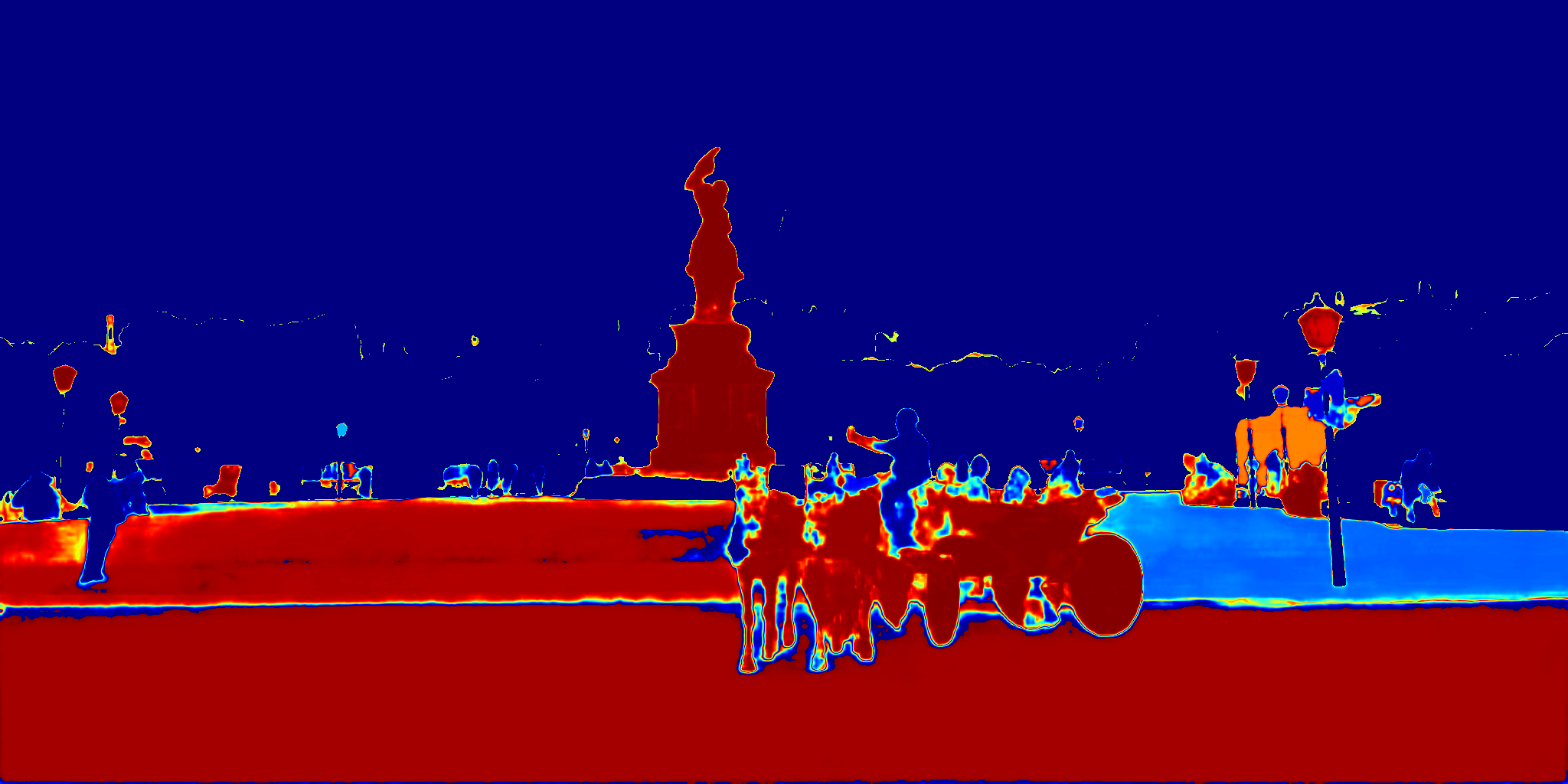}
        \end{subfigure}
        \vspace{0.5cm}
        \centering
        \small \textbf{SegmentMeIfYouCan Anomaly Track Dataset}
    \end{minipage}

    \caption{Visualization of Mask2Anomaly \cite{rai2024mask2anomaly} applied to selected images exhibiting semantic and covariate shifts. The left column presents the original images from various datasets, including Lost and Found \cite{pinggera2016lost}, ACDC Night and Rain \cite{sakaridis2021acdc}, and SegmentMeIfYouCan Anomaly Track \cite{chan2021segmentmeifyoucan} The right column displays the corresponding OOD object-level maps generated by Mask2Anomaly.}
    \label{fig:Mast2Former_Application}
\end{figure}

Pixel-level out-of-distribution (OOD) segmentation methods aim to identify pixels that deviate from the known distribution of semantic classes in training data. 
\textbf{This method is about image-level OOD detection (\enquote{is there any anomaly}), not pixel-level OOD segmentation (\enquote{where is an anomaly})}. We detail the motivation for this in the following.

Practically, input monitoring complements pixel-wise OOD segmentation by enabling more efficient anomaly detection and response for the following reasons:
\begin{itemize}
    \item In case there is a shift in one part of the image, the probability that there are further (undetected) shifts elsewhere in the image is very high. Therefore, a reprocessing of the complete image using a safety mechanism is recommended. This requires an accurate prediction of whether there is \emph{any} anomaly in the image.
    And despite considerable progress in pixel-level OOD segmentation \cite{bevandic2019simultaneous, grcic2023advantages, vojivr2023image, rottmann2020prediction, nayal2023rba, chan2021entropy, kendall2017uncertainties, rai2024mask2anomaly, grcic2023advantages, zhang2024csl, grcic2022densehybrid, liu2023residual}, these are still challenged by a \textbf{{high rate of false positives}} (illustrated in \cref{fig:Mast2Former_Application}).
    This lies in the nature of problem formulation: Ensuring that there is no false positive among the tremendous amount of pixels in an image requires way more accurate predictions than a single image level OOD detection.
    This issue is illustrated in \cref{fig:Mast2Former_Application} via representative examples of Mask2Anomaly \cite{rai2024mask2anomaly}, a leading method from the SMIYC benchmark \cite{chan2021segmentmeifyoucan}. While pixel-wise OOD segmentation provides detailed information about environmental anomalies, its practical deployment in safety-critical applications remains challenged by high false positive rates. 
    \item Furthermore, pixel-level OOD detection methods typically \textbf{{require supervision during training}} and exhibit performance degradation when encountering objects or scenes outside their training domain. Consequently, deploying pixel-level OOD segmentation methods in real-world applications requires an additional system capable of determining when these models can be reliably employed. Given that the operational domain of AD encompasses diverse semantic and covariate shifts, deploying such methods necessitates an additional system to determine their applicability.
\end{itemize} 
These limitations underscore the importance of operation-level input monitoring for robust assessment of perception system reliability across varying environmental conditions.

Recently, Gao et~al.\ \cite{gao2024generalize} addressed the challenge of pixel-level semantic segmentation under domain shift by leveraging generative models for training data augmentation. While their implementation is not yet publicly available, evaluating their approach across diverse covariate shift scenarios presents a promising direction for future investigation.

\subsection{Why VFMs for Feature Extraction Instead of Standard Vision Models}
\label{sec:ComparativeBaselineAnalysis}

In this section we provide \textbf{evidence that VFM feature encodings are clearly superior to those of classical feature extractor DNNs}.
This motivated our choice to restrict further analysis of OOD detection capabilities to the scope of VFM feature encodings.

To contextualize the performance of VFMs, we establish comparative baselines using analogous architectures pretrained on ImageNet \cite{deng2009imagenet} and Autoencoders (AEs) trained on our target domain data. These experiments provide detailed evidence supporting our main paper's finding that VFMs outperform non-foundation model latent representations in input monitoring. Our baseline models include ImageNet-pretrained architectures: ResNet variants (\eg RN50, RN101, RN152) \cite{he2016deep}, Vision Transformers (\eg ViT-B/16, ViT-L/16) \cite{dosovitskiy2020vit}, and Swin Transformers (\ie SwinB, SwinT) \cite{liu2021swin}\footnote{Implementation details: ResNet models from torchvision (\url{https://github.com/pytorch/vision}), ViT and Swin Transformer models from timm (\url{https://github.com/huggingface/pytorch-image-models})}. Additionally, we train domain-specific AEs with varying latent space dimensions on the Cityscapes training set \cite{cordts2016cityscapes}.

\paragraph{Baseline Setup}
For all architectures (ImageNet-pretrained models and AEs), we model their latent space distributions using the approaches presented in the main paper: APS, OC-SVM, GMM, and NF. We evaluate these models using the Cityscapes dataset, where the training set serves as $\mathcal{X}_\mathrm{monitor}$ and the validation set as $\mathcal{X}_\mathrm{ID}$. To ensure comprehensive evaluation, we test on both semantic and covariate shift scenarios. Detailed implementation specifications and architectural configurations are provided in \cref{sec:Implementation Details}.

\paragraph{Results}
Based on the comparative analysis in \cref{tab:baselines}, \finding{VFMs consistently outperform ImageNet-trained models and AEs} across datasets and different shifts, \ie semantic and covariate. We observe three critical findings through the lens of FPR95: First, VFMs demonstrate superior false positive control, with CLIP-RN50x64 achieving FPR95 of 0.24 on Lost and Found, outperforming ImageNet-trained counterparts. Second, the choice of density estimator proves crucial - while GMM provides competitive results, normalizing flows unlock exceptional performance for hierarchical architectures, enabling Grounding DINO's SwinB to achieve state-of-the-art FPR95 of 0.16 on Lost and Found and perfect detection (FPR95 = 0.00) on Bravo-Synrain. Finally, despite their structural simplicity, autoencoders struggle with false positive control (FPR95 $>$ 0.65 across all configurations), highlighting that architectural sophistication in VFMs, rather than mere representation learning, is key to reliable input monitoring.

\begin{table*}[ht!]
\centering
\resizebox{\textwidth}{!}{
\begin{tabular}{cc||ccc|ccc|ccc|ccc|ccc}
\hline
\multirow{3}{*}{\textbf{Model}} & \multicolumn{1}{c||}{\multirow{3}{*}{\textbf{Backbone}}} & \multicolumn{15}{c}{\textbf{Lost and Found}} \\ \cline{3-17} 
& \multicolumn{1}{c||}{} & \multicolumn{3}{c|}{\textbf{APS}} & \multicolumn{3}{c|}{\textbf{MFS}} & \multicolumn{3}{c|}{\textbf{OC-SVM}} & \multicolumn{3}{c|}{\textbf{GMM}} & \multicolumn{3}{c}{\textbf{NF}} \\ \cline{3-17} 
& \multicolumn{1}{c||}{} & \multicolumn{1}{c|}{\textbf{AUROC}↑} & \multicolumn{1}{c|}{\textbf{AUPR}↑} & \multicolumn{1}{c|}{\textbf{FPR95}↓} & \multicolumn{1}{c|}{\textbf{AUROC}↑} & \multicolumn{1}{c|}{\textbf{AUPR}↑} & \multicolumn{1}{c|}{\textbf{FPR95}↓} & \multicolumn{1}{c|}{\textbf{AUROC}↑} & \multicolumn{1}{c|}{\textbf{AUPR}↑} & \multicolumn{1}{c|}{\textbf{FPR95}↓} & \multicolumn{1}{c|}{\textbf{AUROC}↑} & \multicolumn{1}{c|}{\textbf{AUPR}↑} & \multicolumn{1}{c|}{\textbf{FPR95}↓} & \multicolumn{1}{c|}{\textbf{AUROC}↑} & \multicolumn{1}{c|}{\textbf{AUPR}↑} & \textbf{FPR95}↓ \\ \hline

\multirow{7}{*}{\textbf{\begin{tabular}[c]{@{}c@{}}Trained\\ on\\ ImageNet\end{tabular}}} 
& \multicolumn{1}{c||}{\textbf{RN50}} & 0.73 & 0.69 & 0.70 & 0.73 & 0.69 & 0.70 & 0.76 & 0.71 & 0.62 & 0.70 & 0.67 & 0.78 & 0.75 & 0.71 & 0.63 \\
& \multicolumn{1}{c||}{\textbf{RN101}} & 0.70 & 0.64 & 0.68 & 0.70 & 0.64 & 0.68 & 0.72 & 0.65 & 0.60 & 0.69 & 0.65 & 0.81 & 0.75 & 0.70 & 0.63 \\
& \multicolumn{1}{c||}{\textbf{RN152}} & 0.68 & 0.63 & 0.74 & 0.68 & 0.63 & 0.74 & 0.72 & 0.68 & 0.77 & 0.67 & 0.62 & 0.77 & 0.70 & 0.65 & 0.73 \\
& \multicolumn{1}{c||}{\textbf{ViT-B/16}} & 0.79 & 0.75 & 0.53 & 0.79 & 0.75 & 0.53 & 0.83 & 0.81 & 0.53 & 0.81 & 0.78 & 0.53 & 0.82 & 0.79 & 0.56 \\
& \multicolumn{1}{c||}{\textbf{ViT-L/16}} & 0.88 & 0.85 & 0.39 & 0.88 & 0.85 & 0.39 & 0.86 & 0.83 & 0.47 & 0.88 & 0.85 & 0.40 & 0.88 & 0.85 & 0.42 \\
& \multicolumn{1}{c||}{\textbf{SwinB}} & 0.84 & 0.82 & 0.53 & 0.84 & 0.82 & 0.53 & 0.83 & 0.84 & 0.54 & 0.85 & 0.84 & 0.53 & 0.86 & 0.85 & 0.44 \\
& \multicolumn{1}{c||}{\textbf{SwinT}} & 0.72 & 0.64 & 0.62 & 0.72 & 0.64 & 0.62 & 0.74 & 0.70 & 0.75 & 0.73 & 0.66 & 0.65 & 0.72 & 0.67 & 0.77 \\ \hline

\multirow{4}{*}{\textbf{Autoencoder}} 
& \multicolumn{1}{c||}{\textbf{128}} & 0.66 & 0.61 & 0.71 & 0.55 & 0.52 & 0.80 & 0.62 & 0.56 & 0.74 & 0.62 & 0.55 & 0.79 & 0.42 & 0.43 & 0.94 \\
& \multicolumn{1}{c||}{\textbf{256}} & 0.68 & 0.62 & 0.69 & 0.70 & 0.67 & 0.69 & 0.68 & 0.64 & 0.77 & 0.52 & 0.48 & 0.87 & 0.60 & 0.56 & 0.82 \\
& \multicolumn{1}{c||}{\textbf{512}} & 0.65 & 0.61 & 0.73 & 0.68 & 0.63 & 0.70 & 0.62 & 0.58 & 0.74 & 0.61 & 0.58 & 0.88 & 0.55 & 0.53 & 0.88 \\
& \multicolumn{1}{c||}{\textbf{1024}} & 0.70 & 0.67 & 0.74 & 0.77 & 0.74 & 0.56 & 0.62 & 0.57 & 0.84 & 0.65 & 0.60 & 0.83 & 0.68 & 0.63 & 0.65 \\ \hline

\multirow{3}{*}{\textbf{Ours}} 
& \multicolumn{1}{c||}{\textbf{Clip-RN50}} & 0.85 & 0.82 & 0.49 & 0.85 & 0.82 & 0.49 & 0.90 & 0.88 & 0.36 & 0.85 & 0.83 & 0.51 & 0.82 & 0.78 & 0.52 \\
& \multicolumn{1}{c||}{\textbf{Clip-RN50x64}} & \textbf{0.94} & \textbf{0.93} & \textbf{0.24} & \textbf{0.94} & \textbf{0.93} & \textbf{0.24} & \textbf{0.94} & \textbf{0.93} & \textbf{0.23} & \textbf{0.94} & \textbf{0.93} & \textbf{0.24} & 0.88 & 0.84 & 0.37 \\
& \multicolumn{1}{c||}{\textbf{SwinB}} & 0.88 & 0.84 & 0.47 & 0.88 & 0.84 & 0.47 & 0.91 & 0.89 & 0.32 & 0.84 & 0.80 & 0.54 & \textbf{0.95} & \textbf{0.94} & \textbf{0.16} \\ \hline

& \multicolumn{1}{c||}{} & \multicolumn{15}{c}{\textbf{Bravo Synrain}} \\ \hline

\multirow{7}{*}{\textbf{\begin{tabular}[c]{@{}c@{}}Trained\\ on\\ ImageNet\end{tabular}}} 
& \multicolumn{1}{c||}{\textbf{RN50}} & 0.90 & 0.89 & 0.33 & 0.79 & 0.83 & 0.82 & 0.94 & 0.93 & 0.23 & 0.92 & 0.91 & 0.27 & 0.97 & 0.97 & 0.12 \\
& \multicolumn{1}{c||}{\textbf{RN101}} & 0.89 & 0.87 & 0.36 & 0.81 & 0.85 & 0.79 & 0.93 & 0.92 & 0.27 & 0.91 & 0.90 & 0.30 & 0.97 & 0.96 & 0.12 \\
& \multicolumn{1}{c||}{\textbf{RN152}} & 0.90 & 0.89 & 0.35 & 0.80 & 0.84 & 0.78 & 0.95 & 0.94 & 0.18 & 0.93 & 0.91 & 0.29 & 0.97 & 0.96 & 0.10 \\
& \multicolumn{1}{c||}{\textbf{ViT-B/16}} & 0.86 & 0.82 & 0.38 & 0.81 & 0.81 & 0.69 & 0.93 & 0.92 & 0.24 & 0.91 & 0.88 & 0.26 & 0.97 & 0.96 & 0.12 \\
& \multicolumn{1}{c||}{\textbf{ViT-L/16}} & 0.88 & 0.84 & 0.31 & 0.85 & 0.88 & 0.70 & 0.92 & 0.91 & 0.24 & 0.91 & 0.88 & 0.23 & 0.98 & 0.97 & 0.07 \\
& \multicolumn{1}{c||}{\textbf{SwinB}} & 0.90 & 0.88 & 0.31 & 0.84 & 0.86 & 0.67 & 0.96 & 0.96 & 0.14 & 0.95 & 0.93 & 0.19 & 0.99 & 0.99 & 0.05 \\
& \multicolumn{1}{c||}{\textbf{SwinT}} & 0.88 & 0.84 & 0.31 & 0.76 & 0.75 & 0.77 & 0.95 & 0.93 & 0.16 & 0.93 & 0.90 & 0.20 & 0.99 & 0.98 & 0.04 \\ \hline

\multirow{4}{*}{\textbf{Autoencoder}} 
& \multicolumn{1}{c||}{\textbf{128}} & 0.60 & 0.59 & 0.87 & 0.58 & 0.55 & 0.87 & 0.59 & 0.56 & 0.86 & 0.60 & 0.58 & 0.88 & 0.55 & 0.52 & 0.87 \\
& \multicolumn{1}{c||}{\textbf{256}} & 0.60 & 0.59 & 0.85 & 0.57 & 0.55 & 0.87 & 0.58 & 0.55 & 0.87 & 0.54 & 0.53 & 0.90 & 0.59 & 0.55 & 0.85 \\
& \multicolumn{1}{c||}{\textbf{512}} & 0.60 & 0.60 & 0.84 & 0.56 & 0.54 & 0.88 & 0.56 & 0.55 & 0.87 & 0.60 & 0.59 & 0.88 & 0.57 & 0.54 & 0.85 \\
& \multicolumn{1}{c||}{\textbf{1024}} & 0.57 & 0.55 & 0.87 & 0.59 & 0.56 & 0.86 & 0.60 & 0.57 & 0.86 & 0.52 & 0.51 & 0.86 & 0.63 & 0.59 & 0.81 \\ \hline

\multirow{3}{*}{\textbf{Ours}} 
& \multicolumn{1}{c||}{\textbf{Clip-RN50}} & 0.76 & 0.72 & 0.59 & 0.76 & 0.72 & 0.59 & 0.85 & 0.82 & 0.43 & 0.75 & 0.71 & 0.64 & 0.95 & 0.92 & 0.16 \\
& \multicolumn{1}{c||}{\textbf{Clip-RN50x64}} & 0.99 & 0.99 & 0.03 & 0.99 & 0.99 & 0.03 & \textbf{1.00} & \textbf{1.00} & 0.01 & \textbf{1.00} & \textbf{1.00} & \textbf{0.01} & 0.99 & 0.99 & 0.04 \\
& \multicolumn{1}{c||}{\textbf{SwinB}} & \textbf{1.00} & \textbf{1.00} & \textbf{0.02} & \textbf{1.00} & \textbf{1.00} & \textbf{0.02} & \textbf{1.00} & \textbf{1.00} & \textbf{0.00} & \textbf{1.00} & 0.99 & 0.02 & \textbf{1.00} & \textbf{1.00} & \textbf{0.00} \\ \hline

\end{tabular}
}
\caption{Comparative evaluation of out-of-distribution detection performance using ImageNet-pretrained models, autoencoders, and VFMs on Lost and Found \cite{pinggera2016lost} and Bravo-Synrain \cite{loiseau2023reliability} benchmarks. We evaluate using AUROC, AUPR, and FPR95 metrics across five methods: APS, MFS, OC-SVM, GMM, and NF. For VFMs, we utilize image encoders from CLIP \cite{radford2021learning} with ResNet-50 and ResNet-50x64 backbones, and from Grounding-DINO \cite{liu2023grounding} with SwinB backbone. Best performances for each metric and dataset are highlighted in bold.}
\label{tab:baselines}
\end{table*}

\subsection{Real-time Capability of Chosen Models}
\label{sec:realtime_capability}

A critical consideration for practical deployment in AD systems is computational efficiency. Our approach leverages only the image encoder components of Vision Foundation Models (VFMs), specifically employing CLIP's ResNet variants \cite{radford2021learning} and Grounding DINO's Swin Transformer backbones \cite{liu2023grounding}. Recent literature substantiates the real-time capabilities of these architectures:

\begin{itemize}
\item Zhao et al.\cite{zhao2024detrs} demonstrate that \textbf{DETR-based architectures achieve real-time performance} while surpassing YOLO-based models in detection accuracy.
\item Kassab et al.\cite{kassab2024language} have \textbf{successfully deployed CLIP's ResNet backbones for real-time} visual scene understanding in SLAM applications, validating their computational feasibility.
\end{itemize}

Our framework offers additional deployment flexibility through temporal and spatial optimization strategies, including:
\begin{itemize}
\item Processing \textbf{every $n$th frame} to reduce computational load
\item Operating on \textbf{downsampled inputs} while maintaining monitoring efficacy
\end{itemize}

To quantify real-time performance, we benchmarked all feature extractors on hardware (Intel Xeon Gold 6326 CPU, NVIDIA A40 GPU) using randomly sampled 50 full-resolution (2048×1024) images from Cityscapes. Table~\ref{tab:fps_benchmark} presents frames-per-second (FPS) rates with standard deviations across 5 measurement runs.

\begin{table}[t!]
\centering
\tiny 
\setlength{\tabcolsep}{2pt} 
\resizebox{\columnwidth}{!}{%
\begin{tabular}{>{\centering\arraybackslash}p{0.9cm}>{\centering\arraybackslash}p{1.0cm}||>{\centering\arraybackslash}p{1.2cm}>{\centering\arraybackslash}p{1.0cm}}
\hline
\multirow{2}{*}{\textbf{Model}} & \multirow{2}{*}{\textbf{Backbone}} & \multicolumn{2}{c}{\textbf{FPS}} \\
\cline{3-4}
& & \textbf{Mean} & \textbf{Std} \\
\hline
\multirow{8}{*}{\centering\textbf{CLIP}} & \textbf{RN50} & 63.69 & 0.81 \\
& \textbf{RN101} & 63.29 & 0.40 \\
& \textbf{RN50x4} & 56.50 & 0.32 \\
& \textbf{RN50x16} & 42.37 & 0.90 \\
& \textbf{RN50x64} & 31.15 & 0.19 \\
& \textbf{ViT-B/32} & 63.29 & 1.60 \\
& \textbf{ViT-B/16} & 56.18 & 0.32 \\
& \textbf{ViT-L/14} & 49.50 & 0.74 \\
\hline
\multirow{5}{*}{\centering\textbf{DINO}} & \textbf{ViT-S/16} & 59.17 & 2.10 \\
& \textbf{ViT-S/8} & 49.50 & 0.00 \\
& \textbf{ViT-B/16} & 49.02 & 3.37 \\
& \textbf{ViT-B/8} & 42.74 & 0.18 \\
& \textbf{RN50} & 56.50 & 0.00 \\
\hline
\multirow{4}{*}{\centering\textbf{DINOv2}} & \textbf{ViT-S/14} & 53.19 & 1.70 \\
& \textbf{ViT-B/14} & 57.14 & 2.94 \\
& \textbf{ViT-L/14} & 47.62 & 0.23 \\
& \textbf{ViT-G/14} & 35.34 & 0.87 \\
\hline
\multirow{2}{*}{\centering\textbf{\begin{tabular}[c]{@{}c@{}}Grounding \\ DINO\end{tabular}}} & \textbf{SwinB} & 16.31 & 0.08 \\
& \textbf{SwinT} & 26.74 & 2.43 \\
\hline
\end{tabular}
}
\caption{Inference speed of VFMs' feature extractors, measured in frames per second (FPS) per image on high-resolution 2048×1024 input images. The FPS values represent the mean and standard deviation (Std) across multiple runs.}
\label{tab:fps_benchmark}
\end{table}

Our results indicate that CLIP's ResNet and ViT backbones provide the highest throughput, with RN50 and ViT-B/32 both exceeding 63 FPS. The DINOv2 and DINO variants maintain strong performance between 35-59 FPS. While Grounding DINO models demonstrate lower throughput (SwinB: 16.31 FPS, SwinT: 26.74 FPS), they remain viable for real-time applications—particularly given their superior OOD detection performance. These throughput rates can be improved by employing our proposed temporal optimization (processing every $n$th frame) or spatial optimization (downsampling input resolution). Recent work by Zhao et al.~\cite{zhao2024detrs} has shown that DETR-based architectures can surpass YOLO \cite{cheng2024yolo} in the inference with their optimizations which can be used in our case for Grounding-DINO measurements.

For distribution modeling approaches, computational efficiency varies significantly. APS requires computing pairwise similarities between all monitor and test samples, making it computationally prohibitive for real-time applications. In contrast, MFS, OC-SVM, GMM, and NF all offer efficient inference after offline training, as they only require evaluating new samples against pre-computed distribution parameters. Among these approaches, MFS provides the fastest inference but with reduced detection performance, while GMM and NF offer optimal performance-efficiency trade-offs for operational deployment.

\section{Details of Chosen Datasets and Vision Foundation Models}
\label{sec:models_and_datasets}

\subsection{Diversity of Chosen VMFs}
In our experiments, we employ four VFMs: CLIP \cite{radford2021learning}, DINO \cite{caron2021emerging}, DINOv2 \cite{oquab2023dinov2}, and Grounding DINO \cite{liu2023grounding}. A critical consideration in evaluating these models is whether their superior feature representations might be attributed to test data leakage. As shown below, the chosen models
\begin{itemize}
\item cover a wide range of architectures, as well as
\item a \textbf{wide range of different training datasets}
; and
\item \textbf{no test data leakage} influenced our results.
\end{itemize}

\begin{description}
    \item[CLIP \cite{radford2021learning}] was trained on a proprietary dataset of 400 million image-text pairs curated from publicly available internet data, leveraging natural language descriptions paired with corresponding images. In contrast, our datasets---\eg Cityscapes \cite{cordts2016cityscapes}, ACDC \cite{sakaridis2021acdc}, and Lost and Found \cite{pinggera2016lost}---are specifically designed for perception tasks, offering pixel-level semantic labels rather than natural language annotations. Moreover, these datasets' highly specialized characteristics—featuring calibrated front-facing cameras, standardized viewpoints, and tightly controlled acquisition protocols—differ fundamentally from the diverse, unconstrained nature of web-scraped imagery.

    \item[DINO and DINOv2]
    employ self-supervised learning paradigms, representing a distinct approach to training. While DINO is trained exclusively on ImageNet \cite{russakovsky2015imagenet} using self-distillation without labeled data, DINOv2 leverages the larger LVD-142M dataset \cite{oquab2023dinov2}. Although Cityscapes and ACDC are part of the LVD-142M corpus, our strongest results are achieved using CLIP and Grounding-DINO rather than DINOv2, suggesting that performance stems from architectural strengths rather than data memorization.

    \item[Grounding-DINO \cite{liu2023grounding}]
    demonstrates strong performance in our experiments, and was trained on detection datasets (COCO \cite{lin2014microsoft}, Objects365 \cite{shao2019objects365}, OpenImage \cite{krasin2017openimages}) and grounding datasets grouped as GoldG \cite{kamath2021mdetr} (Flickr30k entities \cite{plummer2015flickr30k}, Visual Genome \cite{krishna2017visual}) and RefC (RefCOCO variants)\cite{kamath2021mdetr}. Notably, our evaluation datasets were not part of its training corpus.
\end{description}
These findings indicate that VFMs' superior performance stems from their architectural and representational capabilities rather than training data overlap or test data leakage.

\subsection{Dataset Examples and Characteristics}
\label{sec:Datasets_appendix}

This section provides a comprehensive overview (see also \cref{tab:dataset_organization}) and samples of the datasets employed in our experimental framework and their strategic utilization for evaluating different types of distribution shifts. We categorize our evaluation scenarios into three distinct types of shifts:
\begin{enumerate}
    \item semantic shifts (novel objects or classes),
    \item covariate shifts (environmental and capture condition variations), and 
    \item combined shifts that exhibit both semantic and covariate variations simultaneously.
\end{enumerate}

\begin{figure}[htbp]
    \centering
    

    \begin{subfigure}{0.47\linewidth}
        \centering
        \includegraphics[width=\linewidth]{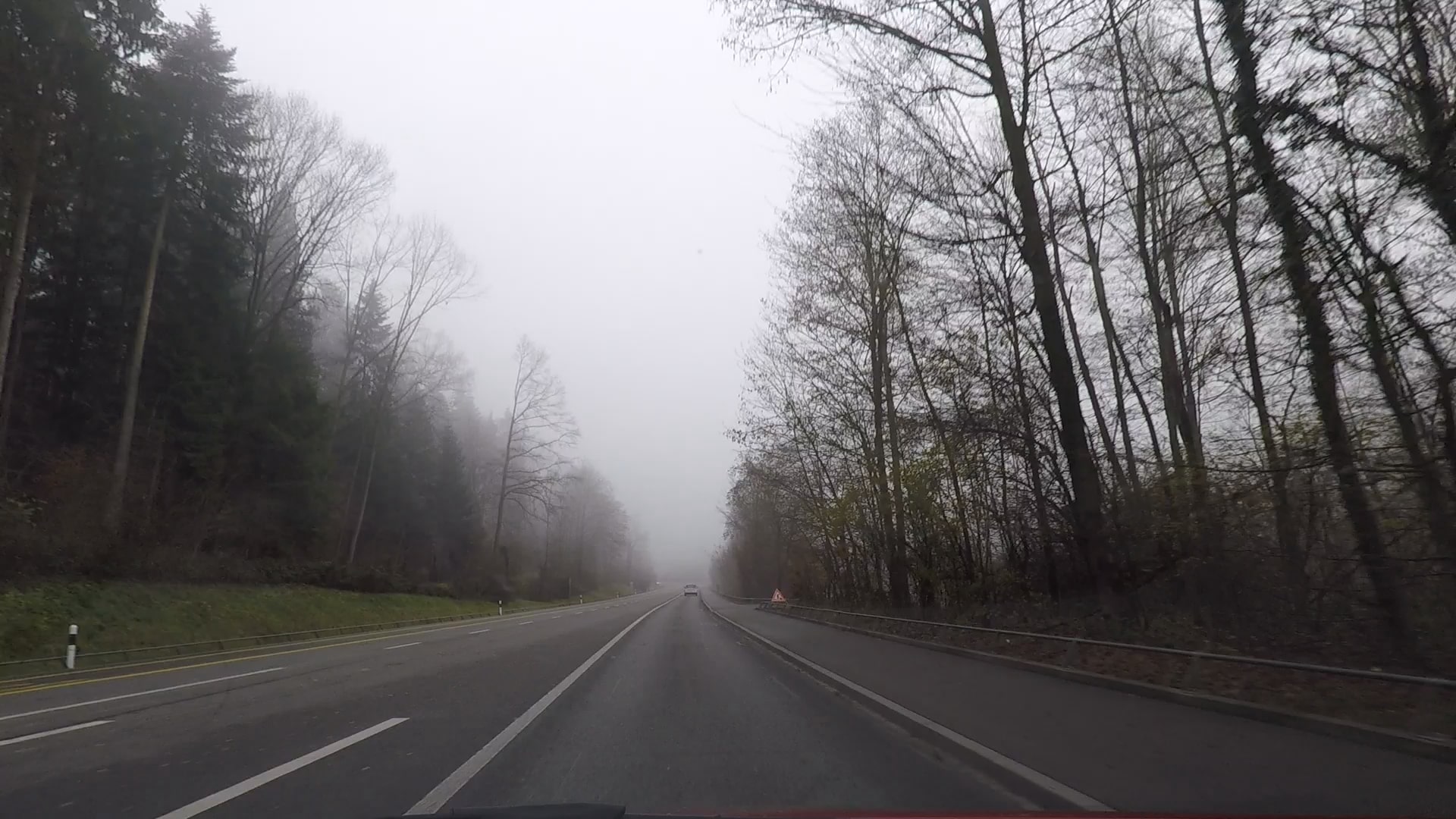}
        \caption{ACDC Fog}
    \end{subfigure}
    \hfill
    \begin{subfigure}{0.47\linewidth}
        \centering
        \includegraphics[width=\linewidth]{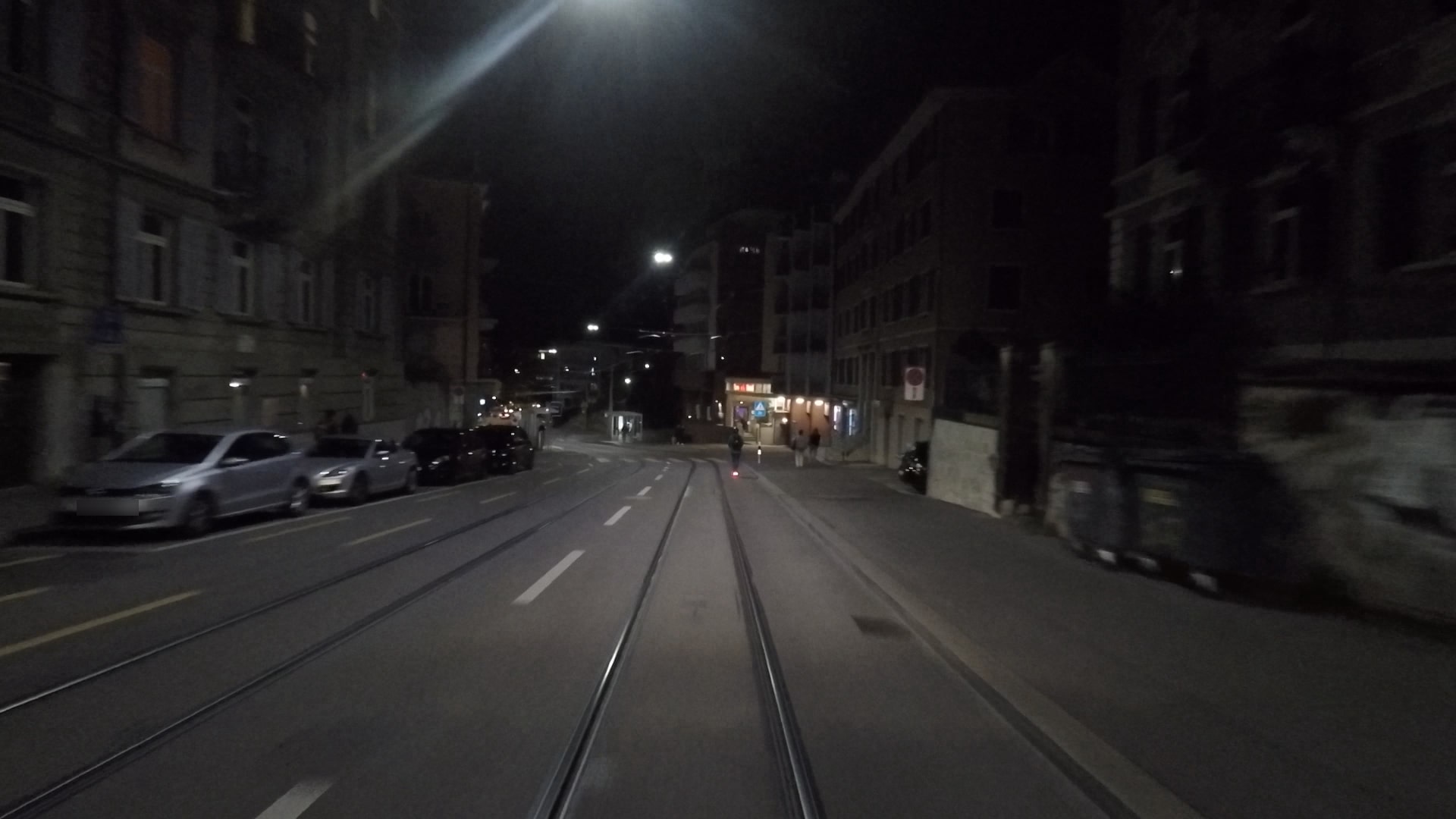}
        \caption{ACDC Night}
    \end{subfigure}

    \vspace{1cm}

     \begin{subfigure}{0.47\linewidth} 
        \centering
        \includegraphics[width=\linewidth]{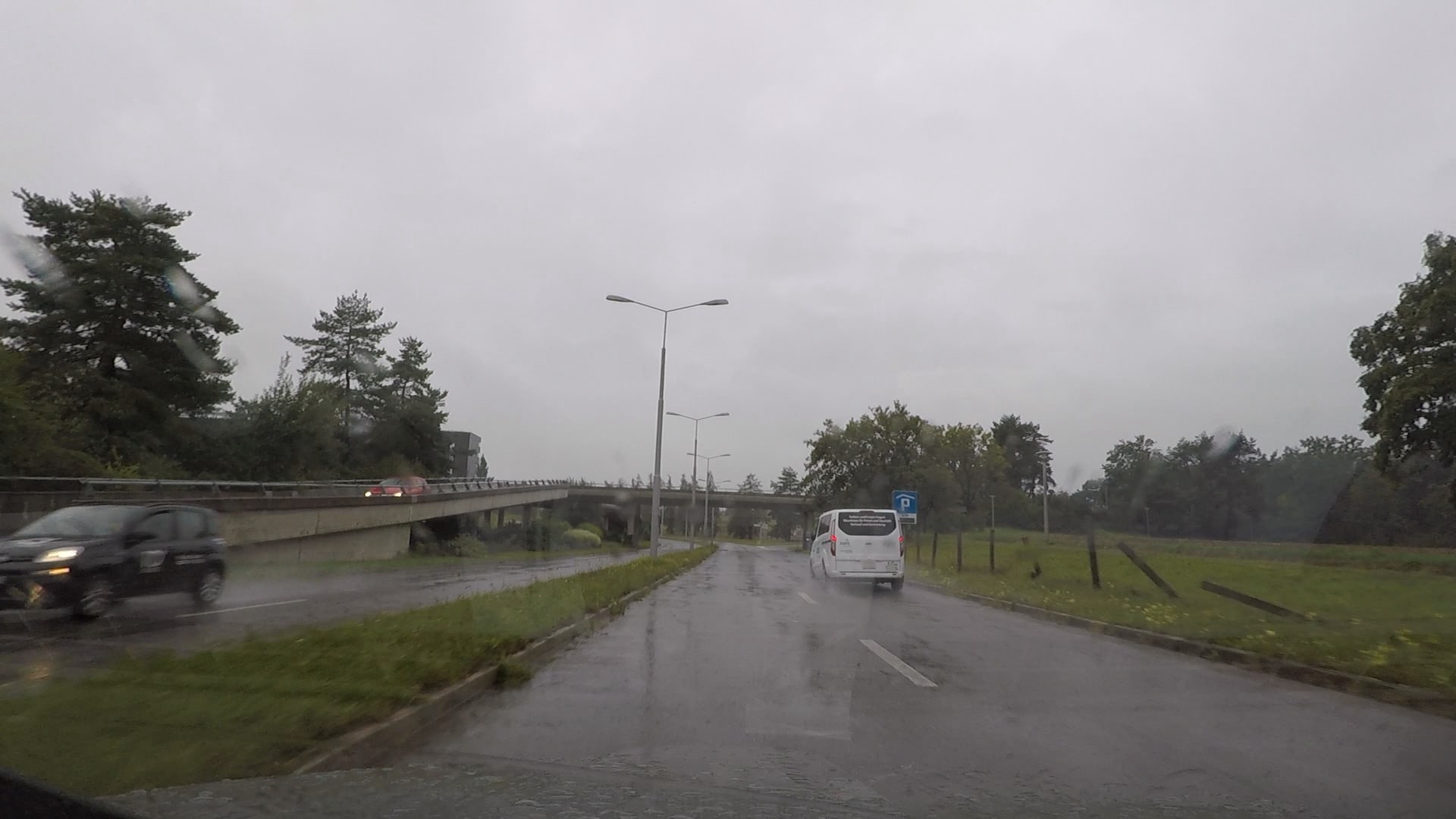}
        \caption{ACDC Rain}
    \end{subfigure}
    \hfill
    \begin{subfigure}{0.47\linewidth} 
        \centering
        \includegraphics[width=\linewidth]{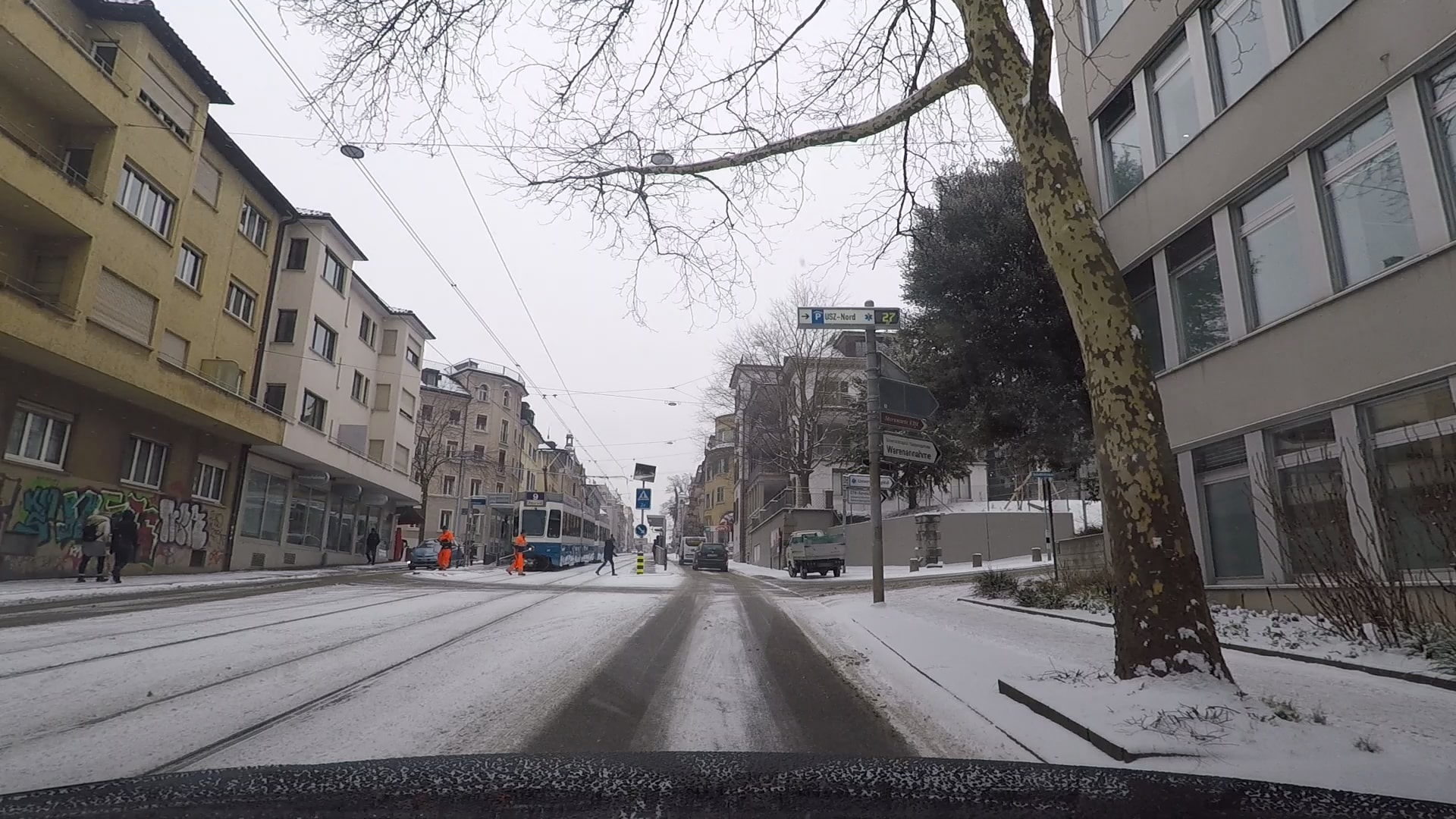}
        \caption{ACDC Snow}
    \end{subfigure}
    
    \caption{Example images from four subsets of ACDC Dataset \cite{sakaridis2021acdc}}
    \label{fig:acdc_example_images}
\end{figure}

\begin{description}
\item[ACDC Dataset:] 
    The ACDC dataset~\cite{sakaridis2021acdc} is designed for semantic scene understanding under adverse driving conditions, comprising four distinct weather scenarios: fog, night, rain, and snow. The dataset contains 4006 images with fine-grained pixel-level panoptic annotations, where each adverse condition image is paired with a corresponding reference image captured under clear conditions at the same location. Example images from each weather scenario are shown in Figure~\ref{fig:acdc_example_images}.

    For our experiments, we leverage the dataset's paired structure: reference images from the training set serve as $\mathcal{X}_\mathrm{monitor}$, reference images from the validation set as $\mathcal{X}_\mathrm{ID}$, and images from each adverse condition (\ie fog, night, rain, and snow) as separate $\mathcal{X}_\mathrm{test}$ sets. This setup enables systematic evaluation of covariate shift detection across different environmental conditions while maintaining consistent scene content.

\begin{figure}[htbp]
    \centering
    

    \begin{subfigure}{0.47\linewidth}
        \centering
        \includegraphics[width=\linewidth]{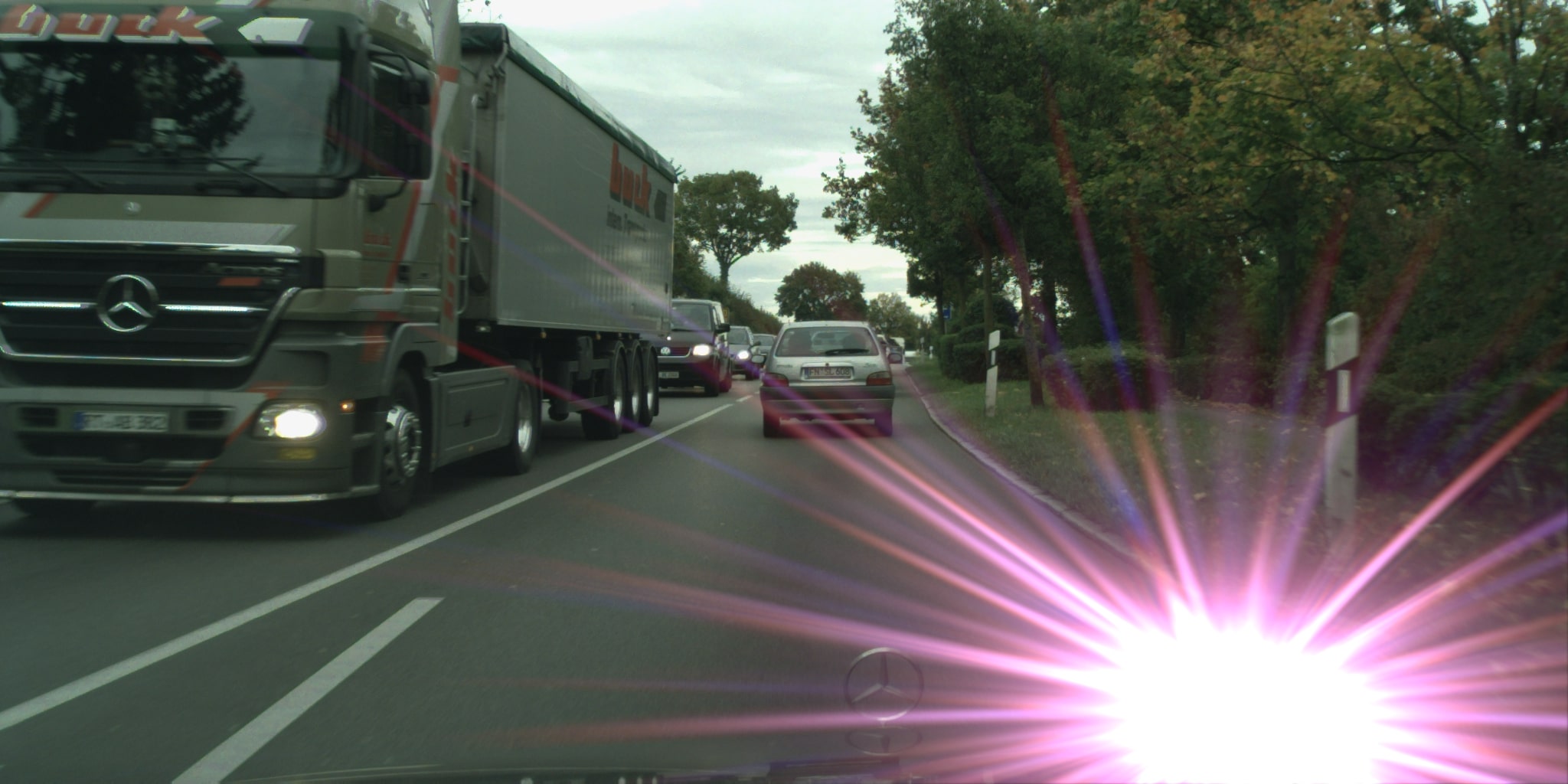}
        \caption{Bravo-Synflare}
    \end{subfigure}
    \hfill
    \begin{subfigure}{0.47\linewidth}
        \centering
        \includegraphics[width=\linewidth]{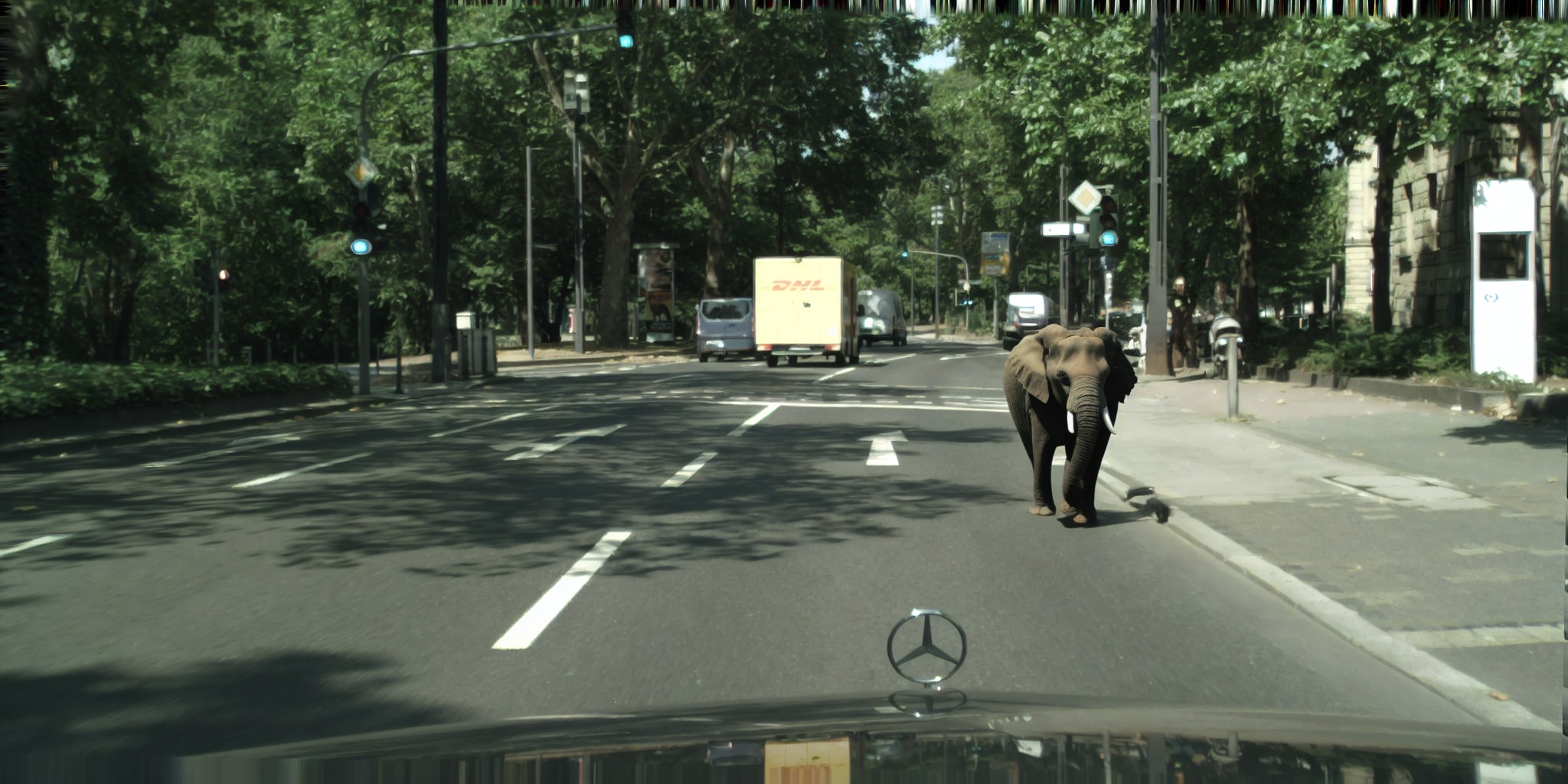}
        \caption{Bravo-Synobjs}
    \end{subfigure}

    \vspace{1cm}

     \begin{subfigure}{0.47\linewidth} 
        \centering
        \includegraphics[width=\linewidth]{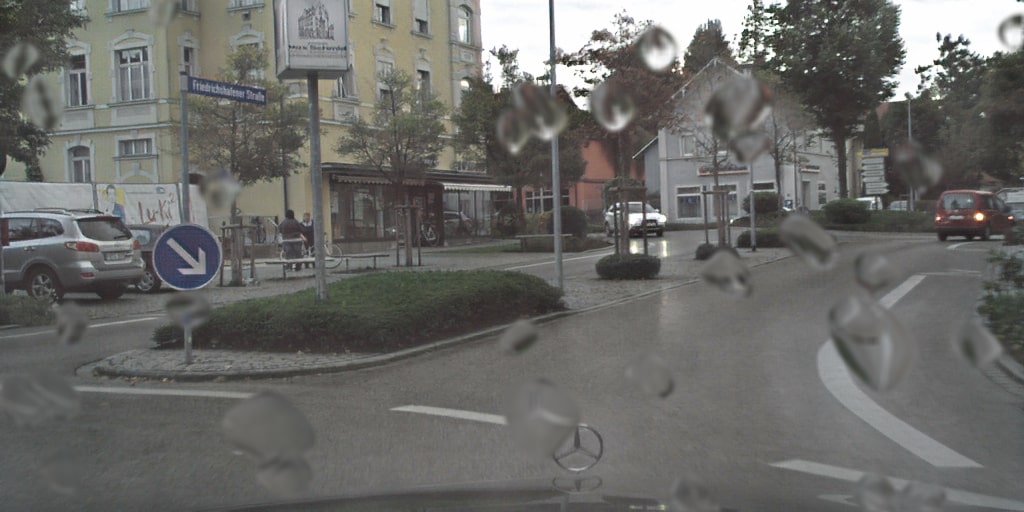}
        \caption{Bravo-Synrain}
    \end{subfigure}
    \hfill
    \begin{subfigure}{0.47\linewidth} 
        \centering
        \includegraphics[width=\linewidth]{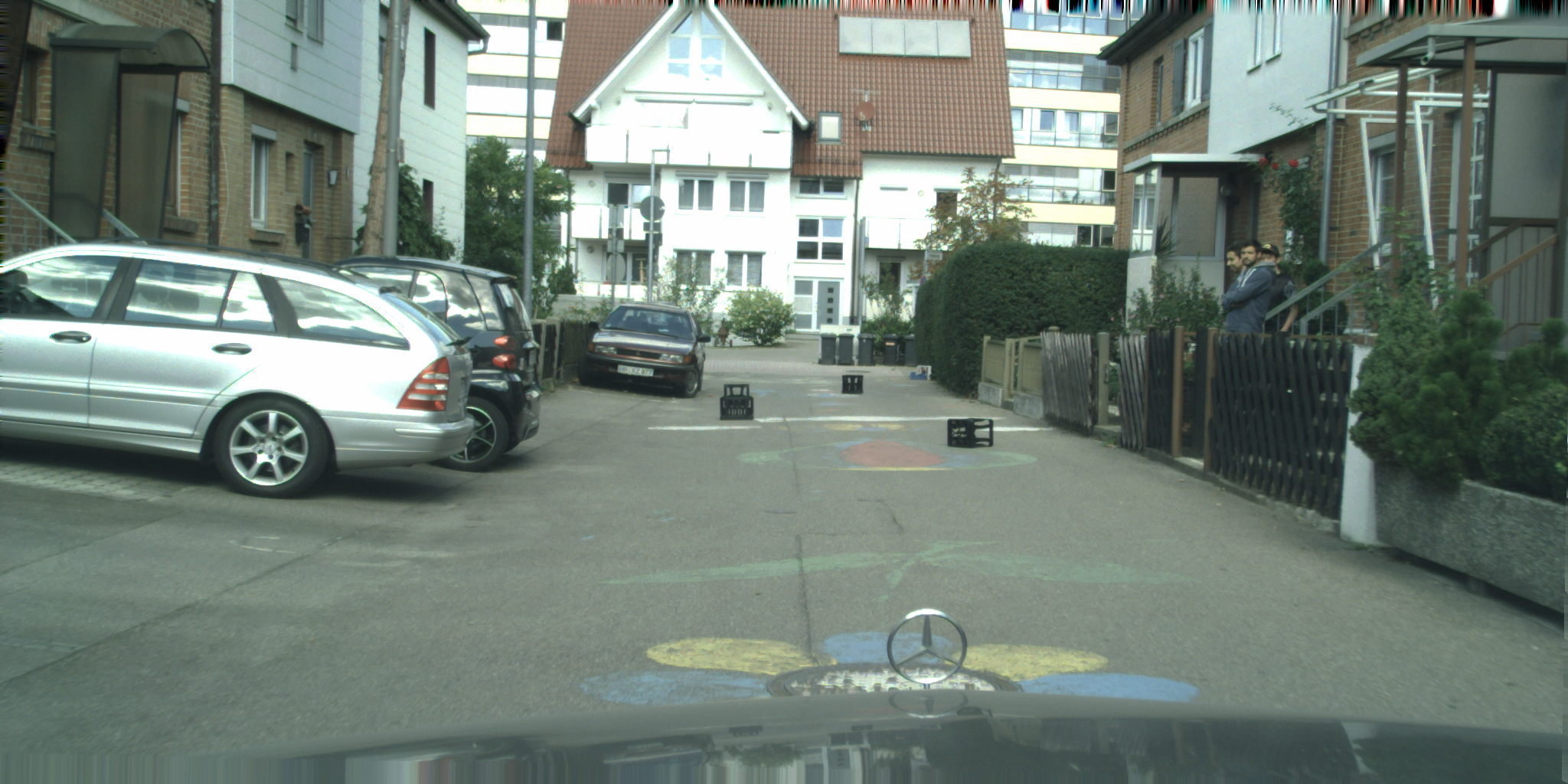}
        \caption{Lost and Found}
    \end{subfigure}
    
    \caption{Example images from Bravo datasets \cite{loiseau2023reliability} and Lost and Found Dataset \cite{pinggera2016lost}}
    \label{fig:bravo_example_images}
\end{figure}

\item[Bravo Datasets:]
    The Bravo dataset family~\cite{loiseau2023reliability} provides synthetic perturbations of Cityscapes images to evaluate model robustness under controlled distribution shifts. The collection comprises three distinct subsets:
    \begin{itemize}
        \item Bravo-Synflare (lens flare effects simulating intense light sources),
        \item Bravo-Synobjs (synthetic object insertions for semantic shifts), and
        \item Bravo-Synrain (synthetic precipitation effects).
    \end{itemize}
    Examples from each subset are shown in \cref{fig:bravo_example_images}.

We structure our experiments using Cityscapes training data as $\mathcal{X}_\mathrm{monitor}$ and validation data as $\mathcal{X}_\mathrm{ID}$. For $\mathcal{X}_\mathrm{test}$, we evaluate covariate shifts using Bravo-Synflare and Bravo-Synrain, while semantic shifts are assessed using Bravo-Synobjs. This configuration enables systematic evaluation of both types of distribution shifts while maintaining consistent base scene content from the Cityscapes domain.

\item[Lost and Found Dataset:]
    The Lost and Found dataset~\cite{pinggera2016lost} is designed for real-world anomaly detection in urban driving environments, specifically focusing on rare and unexpected road hazards such as traffic cones, cargo items, and other small obstacles that are typically absent from standard training datasets. Captured in German cities under similar environmental conditions as Cityscapes~\cite{cordts2016cityscapes}, it maintains consistent scene characteristics while introducing novel objects, making it particularly suitable for evaluating semantic shifts in the Cityscapes domain. Example images from this dataset are shown in \cref{fig:bravo_example_images}.

    For our experiments, we utilize Cityscapes training data as $\mathcal{X}_\mathrm{monitor}$ and its validation set as $\mathcal{X}_\mathrm{ID}$, while Lost and Found samples serve as $\mathcal{X}_\mathrm{test}$. This setup enables evaluation of semantic shift detection capabilities in naturally occurring, safety-critical scenarios while minimizing covariate shifts due to shared geographical and capture conditions.

\begin{figure}[htbp]
    \centering
    

    \begin{subfigure}{0.47\linewidth}
        \centering
        \includegraphics[width=\linewidth]{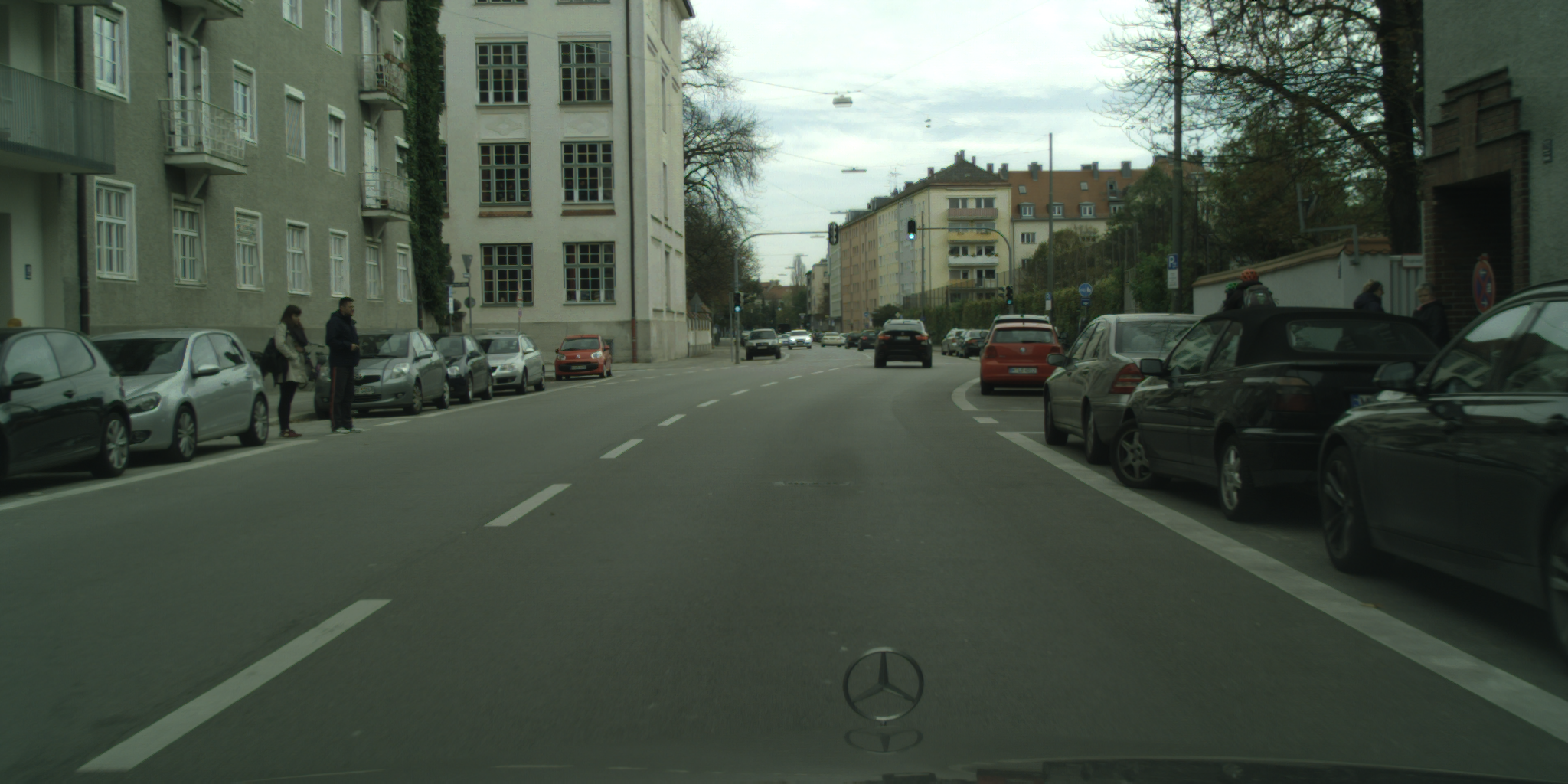}
        \caption{Cityscapes}
    \end{subfigure}
    \hfill
    \begin{subfigure}{0.47\linewidth}
        \centering
        \includegraphics[width=\linewidth]{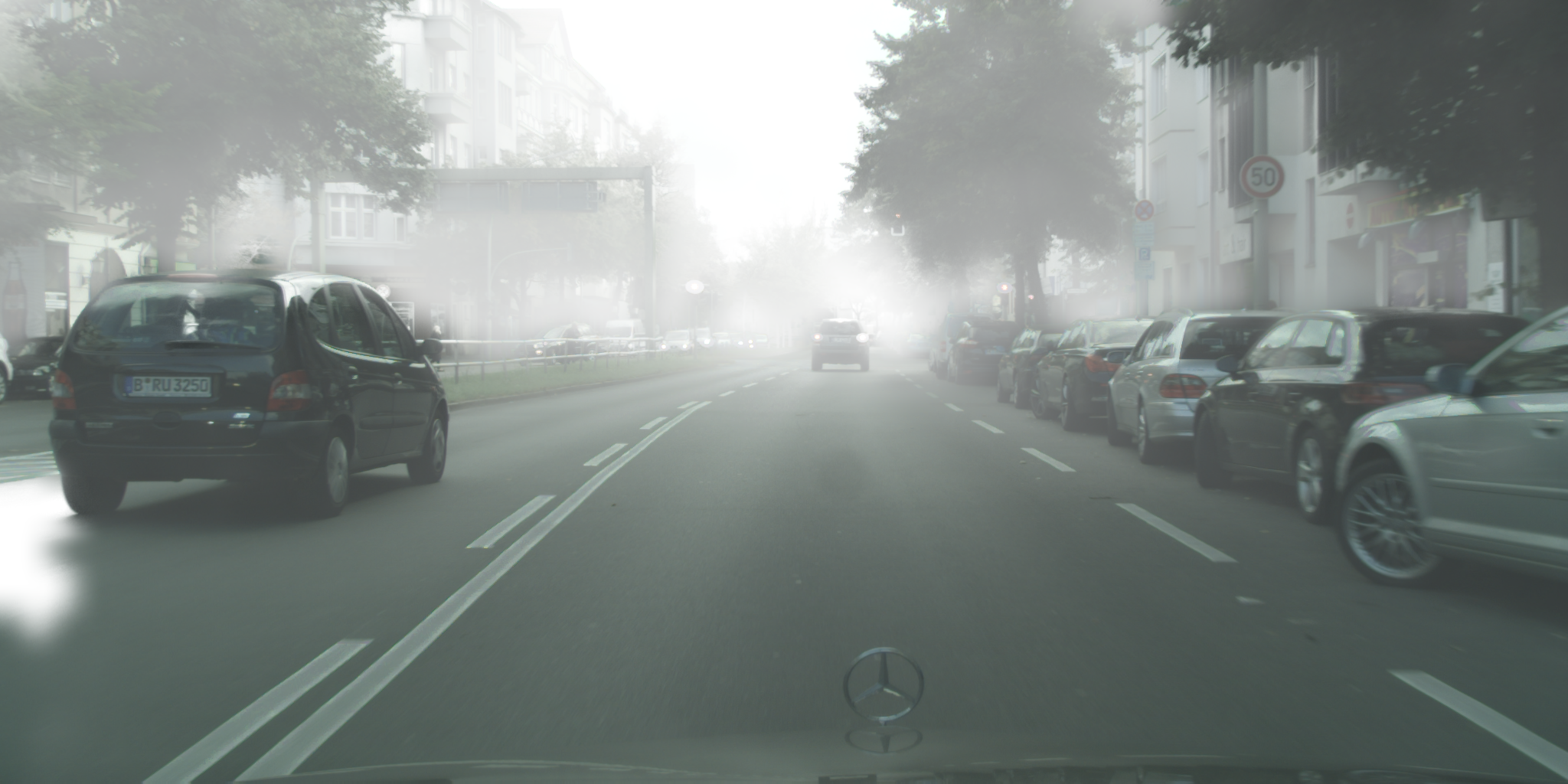}
        \caption{Foggy Cityscapes}
    \end{subfigure}

    \vspace{1cm}

     \begin{subfigure}{0.47\linewidth} 
        \centering
        \includegraphics[width=\linewidth]{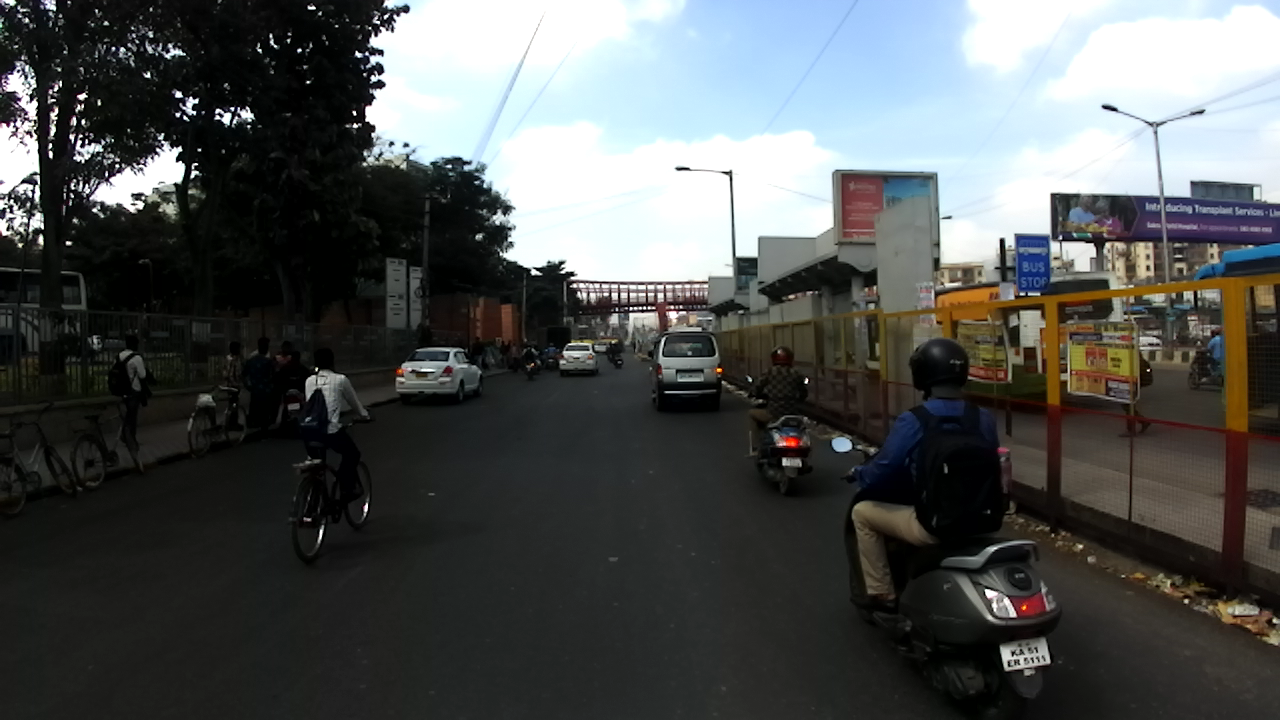}
        \caption{Indian Driving Dataset}
    \end{subfigure}
    \hfill
    \begin{subfigure}{0.47\linewidth} 
        \centering
        \includegraphics[width=\linewidth]{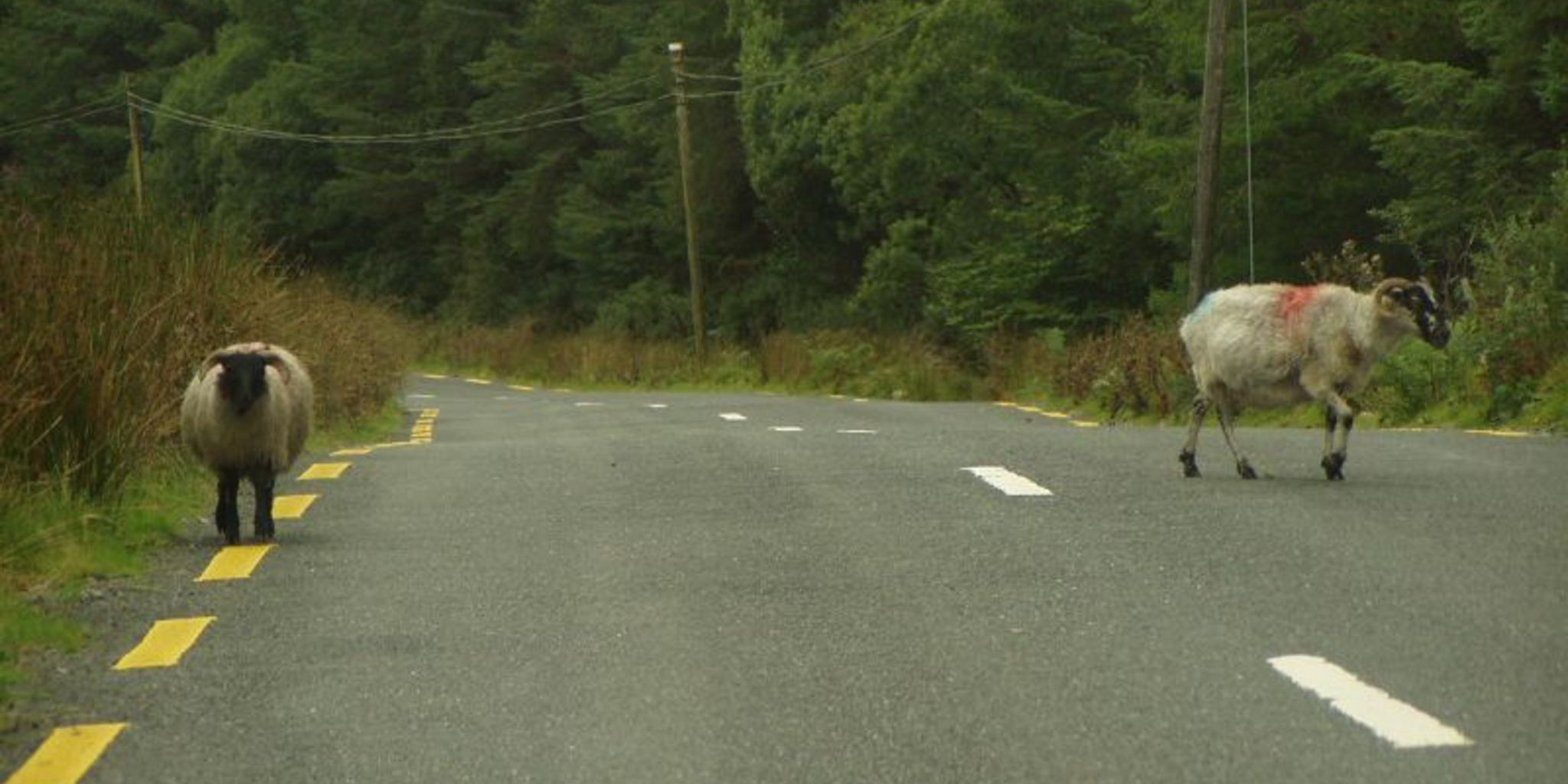}
        \caption{SegmentMeIfYouCan}
    \end{subfigure}
    
    \caption{Example images from Cityscapes~\cite{cordts2016cityscapes}, Cityscapes Fog \cite{sakaridis2018semantic}, Indian Driving dataset\cite{varma2019idd} and SegmentMeIfYouCan \cite{chan2021segmentmeifyoucan}}
    \label{fig:cityscapes_example_images}
\end{figure}

\item[Cityscapes and Foggy Cityscapes Datasets:]
    The Cityscapes~\cite{cordts2016cityscapes} dataset is a widely-used benchmark for semantic urban scene understanding. It contains high-quality pixel-level annotations of images from 50 cities across Germany, captured in various urban driving conditions. The\textbf{ Foggy Cityscapes Dataset} is an adaptation of the original Cityscapes dataset, created by synthetically adding fog to the images. Example images can be seen in Figure \cref{fig:cityscapes_example_images}.

\item[Indian Driving Dataset:]
    The Indian Driving dataset~\cite{varma2019idd} (IDD) is a real-world dataset focused on capturing the complexities of driving scenes in Indian road conditions. It consists of images collected from various urban, rural, and highway scenes across India, representing a diverse set of traffic conditions. An example image can be seen in Figure \cref{fig:cityscapes_example_images}.

\item[SegmentMeIfYouCan Dataset:]
    The SegmentMeIfYouCan \cite{chan2021segmentmeifyoucan} dataset (SMYIC) is designed for evaluating anomaly detection in semantic segmentation tasks. It includes challenging scenarios with diverse visual contexts and synthetic anomalies that closely resemble real-world driving conditions. The dataset focuses on identifying OOD objects that were not present during training. An example image can be seen in Figure~\cref{fig:cityscapes_example_images}.
\end{description}

In our semantic and covariate shift experiments presented in the main paper, we maintain Cityscapes training data as $\mathcal{X}_\mathrm{monitor}$ and its validation set as $\mathcal{X}_\mathrm{ID}$, while samples from IDD~\cite{varma2019idd} and SMIYC~\cite{chan2021segmentmeifyoucan} serve as $\mathcal{X}_\mathrm{test}$. This setup enables evaluation of our approach under real-world scenarios where both types of shifts occur simultaneously.

\begin{table*}[t]
\centering
\begin{tabulary}{\linewidth}{|L||L||L|L|L|}
\hline
\textbf{ID monitor train data $\mathcal{X}_\text{monitor}$} &\textbf{ID test data $\mathcal{X}_{\text{ID}}$}
& \multicolumn{3}{c|}{\textbf{OOD test data $\mathcal{X}_{\text{test}}$}}\\
\cline{3-5}
&& \textbf{Semantic Shift} & \textbf{Covariate Shift} & \textbf{Semantic and Covariate Shift} \\ \hline\hline
\textbf{Cityscapes}\newline train split &\textbf{Cityscapes}\newline validation split & \mbox{Lost and Found}\newline \mbox{Bravo-Synobj} & \mbox{Foggy Cityscapes}\newline \mbox{Bravo-Synflare}\newline 
\mbox{Bravo-Synrain} & \mbox{Indian Driving Dataset}\newline SegmentMeIfYouCan \\ 
\hline
\textbf{ACDC \newline (clear weather)}\newline train split & \textbf{ACDC \newline (clear weather)}\newline validation split & - & ACDC-Fog\newline ACDC-Night\newline ACDC-Rain\newline ACDC-Snow & - \\ 
\hline
\end{tabulary}
\caption{Dataset organization for evaluating different types of distribution shifts.}
\label{tab:dataset_organization}
\end{table*}

Table~\ref{tab:dataset_organization} illustrates our systematic dataset organization. For Cityscapes, we utilize its training set as $\mathcal{X}_\mathrm{monitor}$ and validation set as $\mathcal{X}_\mathrm{ID}$. Similarly for ACDC, we use the reference (clear-weather) images from its training set as $\mathcal{X}_\mathrm{monitor}$ and reference validation set as $\mathcal{X}_\mathrm{ID}$. The remaining datasets are categorized into three distinct evaluation scenarios: semantic shifts (novel objects), covariate shifts (environmental variations), and combined semantic and covariate shifts, each serving as $\mathcal{X}_\mathrm{test}$ for their respective experiments.

\section{Implementation Details}
\label{sec:Implementation Details}

In this section, we detail our feature processing pipeline and implementation specifics. First, we describe the flattening procedure for converting 3-dimensional features from Grounding DINO's~\cite{liu2023grounding} image encoder into single-dimensional vectors. We then present the architecture design and training protocol for the autoencoder models evaluated in \cref{sec:ComparativeBaselineAnalysis}. Finally, we outline our systematic parameter selection process, including kernel optimization for each backbone architecture and the hyperparameter configuration of our normalizing flow models.

\begin{figure*}[htbp]
    \centering
    \includegraphics[width=\linewidth]{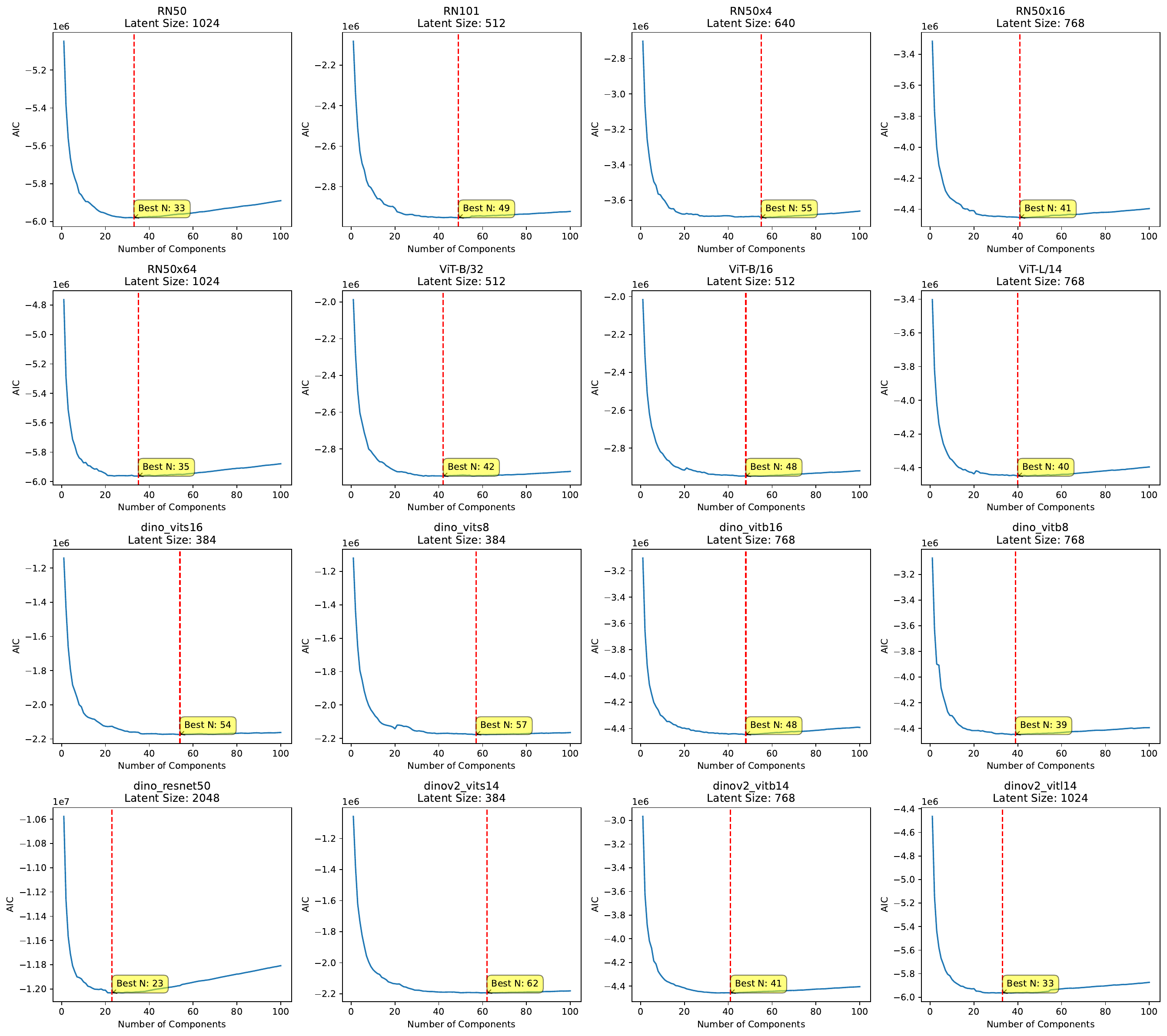}
    \caption{Comparison of AIC values across various model backbones, highlighting the trade-off between model complexity and goodness of fit for GMMs applied to different VFMs. Each plot displays the AIC values for a specific backbone architecture, with the optimal number of GMM components (minimizing AIC) indicated. The size of the latent space vector for each architecture is also annotated.}
    \label{fig:AIC_Values}
\end{figure*}

\paragraph{Selecting Number of Components for GMMs}
\label{sec:SelectingAIC}

To determine the optimal number of components / Gaussians $K$ for the GMM used to model the feature distributions, we employ the Akaike Information Criterion (AIC). The AIC is a model selection technique that strikes a balance between the goodness of fit and the complexity of the model, thereby preventing overfitting.
The AIC for a given GMM with $K$ components is defined as
\begin{equation}
\text{AIC}(K) = 2k(K) - 2\ln(L)
\end{equation}
where $k$ is the number of parameters in the model, and $L$ is the value of the likelihood function for the estimated model, i.e., after likelihood maximization.
For a GMM with $K$ components, each with $d$-dimensional mean vectors and full covariance matrices, the number of parameters is 
\begin{equation}
k(K) = K\left(d + \frac{1}{2}d(d + 1)\right) - 1
\end{equation}
The lower the AIC value, the better the trade-off between goodness of fit and model complexity.
To select the optimal $K$, we fit GMMs with varying numbers of components (e.g., $K = 1, 2, 3, \ldots$) to a hold-out validation set representative of the in-distribution data. For each value of $K$, we compute the corresponding $\text{AIC}(K)$. We then select the value of $K$ that minimizes the AIC, as this represents a balance between model fit and complexity for the given data, which in theory can be optimal under certain assumptions.

\begin{figure*}[htbp]
    \centering
    \includegraphics[width=\linewidth]{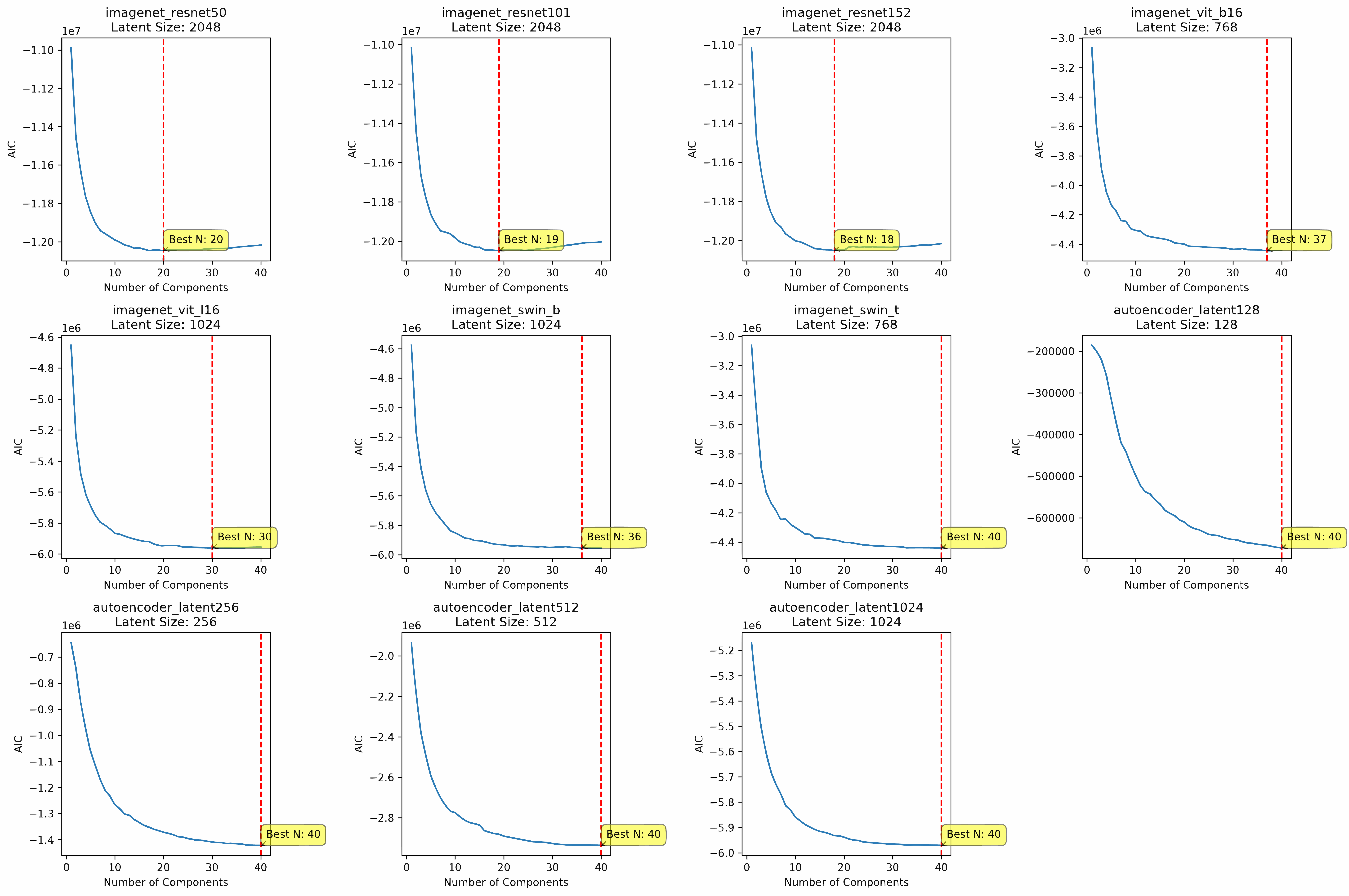}
    \caption{AIC values for ImageNet-trained backbones and autoencoders trained on reference data. Each subplot depicts the AIC profile for a specific architecture or autoencoder configuration, demonstrating the relationship between GMM complexity and fit. The optimal number of components, minimizing AIC, is indicated on each curve.}
    \label{fig:AIC_Values_Benchmark}
\end{figure*}

Figure \ref{fig:AIC_Values} presents a comprehensive evaluation of VFMs. Additionally, Figure \ref{fig:AIC_Values_Benchmark} illustrates the AIC values for ImageNet-trained backbones and autoencoders trained on reference data, corresponding to the experimental setup detailed in \cref{sec:ComparativeBaselineAnalysis}. Each subplot within these figures represents the AIC profile for a specific backbone architecture, elucidating the intricate trade-off between model complexity (quantified by the number of GMM components) and goodness of fit.
The optimal number of components, denoted as $K$, is identified for each backbone as the value corresponding to the global minimum of the AIC curve. This optimal point is explicitly marked on each graph, facilitating direct comparison across different architectural choices. This systematic approach to model selection provides us with a model complexity of the GMM appropriately tailored to the characteristics of each backbone's feature space, potentially enhancing the robustness and efficiency of subsequent anomaly detection tasks.

\paragraph{Grounding DINO Feature Extraction}
\label{sec:GroundingDINOFeatureExtraction}
For feature extraction, we leverage the image encoder of GroundingDINO \cite{liu2023grounding}, which is based on the Swin Transformer \cite{liu2021swin} architecture. Specifically, we extract features from the last three layers of the Swin Transformer and concatenate them to create a multi-scale representation. This approach is analogous to the multi-scale feature utilization in GroundingDINO \cite{liu2023grounding}.

\paragraph{Autoencoder Architecture and Training}
\label{sec:AutoencoderAppendix}

\begin{table}[ht]
\centering
\resizebox{\columnwidth}{!}{%
\begin{tabular}{c}
\hline
Layer \\
\hline
\textbf{Encoder} \\
Input \\
Conv2d (3, 32, 3x3, stride=2, padding=1) \\
BatchNorm2d \\
LeakyReLU \\
Conv2d (32, 64, 3x3, stride=2, padding=1) \\
BatchNorm2d \\
LeakyReLU \\
Conv2d (64, 128, 3x3, stride=2, padding=1) \\
BatchNorm2d \\
LeakyReLU \\
Conv2d (128, 256, 3x3, stride=2, padding=1) \\
BatchNorm2d \\
LeakyReLU \\
Flatten \\
Linear (50,176 → Latent Size) \\
\hline
\textbf{Decoder} \\
Linear (Latent Size → 50,176) \\
Unflatten \\
ConvTranspose2d (256, 128, 3x3, stride=2, padding=1) \\
BatchNorm2d \\
LeakyReLU \\
ConvTranspose2d (128, 64, 3x3, stride=2, padding=1) \\
BatchNorm2d \\
LeakyReLU \\
ConvTranspose2d (64, 32, 3x3, stride=2, padding=1) \\
BatchNorm2d \\
LeakyReLU \\
ConvTranspose2d (32, 3, 3x3, stride=2, padding=1) \\
Sigmoid \\
\hline
\end{tabular}
}
\caption{Architecture of the Autoencoders}
\label{tab:autoencoder_architecture}
\end{table}

The autoencoder is designed to learn compact, latent representations of the reference input data. Once trained, the autoencoder serves as a feature extractor, providing a comparative baseline to compare with the VFMs. The architecture of the autoencoder is detailed in \cref{tab:autoencoder_architecture}.

The training objective for the autoencoder is to minimize the reconstruction error using the mean squared error loss function, defined as
\begin{equation} \mathcal{L} = \frac{1}{n} \sum_{i=1}^{n} (x_i - \hat{x}_i)^2 \end{equation}
where $x_i$ represents the input feature vector, and $\hat{x}_i$ denotes the corresponding reconstructed output. The model is optimized using the Adam optimizer, with a learning rate of $1 \times 10^{-3}$. It is trained for 100 epochs. Early stopping is adopted, triggered by a lack of improvement in the loss.

\paragraph{One-Class SVM Optimization}
\label{sec:OneClassSVM}

To optimize the performance of the One-Class SVM \cite{alam2020one, bounsiar2014one}, we conducted a parameter search across various kernel types and their corresponding hyperparameters. The optimization process focused on three kernel functions: radial basis function (RBF), linear, and polynomial, with the aim of selecting the parameter combination that best fits the ID data. This fit was determined based on the mean decision function score on the training set, with higher scores indicating better alignment with the data.

The parameter space explored in this search includes the following:

\begin{itemize} 
\item \textbf{Kernel types}: RBF, linear, polynomial 
\item \textbf{$\nu$ values}: 0.01, 0.1, 0.5 
\item \textbf{$\gamma$ values (for RBF kernel)}: \textit{scale}, \textit{auto}, 0.1, 1, 10 
\item \textbf{Degree values (for polynomial kernel)}: 2, 3 \end{itemize}

For each backbone type, we systematically evaluated combinations of hyperparameters, selecting the configuration that yielded the highest mean decision function score on the reference data set.

\paragraph{Normalizing Flow Implementation}
\label{sec:NormalizingFlow}

For density estimation using NF, we implement a Real-valued Non-Volume Preserving (Real-NVP) architecture~\cite{dinh2016density}. Our implementation consists of multiple coupling layers with permutation layers and batch normalization between them. Each coupling layer employs a neural network with two hidden layers and ReLU activations, following a half-masking strategy where the first half of the dimensions are used to compute the scale and translation parameters for the second half. We conduct extensive hyperparameter optimization by exploring different architectures with coupling layer topologies of [64, 64], [128, 128], and [256, 256] hidden units, and varying the number of flow steps (2, 4, or 6). The models are trained using the Adam optimizer with a learning rate of $10^{-4}$ and batch sizes of 16 or 32, for either 100 or 200 epochs, using an 80-20 train-validation split for model selection. Following standard practice in flow-based models, we use a standard normal distribution as the base distribution and compute log-probabilities using the change of variables formula with the Jacobian determinant of the transformation.

\section{Detailed Experimental Results}
\label{sec:Detailed Experimental Results}

This section presents the complete set of experimental results discussed in the main paper. In particular. we provide additional tables detailing the performance metrics across all model architectures and backbone configurations for both semantic and covariate shift detection tasks (not only the top 2 backbones as in the space constraint presentation in the main paper).

\subsection{Semantic Shift Experiments}
\label{sec:SemanticShiftFull}

\begin{table*}[ht]
\centering
\resizebox{\textwidth}{!}{
\begin{tabular}{cc||ccc|ccc|ccc|ccc|ccc}
\hline
\multirow{3}{*}{\textbf{Model}} & \multicolumn{1}{c||}{\multirow{3}{*}{\textbf{Backbone}}} & \multicolumn{15}{c}{\textbf{Lost and Found}} \\ \cline{3-17} 
& \multicolumn{1}{c||}{} & \multicolumn{3}{c|}{\textbf{APS}} & \multicolumn{3}{c|}{\textbf{MFS}} & \multicolumn{3}{c|}{\textbf{OC-SVM}} & \multicolumn{3}{c|}{\textbf{GMM}} & \multicolumn{3}{c}{\textbf{NF}} \\ \cline{3-17} 
& \multicolumn{1}{c||}{} & \multicolumn{1}{c|}{\textbf{AUROC}↑} & \multicolumn{1}{c|}{\textbf{AUPR}↑} & \multicolumn{1}{c|}{\textbf{FPR95}↓} & \multicolumn{1}{c|}{\textbf{AUROC}↑} & \multicolumn{1}{c|}{\textbf{AUPR}↑} & \multicolumn{1}{c|}{\textbf{FPR95}↓} & \multicolumn{1}{c|}{\textbf{AUROC}↑} & \multicolumn{1}{c|}{\textbf{AUPR}↑} & \multicolumn{1}{c|}{\textbf{FPR95}↓} & \multicolumn{1}{c|}{\textbf{AUROC}↑} & \multicolumn{1}{c|}{\textbf{AUPR}↑} & \multicolumn{1}{c|}{\textbf{FPR95}↓} & \multicolumn{1}{c|}{\textbf{AUROC}↑} & \multicolumn{1}{c|}{\textbf{AUPR}↑} & \textbf{FPR95}↓ \\ \hline

\multirow{8}{*}{\textbf{CLIP}} 
& \multicolumn{1}{c||}{\textbf{RN50}} & 0.85 & 0.82 & 0.49 & 0.84 & 0.82 & 0.52 & 0.85 & 0.83 & 0.51 & 0.90 & 0.88 & 0.36 & 0.82 & 0.78 & 0.52 \\
& \multicolumn{1}{c||}{\textbf{RN101}} & 0.83 & 0.82 & 0.55 & 0.82 & 0.81 & 0.58 & 0.83 & 0.82 & 0.65 & 0.87 & 0.85 & 0.46 & 0.79 & 0.74 & 0.60 \\
& \multicolumn{1}{c||}{\textbf{RN50x4}} & 0.84 & 0.81 & 0.55 & 0.83 & 0.82 & 0.58 & 0.84 & 0.81 & 0.62 & 0.88 & 0.86 & 0.42 & 0.84 & 0.79 & 0.50 \\
& \multicolumn{1}{c||}{\textbf{RN50x16}} & 0.88 & 0.86 & 0.44 & 0.88 & 0.88 & 0.46 & 0.89 & 0.88 & 0.39 & 0.89 & 0.87 & 0.38 & 0.83 & 0.80 & 0.51 \\
& \multicolumn{1}{c||}{\textbf{RN50x64}} & \textbf{0.94} & \textbf{0.93} & \textbf{0.24} & \textbf{0.93} & \textbf{0.93} & \textbf{0.27} & \textbf{0.94} & \textbf{0.93} & \textbf{0.24} & \textbf{0.94} & \textbf{0.93} & \textbf{0.23} & 0.88 & 0.84 & 0.37 \\
& \multicolumn{1}{c||}{\textbf{ViT-B/32}} & 0.86 & 0.84 & 0.51 & 0.83 & 0.82 & 0.60 & 0.86 & 0.85 & 0.53 & 0.88 & 0.85 & 0.38 & 0.87 & 0.84 & 0.50 \\
& \multicolumn{1}{c||}{\textbf{ViT-B/16}} & 0.88 & 0.86 & 0.38 & 0.86 & 0.85 & 0.41 & 0.89 & 0.88 & 0.43 & 0.91 & 0.89 & 0.33 & 0.91 & 0.88 & 0.34 \\
& \multicolumn{1}{c||}{\textbf{ViT-L/14}} & 0.87 & 0.85 & 0.45 & 0.89 & 0.89 & 0.41 & 0.89 & 0.87 & 0.41 & 0.85 & 0.83 & 0.47 & 0.88 & 0.86 & 0.41 \\ \hline

\multirow{5}{*}{\textbf{DINO}} 
& \multicolumn{1}{c||}{\textbf{ViT-S/16}} & 0.75 & 0.71 & 0.59 & 0.71 & 0.63 & 0.60 & 0.74 & 0.71 & 0.76 & 0.81 & 0.78 & 0.50 & 0.81 & 0.78 & 0.55 \\
& \multicolumn{1}{c||}{\textbf{ViT-S/8}} & 0.71 & 0.65 & 0.61 & 0.66 & 0.59 & 0.66 & 0.75 & 0.71 & 0.60 & 0.82 & 0.81 & 0.51 & 0.79 & 0.77 & 0.60 \\
& \multicolumn{1}{c||}{\textbf{ViT-B/16}} & 0.79 & 0.75 & 0.52 & 0.72 & 0.64 & 0.60 & 0.79 & 0.76 & 0.59 & 0.84 & 0.82 & 0.55 & 0.75 & 0.71 & 0.66 \\
& \multicolumn{1}{c||}{\textbf{ViT-B/8}} & 0.81 & 0.76 & 0.49 & 0.73 & 0.64 & 0.59 & 0.77 & 0.75 & 0.66 & 0.87 & 0.86 & 0.45 & 0.82 & 0.81 & 0.59 \\
& \multicolumn{1}{c||}{\textbf{RN50}} & 0.64 & 0.63 & 0.87 & 0.65 & 0.62 & 0.83 & 0.61 & 0.60 & 0.91 & 0.68 & 0.67 & 0.80 & 0.61 & 0.60 & 0.88 \\ \hline

\multirow{4}{*}{\textbf{DINOv2}} 
& \multicolumn{1}{c||}{\textbf{ViT-S/14}} & 0.86 & 0.80 & 0.34 & 0.86 & 0.80 & 0.36 & 0.88 & 0.83 & 0.31 & 0.81 & 0.76 & 0.51 & 0.78 & 0.74 & 0.65 \\
& \multicolumn{1}{c||}{\textbf{ViT-B/14}} & 0.86 & 0.82 & 0.45 & 0.86 & 0.82 & 0.44 & 0.87 & 0.84 & 0.53 & 0.82 & 0.78 & 0.54 & 0.85 & 0.80 & 0.48 \\
& \multicolumn{1}{c||}{\textbf{ViT-L/14}} & 0.89 & 0.87 & 0.43 & 0.91 & 0.88 & 0.37 & 0.84 & 0.82 & 0.68 & 0.82 & 0.77 & 0.52 & 0.83 & 0.78 & 0.54 \\
& \multicolumn{1}{c||}{\textbf{ViT-G/14}} & 0.90 & 0.87 & 0.30 & 0.89 & 0.86 & 0.35 & 0.85 & 0.82 & 0.63 & 0.86 & 0.78 & 0.39 & 0.83 & 0.75 & 0.46 \\ \hline

\multirow{2}{*}{\textbf{\begin{tabular}[c]{@{}c@{}}Grounding\\ DINO\end{tabular}}} 
& \multicolumn{1}{c||}{\textbf{SwinB}} & 0.88 & 0.84 & 0.47 & 0.82 & 0.76 & 0.52 & 0.84 & 0.80 & 0.54 & 0.91 & 0.89 & 0.32 & 0.95 & 0.94 & \textbf{0.16} \\
& \multicolumn{1}{c||}{\textbf{SwinT}} & 0.85 & 0.80 & 0.53 & 0.79 & 0.73 & 0.60 & 0.83 & 0.79 & 0.59 & 0.92 & 0.90 & 0.28 & \textbf{0.96} & \textbf{0.95} & \textbf{0.16} \\ \hline

& \multicolumn{1}{c||}{} & \multicolumn{15}{c}{\textbf{Bravo-Synobj}} \\ \hline

\multirow{8}{*}{\textbf{CLIP}} 
& \multicolumn{1}{c||}{\textbf{RN50}} & 0.87 & 0.89 & 0.65 & 0.87 & 0.89 & 0.65 & 0.87 & 0.90 & 0.71 & 0.89 & 0.92 & 0.66 & 0.92 & 0.94 & 0.57 \\
& \multicolumn{1}{c||}{\textbf{RN101}} & 0.89 & 0.92 & 0.54 & 0.89 & 0.92 & 0.54 & 0.89 & 0.92 & 0.60 & 0.91 & 0.93 & 0.55 & 0.93 & 0.95 & 0.49 \\
& \multicolumn{1}{c||}{\textbf{RN50x4}} & 0.90 & 0.92 & 0.52 & 0.90 & 0.92 & 0.52 & 0.91 & 0.93 & 0.58 & 0.93 & 0.95 & 0.48 & 0.94 & 0.95 & 0.42 \\
& \multicolumn{1}{c||}{\textbf{RN50x16}} & 0.92 & \textbf{0.94} & \textbf{0.48} & 0.92 & \textbf{0.94} & \textbf{0.49} & 0.92 & 0.94 & \textbf{0.50} & 0.93 & 0.94 & 0.49 & 0.94 & 0.96 & 0.49 \\
& \multicolumn{1}{c||}{\textbf{RN50x64}} & \textbf{0.93} & \textbf{0.94} & 0.54 & \textbf{0.93} & \textbf{0.94} & 0.54 & \textbf{0.93} & \textbf{0.95} & 0.54 & \textbf{0.94} & \textbf{0.95} & 0.53 & 0.94 & 0.96 & 0.43 \\
& \multicolumn{1}{c||}{\textbf{ViT-B/32}} & 0.88 & 0.90 & 0.57 & 0.88 & 0.90 & 0.56 & 0.88 & 0.90 & 0.61 & 0.91 & 0.93 & 0.52 & 0.92 & 0.94 & 0.54 \\
& \multicolumn{1}{c||}{\textbf{ViT-B/16}} & 0.91 & 0.93 & 0.50 & 0.91 & 0.93 & 0.50 & 0.91 & 0.93 & 0.54 & 0.92 & 0.94 & 0.48 & 0.93 & 0.95 & 0.49 \\
& \multicolumn{1}{c||}{\textbf{ViT-L/14}} & 0.92 & \textbf{0.94} & \textbf{0.48} & 0.92 & \textbf{0.94} & 0.48 & 0.92 & 0.94 & 0.53 & 0.93 & \textbf{0.95} & \textbf{0.46} & 0.94 & 0.96 & 0.50 \\ \hline

\multirow{5}{*}{\textbf{DINO}} 
& \multicolumn{1}{c||}{\textbf{ViT-S/16}} & 0.79 & 0.80 & 0.74 & 0.79 & 0.80 & 0.74 & 0.81 & 0.83 & 0.74 & 0.85 & 0.87 & 0.66 & 0.88 & 0.91 & 0.63 \\
& \multicolumn{1}{c||}{\textbf{ViT-S/8}} & 0.83 & 0.85 & 0.69 & 0.83 & 0.85 & 0.69 & 0.84 & 0.87 & 0.76 & 0.87 & 0.90 & 0.61 & 0.89 & 0.91 & 0.64 \\
& \multicolumn{1}{c||}{\textbf{ViT-B/16}} & 0.81 & 0.82 & 0.69 & 0.81 & 0.82 & 0.69 & 0.83 & 0.85 & 0.75 & 0.86 & 0.88 & 0.60 & 0.84 & 0.87 & 0.80 \\
& \multicolumn{1}{c||}{\textbf{ViT-B/8}} & 0.82 & 0.84 & 0.64 & 0.82 & 0.84 & 0.64 & 0.83 & 0.86 & 0.70 & 0.86 & 0.89 & 0.59 & 0.88 & 0.91 & 0.65 \\
& \multicolumn{1}{c||}{\textbf{RN50}} & 0.55 & 0.54 & 0.92 & 0.55 & 0.54 & 0.92 & 0.50 & 0.50 & 0.92 & 0.55 & 0.54 & 0.94 & 0.58 & 0.57 & 0.89 \\ \hline
\multirow{4}{*}{\textbf{DINOv2}} 
& \multicolumn{1}{c||}{\textbf{ViT-S/14}} & 0.75 & 0.77 & 0.77 & 0.75 & 0.77 & 0.77 & 0.75 & 0.78 & 0.82 & 0.79 & 0.82 & 0.77 & 0.85 & 0.88 & 0.74 \\
& \multicolumn{1}{c||}{\textbf{ViT-B/14}} & 0.79 & 0.80 & 0.71 & 0.79 & 0.80 & 0.71 & 0.80 & 0.82 & 0.73 & 0.84 & 0.86 & 0.64 & 0.87 & 0.89 & 0.67 \\
& \multicolumn{1}{c||}{\textbf{ViT-L/14}} & 0.78 & 0.79 & 0.77 & 0.78 & 0.79 & 0.77 & 0.78 & 0.81 & 0.80 & 0.82 & 0.85 & 0.74 & 0.86 & 0.88 & 0.63 \\
& \multicolumn{1}{c||}{\textbf{ViT-G/14}} & 0.79 & 0.80 & 0.73 & 0.79 & 0.80 & 0.73 & 0.81 & 0.83 & 0.69 & 0.82 & 0.85 & 0.71 & 0.86 & 0.89 & 0.69 \\ \hline

\multirow{2}{*}{\textbf{\begin{tabular}[c]{@{}c@{}}Grounding\\ DINO\end{tabular}}} 
& \multicolumn{1}{c||}{\textbf{SwinB}} & 0.60 & 0.56 & 0.83 & 0.61 & 0.56 & 0.82 & 0.58 & 0.54 & 0.85 & 0.76 & 0.68 & 0.58 & \textbf{0.99} & \textbf{0.99} & \textbf{0.02} \\
& \multicolumn{1}{c||}{\textbf{SwinT}} & 0.57 & 0.54 & 0.89 & 0.58 & 0.54 & 0.90 & 0.57 & 0.54 & 0.91 & 0.64 & 0.58 & 0.78 & 0.91 & 0.84 & 0.20 \\ \hline

\end{tabular}
}
\caption{VFM performance comparison across (\ie ResNet, ViT, and Swin) architectures on Lost and Found~\cite{pinggera2016lost} with different intensity values and Bravo-Synobj \cite{loiseau2023reliability} datasets. AUROC↑, AUPR↑, and FPR95↓ metrics shown for APS, MFS, OC-SVM, GMM, and NF methods. Best values in bold.}
\label{tab:Semantic_Shift_Experiments_Full}
\end{table*}

\Cref{tab:Semantic_Shift_Experiments_Full} presents an extensive evaluation complementing our semantic shift analysis from the main paper. encompassing 19 backbone architectures across four VFMs. This comprehensive comparison spans various architectures including ResNet variants (RN50-RN50x64). Vision Transformers (ViT-B/32 to ViT-L/14). and Swin Transformers.

In addition to the findings presented in the main paper. our comprehensive backbone analysis reveals several key insights discussed in the following in more detail:
\begin{description}
\item[Detection Method Sensitivity:] 
    While the main paper highlights the superiority of flow-based methods with Grounding DINO. the full results demonstrate that different VFMs exhibit distinct preferences for detection methods. For instance. CLIP models show remarkable consistency across all four detection methods (APS. OC-SVM. GMM. NF). with RN50x64 maintaining FPR values between 0.37-0.24 on Lost and Found. In contrast. Grounding DINO shows high method sensitivity. with performance varying substantially between GMM (FPR: 0.32) and NF (FPR: 0.16).

\item[Cross-Dataset Generalization:]
    The extended analysis reveals interesting patterns in cross-dataset performance. Models that excel on real-world data don't necessarily maintain their advantage on synthetic data. This is particularly evident in CLIP's RN50x64 performance, which dominates on Lost and Found (AUROC: 0.94) but shows relatively lower performance on Bravo-Synobj (AUROC: 0.93). Conversely. some architectures like CLIP's ViT-L/14 demonstrate more balanced performance across both datasets. suggesting better generalization capabilities.

\item[Resolution Impact:] 
    The full backbone comparison enables analysis of patch size effects in transformer-based architectures. In DINO models, smaller patch sizes (ViT-B/8 vs ViT-B/16) generally lead to improved performance. Particularly evident in GMM results (AUROC: 0.87 vs 0.84 on Lost and Found). This suggests that finer-grained feature extraction benefits semantic shift detection. though at the cost of computational complexity.
\end{description}

\subsection{Covariate Shift Experiments}
\label{sec:CovariateShiftFull}

\begin{table*}[ht]
\centering
\resizebox{\textwidth}{!}{
\begin{tabular}{cc||ccc|ccc|ccc|ccc|ccc}
\hline
\multirow{3}{*}{\textbf{Model}} & \multicolumn{1}{c||}{\multirow{3}{*}{\textbf{Backbone}}} & \multicolumn{15}{c}{\textbf{Foggy Cityscapes (i=0.05)}} \\ \cline{3-17} 
& \multicolumn{1}{c||}{} & \multicolumn{3}{c|}{\textbf{APS}} & \multicolumn{3}{c|}{\textbf{MFS}} & \multicolumn{3}{c|}{\textbf{OC-SVM}} & \multicolumn{3}{c|}{\textbf{GMM}} & \multicolumn{3}{c}{\textbf{NF}} \\ \cline{3-17} 
& \multicolumn{1}{c||}{} & \multicolumn{1}{c|}{\textbf{AUROC}↑} & \multicolumn{1}{c|}{\textbf{AUPR}↑} & \multicolumn{1}{c|}{\textbf{FPR95}↓} & \multicolumn{1}{c|}{\textbf{AUROC}↑} & \multicolumn{1}{c|}{\textbf{AUPR}↑} & \multicolumn{1}{c|}{\textbf{FPR95}↓} & \multicolumn{1}{c|}{\textbf{AUROC}↑} & \multicolumn{1}{c|}{\textbf{AUPR}↑} & \multicolumn{1}{c|}{\textbf{FPR95}↓} & \multicolumn{1}{c|}{\textbf{AUROC}↑} & \multicolumn{1}{c|}{\textbf{AUPR}↑} & \multicolumn{1}{c|}{\textbf{FPR95}↓} & \multicolumn{1}{c|}{\textbf{AUROC}↑} & \multicolumn{1}{c|}{\textbf{AUPR}↑} & \textbf{FPR95}↓ \\ \hline

\multirow{8}{*}{\textbf{CLIP}} 
& \multicolumn{1}{c||}{\textbf{RN50}} & 0.66 & 0.62 & 0.80 & 0.66 & 0.62 & 0.80 & 0.66 & 0.63 & 0.84 & 0.65 & 0.61 & 0.81 & 0.78 & 0.74 & 0.62 \\
& \multicolumn{1}{c||}{\textbf{RN101}} & 0.68 & 0.65 & 0.80 & 0.68 & 0.65 & 0.80 & 0.69 & 0.68 & 0.79 & 0.76 & 0.73 & 0.65 & 0.80 & 0.75 & 0.57 \\
& \multicolumn{1}{c||}{\textbf{RN50x4}} & 0.70 & 0.66 & 0.78 & 0.70 & 0.66 & 0.78 & 0.71 & 0.68 & 0.78 & 0.74 & 0.70 & 0.71 & 0.81 & 0.77 & 0.59 \\
& \multicolumn{1}{c||}{\textbf{RN50x16}} & 0.66 & 0.62 & 0.79 & 0.66 & 0.62 & 0.79 & 0.67 & 0.63 & 0.82 & 0.69 & 0.65 & 0.73 & 0.73 & 0.67 & 0.69 \\
& \multicolumn{1}{c||}{\textbf{RN50x64}} & 0.66 & 0.62 & 0.78 & 0.66 & 0.62 & 0.77 & 0.68 & 0.64 & 0.79 & 0.71 & 0.66 & 0.76 & 0.77 & 0.73 & 0.67 \\
& \multicolumn{1}{c||}{\textbf{ViT-B/32}} & 0.75 & 0.72 & 0.64 & 0.75 & 0.71 & 0.65 & 0.77 & 0.75 & 0.62 & 0.79 & 0.75 & 0.58 & \textbf{0.86} & 0.82 & 0.49 \\
& \multicolumn{1}{c||}{\textbf{ViT-B/16}} & 0.73 & 0.69 & 0.71 & 0.73 & 0.69 & 0.70 & 0.74 & 0.71 & 0.78 & 0.78 & 0.75 & 0.57 & \textbf{0.86} & \textbf{0.83} & 0.48 \\
& \multicolumn{1}{c||}{\textbf{ViT-L/14}} & \textbf{0.80} & \textbf{0.76} & \textbf{0.58} & \textbf{0.80} & \textbf{0.76} & \textbf{0.58} & \textbf{0.83} & \textbf{0.80} & \textbf{0.60} & \textbf{0.84} & \textbf{0.82} & \textbf{0.49} & \textbf{0.86} & 0.82 & 0.46 \\ \hline

\multirow{5}{*}{\textbf{DINO}} 
& \multicolumn{1}{c||}{\textbf{ViT-S/16}} & 0.66 & 0.60 & 0.68 & 0.66 & 0.60 & 0.68 & 0.72 & 0.66 & 0.68 & 0.69 & 0.63 & 0.65 & 0.76 & 0.71 & 0.60 \\
& \multicolumn{1}{c||}{\textbf{ViT-S/8}} & 0.61 & 0.58 & 0.80 & 0.61 & 0.58 & 0.80 & 0.65 & 0.60 & 0.77 & 0.65 & 0.61 & 0.78 & 0.73 & 0.66 & 0.62 \\
& \multicolumn{1}{c||}{\textbf{ViT-B/16}} & 0.65 & 0.60 & 0.73 & 0.65 & 0.60 & 0.73 & 0.71 & 0.65 & 0.69 & 0.67 & 0.63 & 0.72 & 0.75 & 0.72 & 0.74 \\
& \multicolumn{1}{c||}{\textbf{ViT-B/8}} & 0.65 & 0.58 & 0.72 & 0.65 & 0.58 & 0.72 & 0.71 & 0.65 & 0.68 & 0.66 & 0.62 & 0.72 & 0.80 & 0.74 & 0.52 \\
& \multicolumn{1}{c||}{\textbf{RN50}} & 0.59 & 0.56 & 0.88 & 0.59 & 0.56 & 0.88 & 0.58 & 0.57 & 0.90 & 0.60 & 0.57 & 0.85 & 0.57 & 0.56 & 0.91 \\ \hline

\multirow{4}{*}{\textbf{DINOv2}} 
& \multicolumn{1}{c||}{\textbf{ViT-S/14}} & 0.55 & 0.53 & 0.90 & 0.55 & 0.53 & 0.90 & 0.56 & 0.53 & 0.90 & 0.57 & 0.54 & 0.86 & 0.63 & 0.59 & 0.83 \\
& \multicolumn{1}{c||}{\textbf{ViT-B/14}} & 0.58 & 0.55 & 0.88 & 0.58 & 0.55 & 0.88 & 0.60 & 0.57 & 0.92 & 0.59 & 0.56 & 0.85 & 0.66 & 0.61 & 0.77 \\
& \multicolumn{1}{c||}{\textbf{ViT-L/14}} & 0.55 & 0.53 & 0.91 & 0.55 & 0.53 & 0.91 & 0.56 & 0.54 & 0.89 & 0.56 & 0.53 & 0.92 & 0.59 & 0.55 & 0.88 \\
& \multicolumn{1}{c||}{\textbf{ViT-G/14}} & 0.56 & 0.53 & 0.89 & 0.56 & 0.53 & 0.89 & 0.61 & 0.57 & 0.84 & 0.57 & 0.54 & 0.89 & 0.64 & 0.60 & 0.83 \\ \hline

\multirow{2}{*}{\textbf{\begin{tabular}[c]{@{}c@{}}Grounding\\ DINO\end{tabular}}} 
& \multicolumn{1}{c||}{\textbf{SwinB}} & 0.58 & 0.54 & 0.85 & 0.57 & 0.54 & 0.84 & 0.60 & 0.57 & 0.86 & 0.66 & 0.61 & 0.79 & \textbf{0.86} & 0.81 & \textbf{0.36} \\
& \multicolumn{1}{c||}{\textbf{SwinT}} & 0.57 & 0.54 & 0.87 & 0.57 & 0.54 & 0.86 & 0.59 & 0.56 & 0.86 & 0.69 & 0.62 & 0.72 & 0.84 & 0.77 & 0.42 \\ \hline

& \multicolumn{1}{c||}{} & \multicolumn{15}{c}{\textbf{Foggy Cityscapes (i=0.2)}} \\ \hline

\multirow{8}{*}{\textbf{CLIP}} 
& \multicolumn{1}{c||}{\textbf{RN50}} & 0.85 & 0.82 & 0.48 & 0.85 & 0.82 & 0.47 & 0.86 & 0.84 & 0.53 & 0.85 & 0.81 & 0.47 & 0.94 & 0.91 & 0.23 \\
& \multicolumn{1}{c||}{\textbf{RN101}} & 0.87 & 0.84 & 0.43 & 0.87 & 0.84 & 0.44 & 0.88 & 0.88 & 0.47 & 0.94 & 0.93 & 0.26 & 0.94 & 0.92 & 0.20 \\
& \multicolumn{1}{c||}{\textbf{RN50x4}} & 0.90 & 0.87 & 0.36 & 0.90 & 0.87 & 0.36 & 0.91 & 0.89 & 0.37 & 0.94 & 0.93 & 0.26 & 0.96 & 0.95 & 0.18 \\
& \multicolumn{1}{c||}{\textbf{RN50x16}} & 0.88 & 0.86 & 0.40 & 0.88 & 0.86 & 0.40 & 0.89 & 0.87 & 0.42 & 0.90 & 0.88 & 0.31 & 0.92 & 0.88 & 0.29 \\
& \multicolumn{1}{c||}{\textbf{RN50x64}} & 0.90 & 0.86 & 0.30 & 0.90 & 0.86 & 0.31 & 0.91 & 0.89 & 0.32 & 0.92 & 0.91 & 0.26 & 0.94 & 0.92 & 0.21 \\
& \multicolumn{1}{c||}{\textbf{ViT-B/32}} & 0.92 & 0.91 & 0.30 & 0.92 & 0.90 & 0.30 & 0.93 & 0.93 & 0.27 & 0.96 & 0.95 & 0.16 & 0.98 & 0.98 & 0.08 \\
& \multicolumn{1}{c||}{\textbf{ViT-B/16}} & 0.92 & 0.91 & 0.29 & 0.92 & 0.91 & 0.29 & 0.93 & 0.92 & 0.29 & 0.96 & 0.96 & 0.18 & 0.98 & 0.98 & 0.10 \\
& \multicolumn{1}{c||}{\textbf{ViT-L/14}} & \textbf{0.96} & \textbf{0.95} & \textbf{0.14} & \textbf{0.96} & \textbf{0.95} & \textbf{0.15} & \textbf{0.97} & \textbf{0.97} & \textbf{0.11} & \textbf{0.98} & \textbf{0.97} & \textbf{0.07} & 0.97 & 0.96 & 0.10 \\ \hline

\multirow{5}{*}{\textbf{DINO}} 
& \multicolumn{1}{c||}{\textbf{ViT-S/16}} & 0.87 & 0.81 & 0.31 & 0.87 & 0.81 & 0.31 & 0.93 & 0.90 & 0.20 & 0.92 & 0.89 & 0.24 & 0.95 & 0.94 & 0.19 \\
& \multicolumn{1}{c||}{\textbf{ViT-S/8}} & 0.77 & 0.70 & 0.50 & 0.77 & 0.70 & 0.50 & 0.83 & 0.78 & 0.45 & 0.83 & 0.78 & 0.44 & 0.89 & 0.84 & 0.30 \\
& \multicolumn{1}{c||}{\textbf{ViT-B/16}} & 0.85 & 0.78 & 0.34 & 0.85 & 0.78 & 0.34 & 0.91 & 0.88 & 0.22 & 0.89 & 0.85 & 0.31 & 0.89 & 0.88 & 0.34 \\
& \multicolumn{1}{c||}{\textbf{ViT-B/8}} & 0.84 & 0.77 & 0.34 & 0.84 & 0.77 & 0.34 & 0.91 & 0.87 & 0.24 & 0.88 & 0.85 & 0.35 & 0.97 & 0.96 & 0.09 \\
& \multicolumn{1}{c||}{\textbf{RN50}} & 0.73 & 0.68 & 0.70 & 0.73 & 0.68 & 0.70 & 0.70 & 0.67 & 0.79 & 0.73 & 0.67 & 0.73 & 0.65 & 0.63 & 0.84 \\ \hline

\multirow{4}{*}{\textbf{DINOv2}} 
& \multicolumn{1}{c||}{\textbf{ViT-S/14}} & 0.66 & 0.60 & 0.72 & 0.66 & 0.60 & 0.72 & 0.69 & 0.62 & 0.73 & 0.70 & 0.63 & 0.66 & 0.82 & 0.76 & 0.52 \\
& \multicolumn{1}{c||}{\textbf{ViT-B/14}} & 0.69 & 0.63 & 0.73 & 0.69 & 0.63 & 0.73 & 0.71 & 0.66 & 0.77 & 0.74 & 0.68 & 0.66 & 0.84 & 0.77 & 0.48 \\
& \multicolumn{1}{c||}{\textbf{ViT-L/14}} & 0.62 & 0.57 & 0.85 & 0.62 & 0.57 & 0.85 & 0.62 & 0.58 & 0.86 & 0.63 & 0.59 & 0.84 & 0.69 & 0.62 & 0.70 \\
& \multicolumn{1}{c||}{\textbf{ViT-G/14}} & 0.64 & 0.58 & 0.76 & 0.64 & 0.58 & 0.76 & 0.71 & 0.64 & 0.72 & 0.68 & 0.62 & 0.73 & 0.80 & 0.74 & 0.54 \\ \hline

\multirow{2}{*}{\textbf{\begin{tabular}[c]{@{}c@{}}Grounding\\ DINO\end{tabular}}} 
& \multicolumn{1}{c||}{\textbf{SwinB}} & 0.84 & 0.79 & 0.38 & 0.84 & 0.79 & 0.38 & 0.89 & 0.86 & 0.32 & 0.95 & 0.94 & 0.17 & \textbf{1.00} & \textbf{1.00} & \textbf{0.00} \\
& \multicolumn{1}{c||}{\textbf{SwinT}} & 0.87 & 0.84 & 0.34 & 0.87 & 0.84 & 0.34 & 0.90 & 0.89 & 0.30 & 0.96 & 0.94 & 0.10 & \textbf{1.00} & \textbf{1.00} & 0.02 \\ \hline

& \multicolumn{1}{c||}{} & \multicolumn{15}{c}{\textbf{Bravo-Synflare}} \\ \hline

\multirow{8}{*}{\textbf{CLIP}} 
& \multicolumn{1}{c||}{\textbf{RN50}} & 0.80 & 0.81 & 0.67 & 0.80 & 0.81 & 0.68 & 0.79 & 0.80 & 0.77 & 0.85 & 0.86 & 0.54 & 0.89 & 0.90 & 0.42 \\
& \multicolumn{1}{c||}{\textbf{RN101}} & 0.84 & 0.85 & 0.62 & 0.84 & 0.85 & 0.64 & 0.84 & 0.86 & 0.69 & 0.89 & 0.91 & 0.57 & 0.92 & 0.93 & 0.43 \\
& \multicolumn{1}{c||}{\textbf{RN50x4}} & 0.78 & 0.79 & 0.71 & 0.78 & 0.79 & 0.72 & 0.77 & 0.79 & 0.80 & 0.85 & 0.86 & 0.58 & 0.88 & 0.87 & 0.51 \\
& \multicolumn{1}{c||}{\textbf{RN50x16}} & 0.80 & 0.81 & 0.70 & 0.80 & 0.81 & 0.70 & 0.79 & 0.80 & 0.73 & 0.83 & 0.84 & 0.62 & 0.87 & 0.88 & 0.55 \\
& \multicolumn{1}{c||}{\textbf{RN50x64}} & 0.79 & 0.79 & 0.65 & 0.79 & 0.79 & 0.63 & 0.81 & 0.81 & 0.63 & 0.82 & 0.83 & 0.62 & 0.85 & 0.82 & 0.47 \\
& \multicolumn{1}{c||}{\textbf{ViT-B/32}} & 0.84 & 0.84 & 0.60 & 0.84 & 0.84 & 0.61 & 0.83 & 0.84 & 0.62 & 0.91 & 0.91 & 0.47 & 0.93 & 0.94 & 0.37 \\
& \multicolumn{1}{c||}{\textbf{ViT-B/16}} & 0.84 & 0.85 & 0.52 & 0.84 & 0.84 & 0.53 & 0.83 & 0.84 & 0.60 & 0.89 & 0.90 & 0.43 & 0.94 & 0.95 & 0.35 \\
& \multicolumn{1}{c||}{\textbf{ViT-L/14}} & 0.87 & 0.86 & 0.49 & \textbf{0.87} & 0.86 & 0.50 & 0.88 & 0.89 & 0.48 & 0.90 & 0.91 & 0.45 & 0.94 & 0.93 & 0.31 \\ \hline

\multirow{5}{*}{\textbf{DINO}} 
& \multicolumn{1}{c||}{\textbf{ViT-S/16}} & 0.86 & 0.85 & 0.47 & 0.86 & 0.85 & 0.47 & 0.90 & 0.90 & 0.43 & 0.91 & 0.91 & 0.42 & 0.95 & 0.96 & 0.28 \\
& \multicolumn{1}{c||}{\textbf{ViT-S/8}} & 0.82 & 0.81 & 0.55 & 0.82 & 0.81 & 0.55 & 0.86 & 0.85 & 0.51 & 0.87 & 0.88 & 0.54 & 0.93 & 0.93 & 0.35 \\
& \multicolumn{1}{c||}{\textbf{ViT-B/16}} & \textbf{0.87} & \textbf{0.87} & \textbf{0.41} & \textbf{0.87} & \textbf{0.87} & \textbf{0.41} & \textbf{0.91} & 0.91 & \textbf{0.33} & 0.91 & 0.92 & 0.37 & 0.87 & 0.88 & 0.57 \\
& \multicolumn{1}{c||}{\textbf{ViT-B/8}} & \textbf{0.87} & \textbf{0.87} & 0.45 & \textbf{0.87} & \textbf{0.87} & 0.45 & \textbf{0.91} & \textbf{0.92} & 0.35 & 0.90 & 0.91 & 0.51 & 0.95 & 0.96 & 0.33 \\
& \multicolumn{1}{c||}{\textbf{RN50}} & 0.64 & 0.60 & 0.84 & 0.64 & 0.60 & 0.84 & 0.54 & 0.53 & 0.90 & 0.69 & 0.67 & 0.82 & 0.64 & 0.62 & 0.87 \\ \hline

\multirow{4}{*}{\textbf{DINOv2}} 
& \multicolumn{1}{c||}{\textbf{ViT-S/14}} & 0.70 & 0.67 & 0.80 & 0.70 & 0.67 & 0.80 & 0.69 & 0.67 & 0.88 & 0.76 & 0.74 & 0.78 & 0.79 & 0.76 & 0.72 \\
& \multicolumn{1}{c||}{\textbf{ViT-B/14}} & 0.64 & 0.62 & 0.86 & 0.64 & 0.62 & 0.86 & 0.64 & 0.62 & 0.87 & 0.70 & 0.68 & 0.85 & 0.76 & 0.75 & 0.78 \\
& \multicolumn{1}{c||}{\textbf{ViT-L/14}} & 0.60 & 0.58 & 0.88 & 0.60 & 0.58 & 0.88 & 0.60 & 0.58 & 0.89 & 0.61 & 0.61 & 0.88 & 0.67 & 0.65 & 0.80 \\
& \multicolumn{1}{c||}{\textbf{ViT-G/14}} & 0.60 & 0.57 & 0.83 & 0.60 & 0.57 & 0.83 & 0.64 & 0.61 & 0.83 & 0.63 & 0.62 & 0.88 & 0.72 & 0.72 & 0.80 \\ \hline

\multirow{2}{*}{\textbf{\begin{tabular}[c]{@{}c@{}}Grounding\\ DINO\end{tabular}}} 
& \multicolumn{1}{c||}{\textbf{SwinB}} & 0.86 & 0.85 & 0.51 & 0.86 & 0.85 & 0.51 & 0.86 & 0.86 & 0.49 & \textbf{0.95} & \textbf{0.95} & 0.26 & \textbf{0.99} & \textbf{0.99} & \textbf{0.05} \\
& \multicolumn{1}{c||}{\textbf{SwinT}} & 0.85 & 0.84 & 0.49 & 0.85 & 0.84 & 0.49 & 0.84 & 0.84 & 0.55 & 0.94 & 0.94 & \textbf{0.25} & \textbf{0.99} & \textbf{0.99} & 0.07 \\ \hline

\end{tabular}
}
\caption{VFM performance comparison across (\ie ResNet, ViT, and Swin) architectures on Foggy Cityscapes~\cite{pinggera2016lost} with different intensity values and Bravo-Synflare~\cite{loiseau2023reliability} datasets. AUROC↑, AUPR↑, and FPR95↓ metrics shown for APS, MFS, OC-SVM, GMM, and NF methods. Best values in bold.}
\label{tab:Covariate_Shift_Experiments_1}
\end{table*}

\cref{tab:Covariate_Shift_Experiments_1,tab:Covariate_Shift_Experiments_2,tab:Covariate_Shift_Experiments_3,tab:Covariate_Shift_Experiments_4} present comprehensive covariate shift experiments that extend the analysis introduced in the main paper. Specifically, \cref{tab:Covariate_Shift_Experiments_1,tab:Covariate_Shift_Experiments_2} demonstrate experimental results where the monitoring set $\mathcal{X}_\mathrm{monitor}$ is derived from the Cityscapes training set, while \cref{tab:Covariate_Shift_Experiments_3,tab:Covariate_Shift_Experiments_4} utilize the ACDC reference training set as $\mathcal{X}_\mathrm{monitor}$.

Extending beyond the main paper's findings, our comprehensive backbone analysis reveals several key insights:
\begin{description}
\item[Performance Variation in Environmental Conditions:]
    Analysis of Grounding DINO reveals a notable performance discrepancy between synthetic and natural environmental shifts. While achieving near-perfect detection on synthetic perturbations (FPR95: 0.00-0.02 on Foggy Cityscapes i=0.2), performance degrades on natural conditions, particularly in ACDC Rain (FPR95: 0.34-0.35). This degradation is especially interesting as ACDC Rain represents post-rain conditions (\eg, wet road surfaces, residual moisture) rather than active precipitation. .
    \item[Architecture-Specific Scaling Patterns:]
    Our analysis reveals non-linear scaling behaviors across architectural families. Within the CLIP family, model size does not correlate monotonically with performance. While ViT-L/14 demonstrates superior performance on synthetic perturbations in Foggy Cityscapes, this advantage diminishes or reverses in natural conditions such as ACDC Rain, where RN50x64 achieves markedly better performance (FPR95: 0.17 vs 0.41). These findings suggest that architectural design choices may be more crucial than model capacity for robust covariate shift detection.
    \end{description}

\begin{table*}[ht]
\centering
\resizebox{\textwidth}{!}{
\begin{tabular}{cc||ccc|ccc|ccc|ccc|ccc}
\hline
\multirow{3}{*}{\textbf{Model}} & \multicolumn{1}{c||}{\multirow{3}{*}{\textbf{Backbone}}} & \multicolumn{15}{c}{\textbf{Bravo-Synrain}} \\ \cline{3-17} 
& \multicolumn{1}{c||}{} & \multicolumn{3}{c|}{\textbf{APS}} & \multicolumn{3}{c|}{\textbf{MFS}} & \multicolumn{3}{c|}{\textbf{OC-SVM}} & \multicolumn{3}{c|}{\textbf{GMM}} & \multicolumn{3}{c}{\textbf{NF}} \\ \cline{3-17} 
& \multicolumn{1}{c||}{} & \multicolumn{1}{c|}{\textbf{AUROC}↑} & \multicolumn{1}{c|}{\textbf{AUPR}↑} & \multicolumn{1}{c|}{\textbf{FPR95}↓} & \multicolumn{1}{c|}{\textbf{AUROC}↑} & \multicolumn{1}{c|}{\textbf{AUPR}↑} & \multicolumn{1}{c|}{\textbf{FPR95}↓} & \multicolumn{1}{c|}{\textbf{AUROC}↑} & \multicolumn{1}{c|}{\textbf{AUPR}↑} & \multicolumn{1}{c|}{\textbf{FPR95}↓} & \multicolumn{1}{c|}{\textbf{AUROC}↑} & \multicolumn{1}{c|}{\textbf{AUPR}↑} & \multicolumn{1}{c|}{\textbf{FPR95}↓} & \multicolumn{1}{c|}{\textbf{AUROC}↑} & \multicolumn{1}{c|}{\textbf{AUPR}↑} & \textbf{FPR95}↓ \\ \hline

\multirow{8}{*}{\textbf{CLIP}} 
& \multicolumn{1}{c||}{\textbf{RN50}} & 0.76 & 0.72 & 0.59 & 0.76 & 0.72 & 0.60 & 0.75 & 0.71 & 0.64 & 0.85 & 0.82 & 0.43 & 0.95 & 0.92 & 0.16 \\
& \multicolumn{1}{c||}{\textbf{RN101}} & 0.83 & 0.81 & 0.50 & 0.83 & 0.81 & 0.50 & 0.84 & 0.82 & 0.51 & 0.92 & 0.91 & 0.25 & 0.94 & 0.93 & 0.17 \\
& \multicolumn{1}{c||}{\textbf{RN50x4}} & 0.85 & 0.82 & 0.48 & 0.85 & 0.82 & 0.48 & 0.85 & 0.82 & 0.48 & 0.93 & 0.92 & 0.24 & 0.95 & 0.93 & 0.14 \\
& \multicolumn{1}{c||}{\textbf{RN50x16}} & 0.94 & 0.93 & 0.25 & 0.94 & 0.93 & 0.25 & 0.94 & 0.94 & 0.22 & 0.96 & 0.95 & 0.17 & 0.97 & 0.96 & 0.10 \\
& \multicolumn{1}{c||}{\textbf{RN50x64}} & 0.99 & 0.99 & 0.03 & 0.99 & 0.99 & 0.03 & \textbf{1.00} & \textbf{1.00} & \textbf{0.01} & \textbf{1.00} & \textbf{1.00} & \textbf{0.01} & 0.99 & 0.99 & 0.04 \\
& \multicolumn{1}{c||}{\textbf{ViT-B/32}} & 0.82 & 0.79 & 0.53 & 0.82 & 0.79 & 0.54 & 0.82 & 0.80 & 0.50 & 0.91 & 0.89 & 0.31 & 0.96 & 0.95 & 0.13 \\
& \multicolumn{1}{c||}{\textbf{ViT-B/16}} & 0.90 & 0.87 & 0.34 & 0.90 & 0.87 & 0.34 & 0.91 & 0.89 & 0.31 & 0.94 & 0.93 & 0.20 & 0.97 & 0.95 & 0.12 \\
& \multicolumn{1}{c||}{\textbf{ViT-L/14}} & 0.89 & 0.85 & 0.31 & 0.89 & 0.85 & 0.32 & 0.92 & 0.90 & 0.27 & 0.94 & 0.91 & 0.22 & 0.97 & 0.94 & 0.09 \\ \hline

\multirow{5}{*}{\textbf{DINO}} 
& \multicolumn{1}{c||}{\textbf{ViT-S/16}} & 0.87 & 0.83 & 0.33 & 0.87 & 0.83 & 0.33 & 0.93 & 0.91 & 0.20 & 0.94 & 0.93 & 0.21 & \textbf{1.00} & \textbf{1.00} & 0.02 \\
& \multicolumn{1}{c||}{\textbf{ViT-S/8}} & 0.89 & 0.84 & 0.30 & 0.89 & 0.84 & 0.30 & 0.95 & 0.93 & 0.16 & 0.96 & 0.95 & 0.13 & 0.99 & 0.99 & 0.02 \\
& \multicolumn{1}{c||}{\textbf{ViT-B/16}} & 0.90 & 0.87 & 0.25 & 0.90 & 0.87 & 0.25 & 0.96 & 0.95 & 0.11 & 0.95 & 0.94 & 0.17 & 0.99 & 0.99 & 0.03 \\
& \multicolumn{1}{c||}{\textbf{ViT-B/8}} & 0.95 & 0.94 & 0.15 & 0.95 & 0.94 & 0.15 & 0.99 & 0.99 & 0.06 & 0.98 & 0.98 & 0.10 & \textbf{1.00} & \textbf{1.00} & \textbf{0.00} \\
& \multicolumn{1}{c||}{\textbf{RN50}} & 0.49 & 0.47 & 0.92 & 0.49 & 0.47 & 0.92 & 0.40 & 0.42 & 0.97 & 0.56 & 0.52 & 0.87 & 0.65 & 0.64 & 0.85 \\ \hline

\multirow{4}{*}{\textbf{DINOv2}} 
& \multicolumn{1}{c||}{\textbf{ViT-S/14}} & 0.89 & 0.83 & 0.33 & 0.89 & 0.83 & 0.33 & 0.89 & 0.84 & 0.35 & 0.96 & 0.92 & 0.16 & 0.91 & 0.84 & 0.20 \\
& \multicolumn{1}{c||}{\textbf{ViT-B/14}} & 0.80 & 0.75 & 0.58 & 0.80 & 0.75 & 0.58 & 0.81 & 0.76 & 0.61 & 0.88 & 0.83 & 0.45 & 0.94 & 0.90 & 0.18 \\
& \multicolumn{1}{c||}{\textbf{ViT-L/14}} & 0.73 & 0.67 & 0.64 & 0.73 & 0.67 & 0.64 & 0.74 & 0.68 & 0.67 & 0.80 & 0.75 & 0.56 & 0.88 & 0.83 & 0.37 \\
& \multicolumn{1}{c||}{\textbf{ViT-G/14}} & 0.71 & 0.65 & 0.66 & 0.71 & 0.65 & 0.66 & 0.77 & 0.71 & 0.57 & 0.79 & 0.74 & 0.61 & 0.90 & 0.87 & 0.35 \\ \hline

\multirow{2}{*}{\textbf{\begin{tabular}[c]{@{}c@{}}Grounding\\ DINO\end{tabular}}} 
& \multicolumn{1}{c||}{\textbf{SwinB}} & \textbf{1.00} & \textbf{1.00} & \textbf{0.02} & \textbf{1.00} & \textbf{1.00} & \textbf{0.02} & \textbf{1.00} & 0.99 & 0.02 & \textbf{1.00} & \textbf{1.00} & \textbf{0.00} & \textbf{1.00} & \textbf{1.00} & \textbf{0.00} \\
& \multicolumn{1}{c||}{\textbf{SwinT}} & 0.99 & 0.99 & \textbf{0.02} & 0.99 & 0.99 & \textbf{0.02} & 0.99 & 0.99 & 0.03 & \textbf{1.00} & \textbf{1.00} & \textbf{0.00} & \textbf{1.00} & \textbf{1.00} & \textbf{0.00} \\ \hline

\end{tabular}
}
\caption{VFM performance comparison across (\ie ResNet, ViT, and Swin) architectures on Bravo-Synrain~\cite{loiseau2023reliability} dataset. AUROC↑, AUPR↑, and FPR95↓ metrics shown for APS, MFS, OC-SVM, GMM, and NF methods. Best values in bold.}
\label{tab:Covariate_Shift_Experiments_2}
\end{table*}

\begin{table*}[]
\centering
\resizebox{\textwidth}{!}{
\begin{tabular}{cc||ccc|ccc|ccc|ccc|ccc}
\hline
\multirow{3}{*}{\textbf{Model}} & \multicolumn{1}{c||}{\multirow{3}{*}{\textbf{Backbone}}} & \multicolumn{15}{c}{\textbf{ACDC Fog }} \\ \cline{3-17} 
& \multicolumn{1}{c||}{} & \multicolumn{3}{c|}{\textbf{APS}} & \multicolumn{3}{c|}{\textbf{MFS}} & \multicolumn{3}{c|}{\textbf{OC-SVM}} & \multicolumn{3}{c|}{\textbf{GMM}} & \multicolumn{3}{c}{\textbf{NF}} \\ \cline{3-17} 
& \multicolumn{1}{c||}{} & \multicolumn{1}{c|}{\textbf{AUROC}↑} & \multicolumn{1}{c|}{\textbf{AUPR}↑} & \multicolumn{1}{c|}{\textbf{FPR95}↓} & \multicolumn{1}{c|}{\textbf{AUROC}↑} & \multicolumn{1}{c|}{\textbf{AUPR}↑} & \multicolumn{1}{c|}{\textbf{FPR95}↓} & \multicolumn{1}{c|}{\textbf{AUROC}↑} & \multicolumn{1}{c|}{\textbf{AUPR}↑} & \multicolumn{1}{c|}{\textbf{FPR95}↓} & \multicolumn{1}{c|}{\textbf{AUROC}↑} & \multicolumn{1}{c|}{\textbf{AUPR}↑} & \multicolumn{1}{c|}{\textbf{FPR95}↓} & \multicolumn{1}{c|}{\textbf{AUROC}↑} & \multicolumn{1}{c|}{\textbf{AUPR}↑} & \textbf{FPR95}↓ \\ \hline

\multirow{8}{*}{\textbf{CLIP}} 
& \multicolumn{1}{c||}{\textbf{RN50}} & 0.95 & 0.94 & 0.22 & 0.95 & 0.94 & 0.22 & 0.94 & 0.93 & 0.22 & 0.98 & 0.98 & 0.07 & 0.97 & 0.94 & 0.09 \\
& \multicolumn{1}{c||}{\textbf{RN101}} & 0.94 & 0.94 & 0.25 & 0.94 & 0.94 & 0.25 & 0.94 & 0.93 & 0.28 & 0.98 & 0.97 & 0.12 & 0.98 & 0.96 & 0.09 \\
& \multicolumn{1}{c||}{\textbf{RN50x4}} & 0.97 & 0.97 & 0.13 & 0.97 & 0.97 & 0.13 & 0.97 & 0.96 & 0.16 & 0.99 & 0.98 & 0.07 & 0.98 & 0.98 & 0.06 \\
& \multicolumn{1}{c||}{\textbf{RN50x16}} & 0.97 & 0.97 & 0.11 & 0.97 & 0.97 & 0.11 & 0.97 & 0.97 & 0.12 & 0.98 & 0.98 & 0.08 & 0.98 & 0.96 & 0.08 \\
& \multicolumn{1}{c||}{\textbf{RN50x64}} & \textbf{0.98} & \textbf{0.98} & \textbf{0.07} & \textbf{0.98} & \textbf{0.98} & \textbf{0.07} & \textbf{0.99} & \textbf{0.99} & \textbf{0.06} & 0.98 & 0.98 & 0.06 & \textbf{0.99} & 0.98 & 0.04 \\
& \multicolumn{1}{c||}{\textbf{ViT-B/32}} & 0.97 & 0.97 & 0.12 & 0.97 & 0.97 & 0.12 & 0.97 & 0.97 & 0.11 & 0.99 & \textbf{0.99} & 0.03 & \textbf{0.99} & \textbf{0.99} & 0.04 \\
& \multicolumn{1}{c||}{\textbf{ViT-B/16}} & 0.95 & 0.95 & 0.20 & 0.95 & 0.95 & 0.20 & 0.95 & 0.95 & 0.18 & 0.98 & 0.98 & 0.07 & 0.98 & 0.98 & 0.07 \\
& \multicolumn{1}{c||}{\textbf{ViT-L/14}} & \textbf{0.98} & 0.97 & 0.09 & \textbf{0.98} & 0.97 & 0.09 & 0.98 & 0.97 & 0.09 & \textbf{1.00} & \textbf{0.99} & \textbf{0.02} & \textbf{0.99} & 0.98 & \textbf{0.03} \\ \hline

\multirow{5}{*}{\textbf{DINO}} 
& \multicolumn{1}{c||}{\textbf{ViT-S/16}} & 0.80 & 0.77 & 0.57 & 0.80 & 0.77 & 0.57 & 0.79 & 0.76 & 0.60 & 0.90 & 0.88 & 0.34 & 0.87 & 0.84 & 0.44 \\
& \multicolumn{1}{c||}{\textbf{ViT-S/8}} & 0.68 & 0.63 & 0.71 & 0.68 & 0.63 & 0.71 & 0.70 & 0.65 & 0.70 & 0.77 & 0.70 & 0.56 & 0.81 & 0.77 & 0.52 \\
& \multicolumn{1}{c||}{\textbf{ViT-B/16}} & 0.74 & 0.70 & 0.66 & 0.74 & 0.70 & 0.66 & 0.74 & 0.70 & 0.65 & 0.86 & 0.82 & 0.43 & 0.85 & 0.83 & 0.50 \\
& \multicolumn{1}{c||}{\textbf{ViT-B/8}} & 0.81 & 0.77 & 0.48 & 0.81 & 0.77 & 0.48 & 0.83 & 0.80 & 0.46 & 0.90 & 0.88 & 0.35 & 0.91 & 0.90 & 0.35 \\
& \multicolumn{1}{c||}{\textbf{RN50}} & 0.92 & 0.91 & 0.42 & 0.92 & 0.91 & 0.42 & 0.92 & 0.92 & 0.38 & 0.90 & 0.90 & 0.38 & 0.64 & 0.65 & 0.89 \\ \hline

\multirow{4}{*}{\textbf{DINOv2}} 
& \multicolumn{1}{c||}{\textbf{ViT-S/14}} & 0.82 & 0.79 & 0.54 & 0.82 & 0.79 & 0.54 & 0.79 & 0.75 & 0.58 & 0.91 & 0.87 & 0.23 & 0.90 & 0.83 & 0.24 \\
& \multicolumn{1}{c||}{\textbf{ViT-B/14}} & 0.74 & 0.71 & 0.67 & 0.74 & 0.71 & 0.67 & 0.73 & 0.68 & 0.69 & 0.84 & 0.78 & 0.46 & 0.85 & 0.79 & 0.44 \\
& \multicolumn{1}{c||}{\textbf{ViT-L/14}} & 0.61 & 0.58 & 0.84 & 0.61 & 0.58 & 0.84 & 0.60 & 0.56 & 0.79 & 0.72 & 0.66 & 0.69 & 0.75 & 0.67 & 0.59 \\
& \multicolumn{1}{c||}{\textbf{ViT-G/14}} & 0.62 & 0.60 & 0.82 & 0.62 & 0.60 & 0.82 & 0.67 & 0.63 & 0.71 & 0.77 & 0.73 & 0.63 & 0.83 & 0.77 & 0.46 \\ \hline

\multirow{2}{*}{\textbf{\begin{tabular}[c]{@{}c@{}}Grounding\\ DINO\end{tabular}}} 
& \multicolumn{1}{c||}{\textbf{SwinB}} & 0.76 & 0.73 & 0.57 & 0.75 & 0.72 & 0.58 & 0.82 & 0.81 & 0.50 & 0.96 & 0.95 & 0.16 & 0.98 & 0.97 & 0.08 \\
& \multicolumn{1}{c||}{\textbf{SwinT}} & 0.80 & 0.79 & 0.54 & 0.80 & 0.78 & 0.55 & 0.83 & 0.83 & 0.52 & 0.98 & 0.97 & 0.08 & 0.96 & 0.95 & 0.13 \\ \hline

& \multicolumn{1}{c||}{} & \multicolumn{15}{c}{\textbf{ACDC Night}} \\ \hline

\multirow{8}{*}{\textbf{CLIP}} 
& \multicolumn{1}{c||}{\textbf{RN50}} & 0.94 & 0.91 & 0.17 & 0.94 & 0.91 & 0.17 & 0.92 & 0.89 & 0.23 & 0.93 & 0.90 & 0.27 & 0.92 & 0.84 & 0.19 \\
& \multicolumn{1}{c||}{\textbf{RN101}} & 0.96 & 0.94 & 0.15 & 0.96 & 0.94 & 0.16 & 0.92 & 0.89 & 0.25 & 0.87 & 0.83 & 0.39 & 0.91 & 0.84 & 0.22 \\
& \multicolumn{1}{c||}{\textbf{RN50x4}} & \textbf{0.98} & \textbf{0.98} & \textbf{0.07} & \textbf{0.98} & \textbf{0.98} & \textbf{0.07} & 0.94 & 0.93 & 0.19 & 0.87 & 0.83 & 0.43 & 0.83 & 0.75 & 0.34 \\
& \multicolumn{1}{c||}{\textbf{RN50x16}} & 0.97 & 0.96 & 0.09 & 0.97 & 0.96 & 0.10 & 0.95 & 0.93 & 0.16 & 0.85 & 0.82 & 0.46 & 0.88 & 0.78 & 0.27 \\
& \multicolumn{1}{c||}{\textbf{RN50x64}} & 0.97 & 0.96 & 0.09 & 0.97 & 0.96 & 0.09 & \textbf{0.96} & \textbf{0.95} & \textbf{0.12} & 0.96 & 0.94 & 0.13 & 0.90 & 0.84 & 0.23 \\
& \multicolumn{1}{c||}{\textbf{ViT-B/32}} & 0.90 & 0.88 & 0.32 & 0.90 & 0.88 & 0.32 & 0.87 & 0.84 & 0.37 & 0.91 & 0.88 & 0.27 & 0.90 & 0.84 & 0.23 \\
& \multicolumn{1}{c||}{\textbf{ViT-B/16}} & 0.91 & 0.89 & 0.25 & 0.91 & 0.89 & 0.26 & 0.89 & 0.85 & 0.31 & 0.88 & 0.85 & 0.34 & 0.93 & 0.88 & 0.16 \\
& \multicolumn{1}{c||}{\textbf{ViT-L/14}} & 0.91 & 0.88 & 0.27 & 0.91 & 0.88 & 0.27 & 0.88 & 0.85 & 0.38 & 0.86 & 0.82 & 0.35 & 0.92 & 0.86 & 0.20 \\ \hline

\multirow{5}{*}{\textbf{DINO}} 
& \multicolumn{1}{c||}{\textbf{ViT-S/16}} & 0.90 & 0.85 & 0.23 & 0.90 & 0.85 & 0.23 & 0.89 & 0.83 & 0.28 & 0.82 & 0.76 & 0.45 & 0.87 & 0.82 & 0.36 \\
& \multicolumn{1}{c||}{\textbf{ViT-S/8}} & 0.85 & 0.77 & 0.31 & 0.85 & 0.77 & 0.31 & 0.85 & 0.77 & 0.32 & 0.76 & 0.67 & 0.48 & 0.82 & 0.78 & 0.50 \\
& \multicolumn{1}{c||}{\textbf{ViT-B/16}} & 0.88 & 0.82 & 0.26 & 0.88 & 0.82 & 0.26 & 0.88 & 0.82 & 0.28 & 0.87 & 0.82 & 0.33 & 0.94 & 0.92 & 0.23 \\
& \multicolumn{1}{c||}{\textbf{ViT-B/8}} & 0.93 & 0.88 & 0.18 & 0.93 & 0.88 & 0.18 & 0.92 & 0.87 & 0.20 & 0.78 & 0.71 & 0.48 & 0.91 & 0.88 & 0.27 \\
& \multicolumn{1}{c||}{\textbf{RN50}} & 0.90 & 0.87 & 0.34 & 0.90 & 0.87 & 0.34 & 0.84 & 0.82 & 0.57 & 0.92 & 0.90 & 0.26 & 0.71 & 0.68 & 0.76 \\ \hline

\multirow{4}{*}{\textbf{DINOv2}} 
& \multicolumn{1}{c||}{\textbf{ViT-S/14}} & 0.89 & 0.83 & 0.25 & 0.89 & 0.83 & 0.25 & 0.85 & 0.78 & 0.35 & 0.77 & 0.71 & 0.63 & 0.79 & 0.70 & 0.53 \\
& \multicolumn{1}{c||}{\textbf{ViT-B/14}} & 0.86 & 0.79 & 0.35 & 0.86 & 0.79 & 0.35 & 0.85 & 0.78 & 0.39 & 0.85 & 0.79 & 0.40 & 0.82 & 0.75 & 0.46 \\
& \multicolumn{1}{c||}{\textbf{ViT-L/14}} & 0.84 & 0.78 & 0.42 & 0.84 & 0.78 & 0.42 & 0.85 & 0.79 & 0.39 & 0.84 & 0.78 & 0.48 & 0.85 & 0.79 & 0.39 \\
& \multicolumn{1}{c||}{\textbf{ViT-G/14}} & 0.87 & 0.81 & 0.35 & 0.87 & 0.81 & 0.35 & 0.88 & 0.83 & 0.31 & 0.89 & 0.86 & 0.31 & 0.89 & 0.86 & 0.31 \\ \hline

\multirow{2}{*}{\textbf{\begin{tabular}[c]{@{}c@{}}Grounding\\ DINO\end{tabular}}} 
& \multicolumn{1}{c||}{\textbf{SwinB}} & 0.96 & 0.95 & 0.13 & 0.96 & 0.95 & 0.13 & 0.97 & 0.96 & 0.13 & 0.98 & 0.97 & \textbf{0.10} & \textbf{0.99} & \textbf{0.99} & \textbf{0.02} \\
& \multicolumn{1}{c||}{\textbf{SwinT}} & 0.96 & 0.95 & 0.13 & 0.96 & 0.95 & 0.13 & 0.96 & 0.95 & 0.14 & \textbf{0.99} & \textbf{0.99} & 0.04 & 0.98 & 0.98 & 0.08 \\ \hline

\end{tabular}
}
\caption{VFM performance comparison across (\ie ResNet, ViT, and Swin) architectures on ACDC Fog and ACDC Night~\cite{sakaridis2021acdc} datasets. AUROC↑, AUPR↑, and FPR95↓ metrics shown for APS, MFS, OC-SVM, GMM, and NF methods. Best values in bold.}
\label{tab:Covariate_Shift_Experiments_3}
\end{table*}

\begin{table*}[ht]
\centering
\resizebox{\textwidth}{!}{
\begin{tabular}{cc||ccc|ccc|ccc|ccc|ccc}
\hline
\multirow{3}{*}{\textbf{Model}} & \multicolumn{1}{c||}{\multirow{3}{*}{\textbf{Backbone}}} & \multicolumn{15}{c}{\textbf{ACDC Rain}} \\ \cline{3-17} 
& \multicolumn{1}{c||}{} & \multicolumn{3}{c|}{\textbf{APS}} & \multicolumn{3}{c|}{\textbf{MFS}} & \multicolumn{3}{c|}{\textbf{OC-SVM}} & \multicolumn{3}{c|}{\textbf{GMM}} & \multicolumn{3}{c}{\textbf{NF}} \\ \cline{3-17} 
& \multicolumn{1}{c||}{} & \multicolumn{1}{c|}{\textbf{AUROC}↑} & \multicolumn{1}{c|}{\textbf{AUPR}↑} & \multicolumn{1}{c|}{\textbf{FPR95}↓} & \multicolumn{1}{c|}{\textbf{AUROC}↑} & \multicolumn{1}{c|}{\textbf{AUPR}↑} & \multicolumn{1}{c|}{\textbf{FPR95}↓} & \multicolumn{1}{c|}{\textbf{AUROC}↑} & \multicolumn{1}{c|}{\textbf{AUPR}↑} & \multicolumn{1}{c|}{\textbf{FPR95}↓} & \multicolumn{1}{c|}{\textbf{AUROC}↑} & \multicolumn{1}{c|}{\textbf{AUPR}↑} & \multicolumn{1}{c|}{\textbf{FPR95}↓} & \multicolumn{1}{c|}{\textbf{AUROC}↑} & \multicolumn{1}{c|}{\textbf{AUPR}↑} & \textbf{FPR95}↓ \\ \hline

\multirow{8}{*}{\textbf{CLIP}} 
& \multicolumn{1}{c||}{\textbf{RN50}} & 0.89 & 0.86 & 0.36 & 0.89 & 0.86 & 0.37 & 0.90 & 0.87 & 0.32 & \textbf{0.93} & \textbf{0.92} & 0.24 & 0.94 & 0.90 & 0.21 \\
& \multicolumn{1}{c||}{\textbf{RN101}} & 0.87 & 0.84 & 0.42 & 0.87 & 0.84 & 0.42 & 0.87 & 0.84 & 0.39 & 0.91 & 0.89 & 0.30 & 0.91 & 0.87 & 0.25 \\
& \multicolumn{1}{c||}{\textbf{RN50x4}} & 0.89 & 0.86 & 0.36 & 0.89 & 0.86 & 0.36 & 0.90 & 0.88 & 0.30 & \textbf{0.93} & \textbf{0.92} & \textbf{0.22} & 0.93 & 0.89 & 0.21 \\
& \multicolumn{1}{c||}{\textbf{RN50x16}} & 0.89 & 0.86 & 0.31 & 0.89 & 0.86 & 0.31 & 0.90 & 0.87 & 0.28 & 0.91 & 0.89 & 0.28 & 0.93 & 0.89 & 0.19 \\
& \multicolumn{1}{c||}{\textbf{RN50x64}} & \textbf{0.95} & \textbf{0.94} & \textbf{0.17} & \textbf{0.95} & \textbf{0.94} & \textbf{0.17} & \textbf{0.96} & \textbf{0.96} & \textbf{0.14} & 0.92 & 0.90 & 0.26 & 0.92 & 0.87 & 0.22 \\
& \multicolumn{1}{c||}{\textbf{ViT-B/32}} & 0.85 & 0.83 & 0.41 & 0.85 & 0.83 & 0.41 & 0.87 & 0.84 & 0.39 & 0.92 & 0.90 & 0.26 & \textbf{0.95} & \textbf{0.92} & \textbf{0.18} \\
& \multicolumn{1}{c||}{\textbf{ViT-B/16}} & 0.84 & 0.79 & 0.44 & 0.84 & 0.79 & 0.44 & 0.86 & 0.81 & 0.38 & 0.91 & 0.89 & 0.27 & 0.90 & 0.85 & 0.25 \\
& \multicolumn{1}{c||}{\textbf{ViT-L/14}} & 0.84 & 0.78 & 0.41 & 0.84 & 0.78 & 0.42 & 0.87 & 0.81 & 0.35 & 0.86 & 0.83 & 0.42 & 0.92 & 0.87 & 0.23 \\ \hline

\multirow{5}{*}{\textbf{DINO}} 
& \multicolumn{1}{c||}{\textbf{ViT-S/16}} & 0.78 & 0.73 & 0.64 & 0.78 & 0.73 & 0.64 & 0.80 & 0.76 & 0.60 & 0.82 & 0.80 & 0.61 & 0.84 & 0.84 & 0.52 \\
& \multicolumn{1}{c||}{\textbf{ViT-S/8}} & 0.80 & 0.77 & 0.62 & 0.80 & 0.77 & 0.62 & 0.82 & 0.80 & 0.54 & 0.83 & 0.83 & 0.55 & 0.86 & 0.87 & 0.56 \\
& \multicolumn{1}{c||}{\textbf{ViT-B/16}} & 0.81 & 0.77 & 0.56 & 0.81 & 0.77 & 0.56 & 0.84 & 0.80 & 0.53 & 0.85 & 0.84 & 0.53 & 0.81 & 0.83 & 0.68 \\
& \multicolumn{1}{c||}{\textbf{ViT-B/8}} & 0.82 & 0.80 & 0.51 & 0.82 & 0.80 & 0.51 & 0.85 & 0.82 & 0.47 & 0.84 & 0.84 & 0.52 & 0.84 & 0.86 & 0.58 \\
& \multicolumn{1}{c||}{\textbf{RN50}} & 0.66 & 0.62 & 0.78 & 0.66 & 0.62 & 0.78 & 0.59 & 0.57 & 0.88 & 0.70 & 0.67 & 0.81 & 0.55 & 0.54 & 0.91 \\ \hline

\multirow{4}{*}{\textbf{DINOv2}} 
& \multicolumn{1}{c||}{\textbf{ViT-S/14}} & 0.75 & 0.71 & 0.69 & 0.75 & 0.71 & 0.69 & 0.78 & 0.74 & 0.70 & 0.79 & 0.78 & 0.66 & 0.84 & 0.82 & 0.62 \\
& \multicolumn{1}{c||}{\textbf{ViT-B/14}} & 0.70 & 0.67 & 0.74 & 0.70 & 0.67 & 0.74 & 0.71 & 0.68 & 0.71 & 0.71 & 0.70 & 0.78 & 0.73 & 0.69 & 0.75 \\
& \multicolumn{1}{c||}{\textbf{ViT-L/14}} & 0.63 & 0.61 & 0.80 & 0.63 & 0.61 & 0.80 & 0.67 & 0.65 & 0.79 & 0.67 & 0.66 & 0.83 & 0.75 & 0.72 & 0.67 \\
& \multicolumn{1}{c||}{\textbf{ViT-G/14}} & 0.64 & 0.61 & 0.77 & 0.64 & 0.61 & 0.77 & 0.71 & 0.67 & 0.69 & 0.72 & 0.69 & 0.71 & 0.78 & 0.73 & 0.58 \\ \hline

\multirow{2}{*}{\textbf{\begin{tabular}[c]{@{}c@{}}Grounding\\ DINO\end{tabular}}} 
& \multicolumn{1}{c||}{\textbf{SwinB}} & 0.75 & 0.70 & 0.63 & 0.76 & 0.71 & 0.62 & 0.72 & 0.66 & 0.67 & 0.77 & 0.73 & 0.60 & 0.89 & 0.85 & 0.35 \\
& \multicolumn{1}{c||}{\textbf{SwinT}} & 0.71 & 0.66 & 0.71 & 0.71 & 0.66 & 0.70 & 0.67 & 0.62 & 0.68 & 0.81 & 0.76 & 0.52 & 0.89 & 0.86 & 0.34 \\ \hline

& \multicolumn{1}{c||}{} & \multicolumn{15}{c}{\textbf{ACDC Snow}} \\ \hline

\multirow{8}{*}{\textbf{CLIP}} 
& \multicolumn{1}{c||}{\textbf{RN50}} & 0.98 & 0.97 & 0.08 & 0.98 & 0.97 & 0.08 & 0.98 & 0.97 & 0.08 & 0.99 & 0.99 & 0.05 & 0.98 & 0.96 & 0.08 \\
& \multicolumn{1}{c||}{\textbf{RN101}} & 0.96 & 0.94 & 0.14 & 0.96 & 0.94 & 0.14 & 0.96 & 0.95 & 0.12 & 0.98 & 0.97 & 0.09 & 0.98 & 0.97 & 0.06 \\
& \multicolumn{1}{c||}{\textbf{RN50x4}} & 0.98 & 0.97 & 0.08 & 0.98 & 0.97 & 0.08 & 0.98 & 0.98 & 0.06 & \textbf{0.99} & \textbf{0.99} & 0.07 & \textbf{0.99} & \textbf{0.99} & 0.04 \\
& \multicolumn{1}{c||}{\textbf{RN50x16}} & 0.98 & 0.97 & 0.07 & 0.98 & 0.97 & 0.07 & 0.98 & 0.97 & 0.06 & \textbf{0.99} & \textbf{0.99} & \textbf{0.04} & 0.98 & 0.96 & 0.05 \\
& \multicolumn{1}{c||}{\textbf{RN50x64}} & \textbf{0.99} & \textbf{0.98} & \textbf{0.05} & \textbf{0.99} & \textbf{0.98} & \textbf{0.05} & \textbf{0.99} & \textbf{0.99} & \textbf{0.04} & \textbf{0.99} & 0.98 & 0.05 & \textbf{0.99} & 0.98 & \textbf{0.03} \\
& \multicolumn{1}{c||}{\textbf{ViT-B/32}} & 0.96 & 0.96 & 0.12 & 0.96 & 0.96 & 0.12 & 0.97 & 0.96 & 0.11 & \textbf{0.99} & \textbf{0.99} & 0.05 & \textbf{0.99} & 0.99 & \textbf{0.03} \\
& \multicolumn{1}{c||}{\textbf{ViT-B/16}} & 0.95 & 0.94 & 0.16 & 0.95 & 0.94 & 0.16 & 0.95 & 0.94 & 0.15 & \textbf{0.99} & \textbf{0.99} & 0.05 & \textbf{0.99} & 0.98 & 0.05 \\
& \multicolumn{1}{c||}{\textbf{ViT-L/14}} & 0.93 & 0.90 & 0.20 & 0.93 & 0.90 & 0.20 & 0.94 & 0.92 & 0.15 & 0.97 & 0.97 & 0.11 & 0.98 & 0.97 & 0.05 \\ \hline

\multirow{5}{*}{\textbf{DINO}} 
& \multicolumn{1}{c||}{\textbf{ViT-S/16}} & 0.81 & 0.76 & 0.46 & 0.81 & 0.76 & 0.46 & 0.83 & 0.79 & 0.45 & 0.87 & 0.84 & 0.43 & 0.86 & 0.83 & 0.51 \\
& \multicolumn{1}{c||}{\textbf{ViT-S/8}} & 0.76 & 0.70 & 0.58 & 0.76 & 0.70 & 0.58 & 0.79 & 0.73 & 0.52 & 0.81 & 0.76 & 0.48 & 0.89 & 0.87 & 0.41 \\
& \multicolumn{1}{c||}{\textbf{ViT-B/16}} & 0.82 & 0.76 & 0.48 & 0.82 & 0.76 & 0.48 & 0.84 & 0.79 & 0.42 & 0.86 & 0.83 & 0.43 & 0.88 & 0.86 & 0.48 \\
& \multicolumn{1}{c||}{\textbf{ViT-B/8}} & 0.83 & 0.78 & 0.40 & 0.83 & 0.78 & 0.40 & 0.87 & 0.82 & 0.33 & 0.89 & 0.86 & 0.38 & 0.92 & 0.92 & 0.34 \\
& \multicolumn{1}{c||}{\textbf{RN50}} & 0.80 & 0.76 & 0.59 & 0.80 & 0.76 & 0.59 & 0.74 & 0.72 & 0.72 & 0.82 & 0.78 & 0.59 & 0.61 & 0.60 & 0.90 \\ \hline

\multirow{4}{*}{\textbf{DINOv2}} 
& \multicolumn{1}{c||}{\textbf{ViT-S/14}} & 0.91 & 0.86 & 0.22 & 0.91 & 0.86 & 0.22 & 0.94 & 0.90 & 0.18 & 0.94 & 0.93 & 0.28 & 0.96 & 0.93 & 0.15 \\
& \multicolumn{1}{c||}{\textbf{ViT-B/14}} & 0.81 & 0.76 & 0.51 & 0.81 & 0.76 & 0.51 & 0.85 & 0.80 & 0.43 & 0.87 & 0.85 & 0.49 & 0.94 & 0.91 & 0.23 \\
& \multicolumn{1}{c||}{\textbf{ViT-L/14}} & 0.74 & 0.67 & 0.63 & 0.74 & 0.67 & 0.63 & 0.78 & 0.72 & 0.55 & 0.79 & 0.74 & 0.62 & 0.88 & 0.83 & 0.38 \\
& \multicolumn{1}{c||}{\textbf{ViT-G/14}} & 0.71 & 0.65 & 0.66 & 0.71 & 0.65 & 0.66 & 0.78 & 0.72 & 0.48 & 0.82 & 0.77 & 0.49 & 0.91 & 0.86 & 0.28 \\ \hline

\multirow{2}{*}{\textbf{\begin{tabular}[c]{@{}c@{}}Grounding\\ DINO\end{tabular}}} 
& \multicolumn{1}{c||}{\textbf{SwinB}} & 0.79 & 0.75 & 0.59 & 0.79 & 0.75 & 0.58 & 0.80 & 0.75 & 0.52 & 0.93 & 0.92 & 0.24 & \textbf{0.99} & 0.98 & 0.04 \\
& \multicolumn{1}{c||}{\textbf{SwinT}} & 0.76 & 0.71 & 0.59 & 0.76 & 0.71 & 0.60 & 0.75 & 0.70 & 0.59 & 0.94 & 0.92 & 0.24 & 0.98 & 0.97 & 0.07 \\ \hline

\end{tabular}
}
\caption{VFM performance comparison across (\ie ResNet, ViT, and Swin) architectures on ACDC Rain and ACDC Snow~\cite{sakaridis2021acdc} datasets. AUROC↑, AUPR↑, and FPR95↓ metrics shown for APS, MFS, OC-SVM, GMM, and NF methods. Best values in bold.}
\label{tab:Covariate_Shift_Experiments_4}
\end{table*}

\section{Limitations and Future Directions}
\label{sec:Limitations_and_Future_Directions}

\begin{description}
\item[Utilizing text-encoders:]
    Our current work focuses exclusively on leveraging the image encoding capabilities of VFMs. However, many VFMs, such as Grounding-DINO~\cite{liu2023grounding} and CLIP~\cite{radford2021learning} feature a shared latent space enabling cross-modal alignment between image and text representations. Our approach currently utilizes only image-level representations to detect distribution shifts. As a result, our approach primarily addresses distribution shifts at the image level, leaving the potential of text-based representations unexplored.

    Incorporating text representations could provide richer characterization of different types of distribution shifts, where the text encoder could generate human-interpretable descriptions of both semantic and covariate shifts, offering valuable insights for subsequent mitigation strategies and system interpretability.
    
\item[Combination with Pixel-wise OOD Segmentation:]
    Recent advances by Ren et~al.\ \cite{ren2024grounding} demonstrate the feasibility of real-time deployment through Grounding-DINO Edge, suggesting that dual-encoder VFMs can be effectively integrated into the operational domain of AD systems. This integration of real-time cross-modal analysis presents a promising direction for future research in robust perception systems, potentially enabling more comprehensive and interpretable OOD detection.

\item[Towards Treatment of OOD Samples:]
    This work solely concentrated on the hard problem to reliably detect OOD input images.
    However, the benefit of the monitor only becomes practically relevant, if it is combined with subsequent mitigation techniques that handle OOD cases.
    Our experiments assumed perfect mitigation techniques (we simply filtered OOD samples out), equivalent to a 100\% reliable handover to a 100\% reliable backup driver. Apart from handovers, alternative backup routines could be, e.g., to apply a safe(r) state (e.g., get to a hold), or query a potentially more expensive expert model for which better results are expected.
    For future work it will be highly interesting to see whether the VFM OOD detectors provide valuable information about the OOD samples that can be leveraged for the OOD case treatment.

\end{description}

\end{document}